\documentclass[sigconf]{acmart}
\newcommand{\eg}{e.g.}
\newcommand{\ie}{i.e.}
\usepackage{url}  %Required
\usepackage{graphicx}  %Required
\usepackage{color}
\usepackage{amsmath}
\usepackage{multirow}
\usepackage{natbib}
\usepackage{algorithm}
\usepackage{algorithmic}
\usepackage{calc}

\usepackage{tikz}
\usepackage{forest}
\usetikzlibrary{trees,positioning,shapes,shadows,arrows.meta}

\usepackage{bm}

\DeclareMathOperator{\E}{\mathbb{E}}

\usepackage{balance} 
\newcommand{\hide}[1]{}

\newcommand{\zh}[1]{{\textsf{\textcolor{violet}{}}}}

\newcommand{\etal}{\textit{et al.}}
\usepackage{pifont}% http://ctan.org/pkg/pifont
\newcommand{\cmark}{\ding{51}}%
\newcommand{\xmark}{\ding{55}}%

\AtBeginDocument{%
  }

\setcopyright{acmlicensed}
\copyrightyear{2018}
\acmYear{2018}
\acmDOI{XXXXXXX.XXXXXXX}

\acmConference[Conference acronym 'XX]{}{June 03--05,
  2018}{Woodstock, NY}
\acmISBN{978-1-4503-XXXX-X/18/06}

\begin{document}

\title{Heterogeneous Contrastive Learning for \\ Foundation Models and Beyond}

% \affiliation{%

\author{Lecheng Zheng}
\affiliation{%
  \institution{University of Illinois at Urbana-Champaign}
  \country{}}
\email{lecheng4@illinois.edu}

\author{Baoyu Jing}
\affiliation{%
  \institution{University of Illinois at Urbana-Champaign}
  \country{}}
\email{baoyu2@illinois.edu}

\author{Zihao Li}
\affiliation{%
  \institution{University of Illinois at Urbana-Champaign}
  \country{}}
\email{zihaoli5@illinois.edu}

\author{Hanghang Tong}
\affiliation{%
  \institution{University of Illinois at Urbana-Champaign}
  \country{}}
\email{htong@illinois.edu}

\author{Jingrui He}
\affiliation{%
  \institution{University of Illinois at Urbana-Champaign}
  \country{}}
\email{jingrui@illinois.edu}

\renewcommand{\shortauthors}{Zheng et al.}

\begin{abstract}
In the era of big data and Artificial Intelligence, an emerging paradigm is to utilize contrastive self-supervised learning to model large-scale heterogeneous data. Many existing foundation models benefit from the generalization capability of contrastive self-supervised learning by learning compact and high-quality representations without relying on any label information. Amidst the explosive advancements in foundation models across multiple domains, including natural language processing and computer vision, a thorough survey on heterogeneous contrastive learning for the foundation model is urgently needed. In response, this survey critically evaluates the current landscape of heterogeneous contrastive learning for foundation models, highlighting the open challenges and future trends of contrastive learning. In particular, we first present how the recent advanced contrastive learning-based methods deal with view heterogeneity and how contrastive learning is applied to train and fine-tune the multi-view foundation models. Then, we move to contrastive learning methods for task heterogeneity, including pretraining tasks and downstream tasks, and show how different tasks are combined with contrastive learning loss for different purposes. Finally, we conclude this survey by discussing the open challenges and shedding light on the future directions of contrastive learning.
\end{abstract}

\maketitle

\section{Introduction}
% \lc{Need to update the introduction .} 
Recent years have witnessed the rapid growth of the volume of big data. A Forbes report shows that the amount of newly created data in the past several years had increased by more than two trillion gigabytes\footnote{\url{https://www.forbes.com/sites/gilpress/2020/01/06/6-predictions-about-data-in-2020-and-the-coming-decade/?sh=3214c68f4fc3}}. One major characteristic of big data is heterogeneity~\cite{wang2017heterogeneous}. Specifically, big data are usually collected from multiple sources and associated with various tasks, exhibiting view or task heterogeneity. For instance, in a social media platform, such as Facebook or Twitter, a post usually consists of a mixture of multiple types of data, such as a recorded video or several photos along with the text description. In the financial domain, taking the stock market for instance, the collected data may include not only the numerical values (\eg, stock price, the statistics from a company quarter report) but also some textual data conveying important information (\eg, a piece of news about a pharmaceutical company receiving the approval from Food and Drug Administration for its new product). 

% The past year has witnessed the rapid development of the foundation models in various domains, such as ChatGPT in Natural Language Processing (NLP), Vision Transformer~\cite{DBLP:conf/iclr/DosovitskiyB0WZ21} in Computer vision, etc. These models greatly improve our lives by enabling quick information access, increasing productivity through automating repetitive tasks, efficient proofreading, etc. 

In response to the challenges posed by the exponential growth of big data, a promising approach is emerging by leveraging contrastive self-supervised learning to pre-train foundational models tailored for large-scale heterogeneous datasets. Recently, Contrastive Learning (CL) has gained an increasing interest in training foundation models~\cite{chen2020simple, gao2021simcse, DBLP:conf/acl/JainZAWNLTNRBMX23}, due to its good generalization capability and the independence of labeled data. Amidst the explosive advancements in foundation models across multiple domains, including natural language processing and computer vision, there is an urgent need for a comprehensive survey on heterogeneous contrastive learning for foundational models.

\begin{table*}[th]
\begin{center}
\scalebox{0.9}{
\begin{tabular}{|c|c|c|c|c|c|c|}
\hline
      &     & \multicolumn{4}{c|}{\bf Research Topics and Coverage}\\ \hline
{\bf Learning Paradigms}    & Survey Papers        & \multicolumn{1}{c|}{View Heterogeneity} & \multicolumn{1}{c|}{Task/Label Heterogeneity} & \multicolumn{1}{c|}{Contrastive Learning} & \multicolumn{1}{c|}{Foundation Model}  \\ \hline
\multirow{6}{*}{} &  Li \etal, 2018~\cite{li2018survey}   &      \cmark  &      \xmark   &      \xmark     &     \xmark    \\ \cline{2-6}
                    % &  Xu \etal, 2013~\cite{xu2013survey}  &     \cmark   &      \xmark   &   \xmark    &  \xmark\\ \cline{2-6}
                    &  Zhang \etal, 2018~\cite{zhang2018binary}   &    \xmark    &    \cmark    &    \xmark      &  \xmark     \\ \cline{2-6}
Non-Contrastive     &  Yan \etal, 2021~\cite{yan2021deep} &    \cmark    &     \xmark  &    \xmark  &   \xmark  \\ \cline{2-6}
Learning            &  Jin \etal, 2023~\cite{DBLP:journals/corr/abs-2312-02783} & \cmark  & \xmark  & \xmark    & \cmark \\ \cline{2-6} 
                    &  Xu \etal, 2024~\cite{xu2024survey}   &   \cmark     &   \xmark     &   \xmark       &     \cmark  \\ \hline     
\multirow{6}{*}{} & Jaiswa \etal~ 2020~\cite{jaiswal2020survey}  & \xmark  & \xmark  & \cmark    & \xmark \\ \cline{2-6}
     & Le-Khac \etal~ 2020~\cite{le2020contrastive} & \xmark  & \xmark  & \cmark    & \xmark \\ \cline{2-6}
Contrastive     & Liu \etal, 2023~\cite{DBLP:journals/tkde/LiuZHMWZT23}    &   \xmark      &    \xmark     &   \cmark       &   \xmark     \\ \cline{2-6}
Learning         & Albelwi 2022~\cite{DBLP:journals/entropy/Albelwi22}    &   \xmark     &  \cmark       & \cmark          &  \xmark     \\ \cline{2-6}
      % &     &        &        &          &       \\ \cline{2-6}
      &  {\bf This survey}   &  \cmark       &   \cmark     &  \cmark        & \cmark       \\ \hline
\end{tabular}
}
\end{center}
\caption{Comparison with the existing related survey papers.}
\label{survey_comparison}
\vspace{-5mm}
\end{table*}
% \subsection{Major difference from the other survey papers}
However, existing surveys on this topic are limited in scope and fail to systematically evaluate the most advanced techniques. Previous survey papers~\cite{yan2021deep, li2018survey, zhang2018binary, jaiswal2020survey, le2020contrastive, DBLP:journals/corr/abs-2312-02783, xu2024survey, DBLP:journals/tkde/LiuZHMWZT23, DBLP:journals/entropy/Albelwi22} mainly focus on investigating single heterogeneity~\cite{li2018survey, zhang2018binary, yan2021deep} (\eg, view heterogeneity, task heterogeneity), contrastive learning~\cite{jaiswal2020survey, le2020contrastive, DBLP:journals/tkde/LiuZHMWZT23, DBLP:journals/entropy/Albelwi22} or multi-modal foundation model~\cite{DBLP:journals/corr/abs-2312-02783, xu2024survey}. The comparison of these surveys is summarized in Table \ref{survey_comparison}. Specifically, \cite{yan2021deep, zhang2018binary, li2018survey} solely focus on heterogeneous machine learning, (\eg, multi-view learning, multi-label learning) and they do not cover any topic about contrastive learning and foundation model. \cite{jaiswal2020survey, le2020contrastive} discuss some contrastive learning methods at the early stage and they fail to include the most recent advanced techniques; ~\cite{DBLP:journals/entropy/Albelwi22, DBLP:journals/tkde/LiuZHMWZT23} investigate the recent advances in contrastive learning, but these two papers are only limited to summarizing the traditional contrastive learning methods. ~\cite{xu2024survey, DBLP:journals/corr/abs-2312-02783} introduce the multi-modal foundation models, but their topics are only limited to multi-modal large language models. This survey critically evaluates the current landscape of heterogeneous contrastive learning for foundation models from both view and task heterogeneities, highlighting the open challenges and future trends of contrastive learning.

Our contributions are summarized as follows:
\begin{itemize}
    \item \textbf{Categorization of Contrastive Foundation Models.} We systematically review the contrastive foundation models and categorize the existing methods into two branches, including the contrastive foundation models for view heterogeneity and task heterogeneity.
    \item \textbf{Systematic Review of Techniques.} We provide a comprehensive review of heterogeneous contrastive learning for foundation models. For both view heterogeneity and task heterogeneity, we summarize the representative methods and make necessary comparisons.
    \item \textbf{Future Directions.} We summarize four possible research directions on heterogeneous contrastive foundation models for future exploration.
\end{itemize}

This paper is organized as follows. In Section 2, we briefly review the basic concept of contrastive learning, and in Section 3, we first introduce the traditional 
% \hh{remove 'small'? 'traditional' already suggests that these works are not for 'large foundation models'}\lc{updated.}
multi-view contrastive learning model as the basis and then present the multi-view contrastive learning for large foundation models. In Section 4, we summarize contrastive learning methods for task heterogeneity, including pretraining tasks and downstream tasks, and show how different tasks are combined with contrastive learning loss for different purposes. In Section 5, we present several open future directions in contrastive learning before we conclude this survey paper in Section 6.

% \begin{figure*}
% \centering
% \includegraphics[width=1.05\linewidth]{taxonomy.png} \\
% \caption{Taxonomy of this Survey with Representative Works.\by{maybe we should replace the authors with the method names.}}
% \label{taxonomy}
% \end{figure*}

\begin{figure*}
    \centering
    
% I am using only the basic, xnode, and tnode styles
\tikzset{
    basic/.style  = {draw, text width=1.4cm, align=center, fill=pink!10!green!20!, font=\scriptsize, rectangle},
    % root/.style   = {basic, rounded corners=2pt, thin, align=center, fill=green!30, text width=2.5cm},
    onode/.style = {basic, thin, rounded corners=2pt, align=center, fill=green!10!blue!15, text width=1.2cm,font=\tiny},
    xnode/.style = {basic, thin, rounded corners=2pt, align=center, fill=pink!40,text width=1.5cm,font=\tiny},
    tnode/.style = {basic, thin, align=left, fill=green!30!yellow!20, text width=7cm, align=center, font=\tiny},
    % wnode/.style = {basic, thin, align=left, fill=pink!10!blue!80!red!10, text width=6.5em},
    edge from parent/.style={draw=black, edge from parent fork right}
}

\begin{forest} for tree={
    grow=east,
    growth parent anchor=west,
    parent anchor=east,
    child anchor=west,
    edge path={\noexpand\path[\forestoption{edge},->, >={latex}] 
         (!u.parent anchor) -- +(6pt,0pt) |-  (.child anchor) 
         \forestoption{edge label};}
}
% l sep is used for arrow distance
[Heterogeneous CL, basic,  l sep=5mm
    [Task \\ Heterogeneity, onode,  l sep=5mm,
        [Downstream Tasks, onode,  l sep=5mm,
            [Connecting Downstream Tasks with CL Strategies, xnode,  l sep=5mm,
                [Task Reformulation, xnode,  l sep=5mm,
        		[CRLC \cite{do2021clustering}\text{,} CL4SRec \cite{xie2022contrastive}\text{,} CL4SRec \cite{xie2022contrastive}\text{,} InfoNCE \cite{oord2018representation}\text{,} NeuTraL \cite{qiu2021neural}\text{,} MSC \cite{reiss2023mean}, tnode]]
                [Multi-task Learning, xnode,  l sep=5mm,
        		[CoCa \cite{yu2022coca}\text{,} XMC-GAN \cite{DBLP:conf/cvpr/0010KBLY21}\text{,} mRASP2 \cite{DBLP:conf/acl/PanWWL20}\text{,} SimCTG \cite{su2022contrastive}\text{,} SimGCL \cite{yu2022graph}\text{,} STERLING \cite{jing2023sterling}, tnode]]
                [Prompting Learning, xnode,  l sep=5mm,
        		[PCE~\cite{DBLP:conf/acl/ParanjapeMGHZ21}\text{,} AKSCP~\cite{ DBLP:conf/acl/ZhengSYLPZXZ23}\text{,} PESCO~\cite{DBLP:conf/acl/WangCZY23}\text{,} CP-Tuning~\cite{DBLP:conf/wsdm/Xu0QLXHH23}\text{,} LM-CPPF~\cite{DBLP:conf/acl/AbaskohiRY23}, tnode]]
                [AutoML, xnode,  l sep=5mm,
        		[JOAO \cite{DBLP:conf/icml/YouCSW21}\text{,} InfoTS \cite{luo2023time}\text{,} AutoSSL \cite{jin2021automated}\text{,} InfoMin \cite{DBLP:conf/nips/Tian0PKSI20}\text{,} AutoCL \cite{jing2024automated}\text{,} ArieL~\cite{feng2022adversarial}, tnode]]
            ]
            [Typical Downstream Tasks, xnode,  l sep=5mm,
                [Time-Series, xnode,  l sep=5mm,
        		[TNC~\cite{tonekaboni2020unsupervised}\text{,} TRNN~\cite{jing2021network}\text{,} CPD~\cite{deldari2021time}\text{,} SR-CNN~\cite{ren2019time}\text{,} CIB~\cite{choi2023conditional}\text{,} CoST~\cite{woo2021cost}, tnode]]
        	[Graph Learning, xnode,  l sep=5mm,
        		[Hdmi~\cite{jing2021hdmi}\text{,} DIM~\cite{velickovic2019deep}\text{,} CL4SRec~\cite{xie2022contrastive}\text{,} DINGAL~\cite{yan2021dynamic}\text{,} CoLA~\cite{liu2021anomaly}\text{,} Gccad~\cite{chen2022gccad}\text{,} GMI~\cite{peng2020graph}, tnode]]
                [NLP, xnode,  l sep=5mm,
        			[mRASP2~\cite{pan2021contrastive}\text{,}  ssSCL-ST~\cite{wang2021cross}\text{,}    CALMS~\cite{DBLP:conf/acl/WangCZQL21}\text{,}  CGT~\cite{ye2021contrastive}\text{,}  SeqCo~\cite{xu2022sequence}\text{,}  CLSC~\cite{wang2020coarse}\text{,}  Contrastnet~\cite{chen2022contrastnet}, tnode]]
                [CV, xnode,  l sep=5mm, 
        			[Mice~\cite{tsai2020mice}\text{,} PCL~\cite{li2020prototypical}\text{,} CC~\cite{DBLP:conf/aaai/Li0LPZ021}\text{,} CAT-Det~\cite{DBLP:conf/cvpr/ZhangC022}\text{,} DiscoFaceGAN~\cite{DBLP:conf/cvpr/DengYCWT20}\text{,}  Contraga~\cite{kang2020contragan}\text{,} XMC-GAN~\cite{DBLP:conf/cvpr/0010KBLY21} , tnode]]
            ]
        ]
        [Pre-training Tasks, onode,  l sep=5mm,
            [Auxiliary Tasks, xnode,  l sep=5mm,
        	   [CLIP \cite{DBLP:conf/icml/RadfordKHRGASAM21}\text{,} TimesURL \cite{liu2023timesurl}\text{,} MUSER \cite{chen2022learning}\text{,} KGCL \cite{yang2022knowledge}\text{,} GeoCL \cite{ayush2021geography}\text{,} Knowledge-CLIP \cite{pan2022contrastive}, tnode]]	
            [Preference Tasks, xnode,  l sep=5mm,
        	   [CPL~\cite{DBLP:journals/corr/abs-2310-13639}\text{,} CLEA~\cite{dennler2024using}\text{,} CL-RLHF~\cite{shen2024improving} , tnode]]
            [Supervised Tasks, xnode,  l sep=5mm,
                [SupCon \cite{abs-2004-11362}\text{,} Sel-CL \cite{li2022selective}\text{,} FSCL \cite{park2022fair}\text{,} UniCL \cite{yang2022unified}\text{,} HeroCon~\cite{DBLP:conf/kdd/ZhengXZH22}, tnode]]
            [Pretext Tasks, xnode,  l sep=5mm,
        	[Model Based Pair Construction, xnode,  l sep=5mm,
        		[ProtoNCE \cite{li2020prototypical}\text{,} InfoMin \cite{DBLP:conf/nips/Tian0PKSI20}\text{,} AutoGCL \cite{yin2022autogcl}\text{,} InfoTS \cite{luo2023time}\text{,} FairCL \cite{zhang2022fairness} , tnode]]
                [Heuristics Based Pair Construction, xnode,  l sep=5mm, 
                    [DIM \cite{DBLP:conf/iclr/HjelmFLGBTB19}\text{,} SimCLR \cite{chen2020simple}\text{,} DeepWalk \cite{perozzi2014deepwalk}\text{,} CLIP \cite{DBLP:conf/icml/RadfordKHRGASAM21}\text{,} SRL \cite{franceschi2019unsupervised}, tnode]]
            ]
        ]
    ]
    [View \\ Heterogeneity, onode,  l sep=5mm,
        [Contrastive Foundation Model, onode,  l sep=5mm,
            [Attempts Towards Other Foundation Model, xnode,  l sep=5mm,
                [TimeCLR \cite{yeh2023toward}\text{,} MA-GCL ~\cite{DBLP:conf/aaai/Gong0S23}\text{,} SimGRACE ~\cite{DBLP:conf/www/XiaWCHL22}\text{,} GraphCL ~\cite{DBLP:conf/nips/YouCSCWS20}, tnode]]
            [Multi-modal Foundation Model, xnode,  l sep=5mm,
                [Graph-lanuage, xnode,  l sep=5mm,
                    [ConGraT ~\cite{DBLP:journals/corr/abs-2305-14321}\text{,} G2P2 ~\cite{DBLP:conf/sigir/Wen023}\text{,} GRENADE ~\cite{DBLP:conf/emnlp/0001DL23}\text{,} MolFM ~\cite{DBLP:journals/corr/abs-2307-09484}\text{,} MolCA ~\cite{DBLP:conf/emnlp/LiuLL00K0C23}\text{,} GIT-Mol~\cite{DBLP:journals/corr/abs-2308-06911}\text{,} MoMu ~\cite{DBLP:journals/corr/abs-2209-05481}, tnode]]
                [Audio-language, xnode,  l sep=5mm,
                    [CLAP~\cite{DBLP:conf/icassp/ElizaldeDIW23}\text{,} C-MCR ~\cite{DBLP:conf/nips/WangZCHLYTLWZZ23}\text{,} CALM ~\cite{DBLP:journals/corr/abs-2202-03587}\text{,} Wav2CLIP ~\cite{DBLP:conf/icassp/WuSKB22}\text{,}  AudioCLIP ~\cite{DBLP:conf/icassp/GuzhovRHD22}\text{,} LAION CLAP ~\cite{DBLP:conf/icassp/WuCZHBD23}\text{,} CLAPSpeech ~\cite{DBLP:conf/acl/YeHRJLHYZ23}, tnode]]
                [Vision-language, xnode,  l sep=5mm,
                    [CLIP~\cite{DBLP:conf/icml/RadfordKHRGASAM21}\text{,}  KELIP ~\cite{DBLP:journals/corr/abs-2203-14463}\text{,} ChineseCLIP ~\cite{DBLP:journals/corr/abs-2211-01335}\text{,} AltCLIP ~\cite{DBLP:conf/acl/ChenLZYW23}\text{,} UniCLIP ~\cite{DBLP:conf/nips/LeeKSKKLK22}\text{,} SLIP ~\cite{DBLP:conf/nips/LeeKSKKLK22}\text{,} RA-CLIP ~\cite{DBLP:conf/cvpr/XieSXZZZ23}\text{,} LA-CLIP ~\cite{DBLP:conf/nips/FanKIKT23}\text{,} GrowCLIP ~\cite{DBLP:conf/iccv/DengSHLXHKZZL23} ,  tnode]]
            ]
            [Large Language Model, xnode,  l sep=5mm,
                [SimCSE ~\cite{gao2021simcse}\text{,} ContraCLM ~\cite{DBLP:conf/acl/JainZAWNLTNRBMX23}\text{,} CLEAR ~\cite{DBLP:journals/corr/abs-2012-15466}\text{,}  MixText ~\cite{DBLP:conf/acl/ChenYY20}\text{,} COCO-LM ~\cite{DBLP:conf/nips/MengXBTBHS21}\text{,} NL-Augmenter ~\cite{DBLP:journals/corr/abs-2112-02721}\text{,} EfficientCL ~\cite{DBLP:conf/emnlp/YeKO21}\text{,} DiffAug \cite{DBLP:conf/emnlp/WangL22}\text{,} DeCLUTR ~\cite{DBLP:conf/acl/GiorgiNWB20}, tnode]]
            [Large Vision Model, xnode,  l sep=5mm,
                [SimCLR~\cite{chen2020simple}\text{,} BYOL \cite{DBLP:conf/nips/GrillSATRBDPGAP20}\text{,} MoCo~\cite{he2020momentum}\text{,} MoCo v2~\cite{DBLP:journals/corr/abs-2003-04297}\text{,} InfoMin~\cite{DBLP:conf/nips/Tian0PKSI20}\text{,} NNCLR~\cite{DBLP:conf/iccv/DwibediATSZ21},  tnode]]
        ]
        [Basis, onode,  l sep=5mm,
         %    [Time-series Model, xnode,  l sep=5mm,
        	% 	[Eldele et al. , tnode]]
         %    [Graph Model, xnode,  l sep=5mm,
        	% 	[Hassani and Ahmadi, tnode]]
        	% [Text Model, xnode,  l sep=5mm,
        	% 	[Lin et al., tnode]]
         %    [Vision Model, xnode,  l sep=5mm,
        	% 	[Lin et al., tnode]]
            [Uni-modal Model, xnode,  l sep=5mm,
        		[GCA~\cite{DBLP:conf/www/0001XYLWW21}\text{,} GraphCL~\cite{DBLP:conf/nips/YouCSCWS20}\text{,} JOAO~\cite{DBLP:conf/icml/YouCSW21}\text{,} VaSCL \cite{DBLP:conf/acl/ZhangXZMA22}\text{,} CC~\cite{DBLP:conf/cvpr/TrostenLJK23}\text{,} ClEAR~\cite{DBLP:journals/corr/abs-2012-15466}, tnode]]
            [Multi-modal Model, xnode,  l sep=5mm,
        		[C-MCR~\cite{DBLP:conf/nips/WangZCHLYTLWZZ23}\text{,} M5Product~\cite{DBLP:conf/cvpr/DongZWWKWLWL22}\text{,} Mulan~\cite{zheng2024multi}\text{,} FairMVC~\cite{zheng2023fairness}\text{,} HeroCon~\cite{DBLP:conf/kdd/ZhengXZH22}\text{,} MFLVC~\cite{DBLP:conf/cvpr/XuT0P0022}\text{,} COMPLETER~\cite{DBLP:conf/cvpr/0001GLL0021}\text{,} FactorCL~\cite{ DBLP:conf/nips/LiangDMZMS23}, tnode]]
        ]  
    ]
]
\end{forest}
    \caption{Taxonomy of this Survey with Representative Works. 
    }
    \vspace{-3mm}
    \label{fig:lit_surv}
\end{figure*}
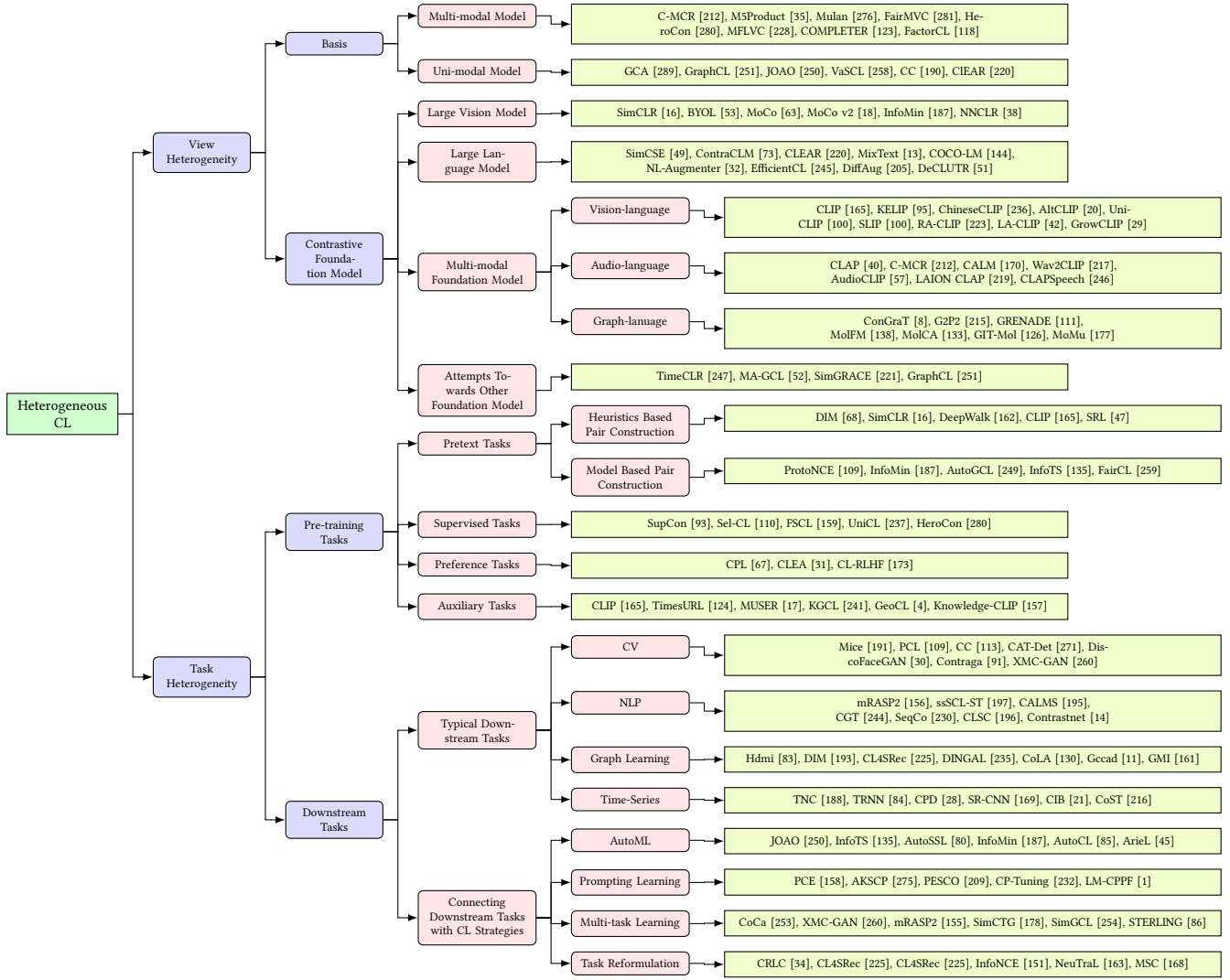
\section{Basic Concept of Contrastive Learning}
% \hh{1. shall we call these content (before sec2.1) as sth like 'basic concept of contrastive learning'? 2. consider to add a notation table to list main symbols and their meanings used in this paper. this helps the readability.} \lc{I have added a table to summarize the main symbols.}
% % \by{A full CL strategy is comprised of (1) data augmentation; (2) embedding augmentation; (3) contrastive pair construction; (4) loss functions. We may also briefly consider to mention them here. After pre-training, (5) the quality of CL strategies is usually evaluated by downstream tasks.}\lc{I have added these information as the basic concept of CL.}
% \by{We need to connect the concepts introduced here with the sections in the survey. maybe we need to modify the components of CL a little bit.}
% \lc{Updated.}

Contrastive learning (CL) aims at learning the compact representation by contrasting the embeddings with one negative sample, following the idea of Noise Contrastive Estimation (NCE) \cite{DBLP:journals/jmlr/GutmannH10}. Its pipeline is comprised of three stages, including augmentation, contrastive pair construction, and loss function formulation. In the first stage, many existing works either augment the raw data (\ie, data augmentation) or the embedding (\ie, embedding augmentation) to get an augmented sample and enrich the negative sets. In the second stage, researchers design how to construct the positive and negative pairs based on the different purposes or different settings (\eg, unsupervised contrastive loss~\cite{chen2020simple} vs supervised contrastive loss~\cite{abs-2004-11362}). The most commonly used contrastive pair construction considers that two samples augmented from the same raw data can form a positive pair and the rest of the samples are treated as negative samples. In the third stage, various types of contrastive learning losses are formulated based on different contrastive pair constructions, \eg, instance-level contrastive loss~\cite{DBLP:conf/nips/ZhangR22, DBLP:conf/iclr/Xiao0ED21}, cluster-level contrastive loss~\cite{DBLP:conf/aaai/Li0LPZ021, DBLP:conf/mm/WangFLG23}, contrastive alignment~\cite{DBLP:conf/cvpr/TrostenLJK23, DBLP:conf/cvpr/TrostenLJK21}, inter-view contrastive loss~\cite{DBLP:conf/cvpr/XuT0P0022, DBLP:conf/cvpr/0001GLL0021}, etc. A detailed discussion of the various types of loss formulation is shown in the next several subsections.

In addition to these stages, many existing works~\cite{perozzi2014deepwalk, DBLP:conf/icml/RadfordKHRGASAM21, yang2022unified} design a variety of CL strategies for pre-training tasks and downstream tasks. During pre-training, various characteristics of the data are injected into the models by pre-training tasks, including pretext tasks \cite{DBLP:conf/iclr/HjelmFLGBTB19, chen2020simple, franceschi2019unsupervised}, supervised tasks~\cite{abs-2004-11362, park2022fair}, preference tasks \cite{DBLP:journals/corr/abs-2310-13639, dennler2024using} and auxiliary tasks~ \cite{pan2022contrastive,yang2022knowledge}.
After pre-training, the models are fine-tuned to learn task-specific patterns of the downstream tasks, including automated machine learning~\cite{jin2021automated, suresh2021adversarial,reed2021selfaugment}, prompt learning~\cite{DBLP:conf/wsdm/Xu0QLXHH23, DBLP:conf/acl/AbaskohiRY23, DBLP:conf/acl/WangCZY23}, multi-task learning~\cite{DBLP:conf/acl/PanWWL20, yu2022graph}, task reformulation~\cite{DBLP:conf/icml/RadfordKHRGASAM21, qiu2021neural, DBLP:conf/aaai/Li0LPZ021}, etc.

% CL aims at learning the compact representation by comparing the embeddings with a Noise Contrastive Estimation (NCE) \cite{DBLP:journals/jmlr/GutmannH10} function defined by:
% \begin{equation}
%     \begin{split}
%          \mathcal{L} = -\log \frac{\exp(sim(\bm{z}, \bm{z^+}))}{\exp(sim(\bm{z}, \bm{z}^+)) + \exp(sim(\bm{z}, \bm{z}^-))}
%     \end{split}
% \end{equation}
% where $\bm{z}$ is the learned representation of a input sample $\bm{x}$, $\bm{z}^+$ is the representation of a positive sample similar to $\bm{x}$ and $\bm{z}^-$ is the representation of a negative sample dissimilar to $\bm{x}$. 

\begin{table}[t]
\caption{Main symbols and notation}
\vspace{-2mm}
\label{notation_table}
\begin{center}
\scalebox{0.95}{
\begin{tabular}{|c|c|}
\hline
$\mathcal{D}$ & The training dataset \\ \hline
$\bm{x}$  &  The input feature \\  \hline
% $v$  &  The total number of views \\  \hline
% $\bm{z}_i^j$  &  The representation of the $j$-th view of the $i$-th sample\\  \hline
$\bm{z}_i^+$ ($\bm{z}_i^-$)  &  A positive (negative) sample for $\bm{z}_i$\\  \hline
$sim(\cdot)$  &  The similarity measurement function \\  \hline
$\tau$  &  The temperature to scale the similarity measurement \\  \hline
% \multirow{2}{*}   &  The representation of the image (text) view \\
%  $\bm{z}_i^I$ ($\bm{z}_i^T$)   & for the $i$-th sample\\ \hline
% $\bm{z}_i^I$ ($\bm{z}_i^T$)  &  Image (text) view representation of sample $i$ \\  \hline
\end{tabular}
}
\end{center}
\vspace{-3mm}
\end{table}

Formally, CL loss follows the idea of NCE loss~\cite{DBLP:journals/jmlr/GutmannH10} by including more negative samples as follows:
\begin{equation}
    \begin{split}
         \mathcal{L} = -\E_{x_i\in \mathcal{D}} \log \frac{\exp(sim(\bm{z}_i, \bm{z}_i^+))}{\exp(sim(\bm{z}_i, \bm{z}_i^+)) + \sum_{k\neq i} \exp(sim(\bm{z}_i, \bm{z}_k^-))}
    \end{split}
\end{equation}
where $\bm{z}_i$ is the learned representation of a input sample $\bm{x}_i$ from the dataset $\mathcal{D}$, $\bm{z}_i^+$ is the representation of a positive sample similar to $\bm{x}_i$ and $\bm{z}_k^-$ is the representation of a negative sample dissimilar to $\bm{x}_i$. $sim(\cdot)$ denotes the similarity measurement functions, (\eg, $sim(\bm{a}, \bm{b})={(\bm{a})}^T \bm{b}/\tau$, where $\tau$ is the temperature). It first constructs a dataset $\mathcal{D}$ with $n$ samples containing $1$ positive samples and $n-1$ negative samples and then maximizes the similarities between $\bm{z}_i$ and $\bm{z}_i^+$.  Table~\ref{notation_table} summarizes the symbols and their meanings.

% \hh{which two types?}\lc{updated.}
% \by{in addition to NCE and InfoNCE, there are many other types of losses: including InfoMax based loss, triplet loss, negative sampling loss etc., we should briefly mention them here.}
There are many other types of losses similar to CL loss, including InfoMax Based loss~\cite{DBLP:conf/iclr/HjelmFLGBTB19}, triplet loss~\cite{DBLP:conf/cvpr/SchroffKP15}, etc. Specifically, ~\cite{DBLP:conf/iclr/HjelmFLGBTB19} proposes to maximize the mutual information between the input feature $\bm{x}$ and the output of the encoder $\bm{z}$ (\ie, $\max \mathcal{I}(\bm{x}_i, \bm{z}_i)$); ~\cite{DBLP:conf/cvpr/SchroffKP15} introduces the triplet loss to compare the representation of the anchor sample $\bm{x}_i$ with the positive and negative samples as follows:
\vspace{-1mm}
\begin{equation}
    \begin{split}
        \mathcal{L} = \sum_i||\bm{z}_i - \bm{z}^+_i||_2^2 - ||\bm{z}_i - \bm{z}^-_i||_2^2 + \alpha
    \end{split}
\end{equation}
\vspace{-3mm}
where $\alpha$ is a margin enforced between positive and negative pairs.

% \hh{my general impression of sections 2 and 3: they read a bit plain. we can consider to add a figure/table (e.g., to compare or summarize key features of different methods) to each section or subsection, and/or a summary paragraph (to summarize pros and cons). This will bring some structure to each (sub-)section, and makes the content less like a laundry list (a common issue with survey paper)} \lc{I plan to summarize each subsection in Figure~\ref{taxonomy}.}

\section{Contrastive Learning For View Heterogeneity}
% In this section, we first provide the audience with the traditional multi-view CL model as the basis and then introduce the multi-view CL for large foundation models. 
In this section, we first present the basis of CL for view heterogeneity and then introduce traditional multi-view CL methods in different domains, including computer vision~\cite{DBLP:conf/nips/WangZCHLYTLWZZ23, DBLP:journals/corr/abs-2105-09401, DBLP:conf/cvpr/XuT0P0022, DBLP:conf/cvpr/0001GLL0021, DBLP:conf/nips/LiangDMZMS23}, natural language processing~\cite{DBLP:conf/emnlp/Lin022, DBLP:conf/acl/ZhangXZMA22, DBLP:conf/emnlp/Zhang0MCTN022, DBLP:conf/acl/OuyangY023, DBLP:journals/corr/abs-2309-08929, DBLP:conf/naacl/ZhangMAHK22}, etc. Based on these traditional CL methods for view heterogeneity, we show how the researchers apply contrastive self-supervised learning to train the multi-view foundation models.

\subsection{Basis of Contrastive Learning for View Heterogeneity}
View heterogeneity refers to situations where data from different sources are available for training a model~\cite{DBLP:conf/www/ZhengCYCH21, DBLP:conf/sdm/ZhengCH19, DBLP:conf/www/ZhouZZLH20}. In CL, view heterogeneity can be categorized into two scenarios. In the first scenario, the raw data is unimodal or single-view (\eg, single-view image, text, or graph data), while in the second scenario, the raw data are collected from multiple data sources and the dataset naturally consists of multiple views (\eg, images with text descriptions in the social media). Different from InfoNCE~\cite{oord2018representation} maximizing the input data $\bm{x}$ and its contextual information, CL for view heterogeneity aims to maximize the mutual information of multiple views of the same sample to extract the shared representations~\cite{tian2020contrastive}. At the early stage, most CL methods tend to first use data augmentation methods to generate the augmented view and then apply CL~\cite{DBLP:conf/nips/ZhangR22, DBLP:conf/iclr/Xiao0ED21, DBLP:conf/cvpr/TrostenLJK23, DBLP:conf/ijcai/LinBB0ZX22, DBLP:conf/nips/PanK21, DBLP:conf/nips/YouCSCWS20, DBLP:conf/icml/YouCSW21, DBLP:conf/icml/HassaniA20}. Different from these methods, CMC~\cite{tian2020contrastive} formally applies the idea of CL to handle the raw data with multiple views. Following CMC, various types of CL losses are proposed to model the multi-modal data, including inter-modality contrastive loss~\cite{DBLP:conf/nips/WangZCHLYTLWZZ23, DBLP:conf/cvpr/DongZWWKWLWL22, zheng2024multi, zheng2023fairness, DBLP:journals/corr/abs-2105-09401, DBLP:conf/kdd/ZhengXZH22, DBLP:conf/cvpr/XuT0P0022, DBLP:conf/cvpr/0001GLL0021, DBLP:conf/nips/LiangDMZMS23, DBLP:conf/ijcai/LinBB0ZX22, DBLP:conf/mm/WangFLG23, DBLP:conf/emnlp/Lin022, DBLP:conf/naacl/ZhangMAHK22, DBLP:journals/corr/abs-2309-08929, DBLP:conf/naacl/ZhangMAHK22, DBLP:conf/emnlp/LiWFO23}, intra-modality contrastive loss~\cite{DBLP:conf/nips/WangZCHLYTLWZZ23,  DBLP:journals/corr/abs-2105-09401, DBLP:conf/kdd/ZhengXZH22}, contrastive alignment~\cite{DBLP:conf/cvpr/TrostenLJK23, DBLP:conf/cvpr/TrostenLJK21, DBLP:conf/acl/OuyangY023}.

% Specifically, in natural language processing domain, CL is applied to model instance-level representations~\cite{DBLP:conf/emnlp/Lin022}, sentence-level or paragraph-level representations~\cite{DBLP:conf/emnlp/Zhang0MCTN022, DBLP:journals/corr/abs-2309-08929, DBLP:conf/naacl/ZhangMAHK22}; 
% and further extends CL in the two-view setting to include the multiple-view setting. The objective function of CMC is formulated as follows:
% \begin{equation}
%     \label{CMC_core_view}
%     \begin{split}
%         \mathcal{L} &= -\sum_{1\leq j \leq l \leq v} \E_{x_i\in \mathcal{D}}[ \log \frac{\exp(sim(\bm{z}_i^j, \bm{z}_i^l))}{\sum_{k} \exp(sim(\bm{z}_i^j, \bm{z}_k^l))}] 
%     \end{split}
% \end{equation}
% % where $sim(\cdot)$ denotes the similarity measurement functions, (\eg, $sim(\bm{Z_i^1}, \bm{Z_i^2})=\frac{(\bm{Z_i^1)}^T \bm{Z_i^2}}{\tau}$ and $\tau$ is the temperature).
% where $\bm{z}_i^j$ denotes the representation of the $j$-th view of the $i$-th sample and $v$ is the number of views. Notice that CMC scales to any number of views and the authors show that the contrastive loss performs the task of cross-view prediction and the more views are involved in the learning task, the better representation it learns to capture underlying scene semantics. 
% \lc{For augmented view, ... for raw dataset with multiple views,....}

% \by{maybe this paragraph should emphasize "views" rather than "losses".}
% \lc{ (1) two augmented views, (2). global view vs local view, (3) }

Unlike multi-modal data which naturally consists of various types of data, when handling single-view data such as images, text, and graphs, researchers often rely on data augmentation techniques to generate one or more augmented views for CL \cite{DBLP:conf/nips/ZhangR22, DBLP:conf/iclr/Xiao0ED21, DBLP:conf/cvpr/TrostenLJK23, DBLP:conf/ijcai/LinBB0ZX22, DBLP:conf/nips/PanK21, DBLP:conf/nips/YouCSCWS20, DBLP:conf/icml/YouCSW21, DBLP:conf/icml/HassaniA20}. Here, we characterize these view construction methods into two main categories: global and local. Global view construction involves augmenting samples globally, creating synthetic data akin to the original, commonly done through methods like random rotation and color jittering in computer vision~\cite{DBLP:conf/iclr/Xiao0ED21, DBLP:conf/cvpr/TrostenLJK23, tian2020contrastive}, or graph-level augmentation in graph mining ~\cite{DBLP:conf/icml/HassaniA20, DBLP:conf/www/0001XYLWW21, DBLP:conf/ijcai/JinZL00P21}, jitter-and-scale and permutation-and-jitter strategies for time-series data~\cite{DBLP:conf/ijcai/Eldele0C000G21}. Conversely, local view construction focuses on augmenting samples locally or partially, often with specific purposes, such as image cropping in computer vision~\cite{DBLP:conf/iclr/Xiao0ED21, DBLP:conf/cvpr/TrostenLJK23, tian2020contrastive} or node-level augmentations and edge-level augmentations in graph mining~\cite{DBLP:conf/nips/YouCSCWS20, DBLP:conf/icml/YouCSW21}, sentence-level augmentation in NLP~\cite{DBLP:conf/acl/ZhangXZMA22, DBLP:journals/corr/abs-2012-15466}, etc. To name a few, TS-TCC~\cite{DBLP:conf/ijcai/Eldele0C000G21} first globally augments the raw data into two different but correlated views and then learns the temporal representation via the temporal contrastive module and the contextual contrasting module; TS2Vec~\cite{DBLP:conf/aaai/YueWDYHTX22} impose hierarchical CL regularization to constrain the locally augmented context views to share the consistent contextual information; DIM~\cite{DBLP:conf/iclr/HjelmFLGBTB19} proposes to maximize the mutual information between local representation and global representation. These approaches collectively enhance the diversity and richness of the data for effective CL.

\subsection{Contrastive Learning for Foundation Model with View Heterogeneity}
\subsubsection{Large Vision Model}
\label{Large_vision_model}
Different from many traditional CL methods, most of the large vision models~\cite{chen2020simple, he2020momentum, DBLP:journals/corr/abs-2003-04297, DBLP:conf/nips/GrillSATRBDPGAP20, DBLP:conf/iccv/CuiZ00J21, DBLP:conf/nips/Tian0PKSI20, DBLP:conf/cvpr/ColeYWAB22} mainly implement the data augmentation methods to generate two augmented views and then apply CL to learn the representations.
% Inspired by the early works of CL~\cite{oord2018representation},
Specifically, SimCLR~\cite{chen2020simple} achieves competitive performance on par with the supervised model after hundreds of iterations of fine-turning with 1\% of the labeled data on the ImageNet dataset. 
% Different from InfoNCE, SimCLR proposes to contrast the representations on the projection space rather than the original hidden space. 
% Specifically, given the representation $\bm{Z^1_i}$ ($\bm{Z^2_i}$) of the sample $\bm{X^1}_i$ ($\bm{X^2}_i$), SimCLR project $\bm{Z^1_i}$ ($\bm{Z^2_i}$) to the projection space denoted as $H^1_i=g_1(\bm{Z^1_i})$ ($H^2_i=g_2(\bm{Z^2_i})$), and then the CL loss is written as follows:
% \begin{equation}
%     \label{SimCLR}
%     \begin{split}
%          \mathcal{L} = -\E_{X_i\in \mathcal{D}}[ \log \frac{\exp(sim(\bm{H_i^1}, \bm{H_i^2}))}{\sum_{k} \exp(sim(\bm{H_i^1}, \bm{H_k^2}))}]
%     \end{split}
% \end{equation}
% After the training stage is completed, $\bm{Z^1_i}$ and $\bm{Z^2_i}$ will be used for the final prediction.
Different from traditional CL methods (e.g., InfoNCE~\cite{oord2018representation}), SimCLR introduces a learnable nonlinear transformation named \textit{projection head} between the representation and the contrastive loss, which alleviates the class collision issue to some extent~\cite{DBLP:conf/iccv/Zheng0Y0Z0021}. Furthermore, SimCLR contributes the success of the unsupervised CL large vision model to stronger data augmentation,  normalized embeddings, an appropriately adjusted temperature parameter, larger batch sizes, longer training iterations, and deeper and wider networks.
% the performance of CL with different commonly used data augmentation methods and the experimental results show that \textit{random color distortions} and \textit{random Gaussian blur} are the two most suitable augmentation methods in image classification task along with CL 
Despite the superiority of CL algorithms~\cite{chen2020simple, oord2018representation} for many tasks, one major drawback of CL is its \textit{high GPU memory requirement} as we need to increase the batch size to achieve better performance. To relax such a constraint, MoCo~\cite{he2020momentum} stores the new encoded representations of the current batch in a dictionary and removes the oldest representation to reduce memory usage but keep a large batch size. Additionally, as a slowly progressing encoder encodes the representations stored in the dictionary, MoCo adopts a momentum-based moving average to maintain consistency for the newest and oldest representations and ensure that the difference among these representations can be small. Chen \etal ~\cite{DBLP:journals/corr/abs-2003-04297} verify the effectiveness of the projection head and stronger data augmentation proposed in SimCLR and then incorporate them into MoCo to get MoCo v2, showing further performance improvement. Following these works, BYOL \cite{DBLP:conf/nips/GrillSATRBDPGAP20} trains the online network to predict the representation of the same image encoded by the target network under a different augmented view; Tian \etal~\cite{DBLP:conf/nips/Tian0PKSI20} investigate the question of "what makes for good views for contrastive learning" and propose \textit{InfoMin} principle to characterize the good views; Cui \etal~\cite{DBLP:conf/iccv/CuiZ00J21} tackle the difficulty of imbalanced learning of CL and introduce PaCo to rebalance the importance of the majority class and the minority classes; Dwibedi \etal~\cite{DBLP:conf/iccv/DwibediATSZ21} propose a nearest-neighbor CL method by sampling the nearest neighbors in the latent space and considering them as positives; Cole \etal~\cite{DBLP:conf/cvpr/ColeYWAB22} find that reducing the number of training data by half only degrades the performance by less than 2\% and current contrastive method may not be sufficient to generalize to many downstream tasks.

\subsubsection{Large Language Model}
% \hh{@Zihao, let's use present tense in this paragraph to be consistent with other parts of the paper} \zh{Fixed.}
CL with view heterogeneity is one of the most prevalent choices when pretraining large language models because of the scarcity of labeled data ~\cite{zhang-etal-2022-contrastive-data}, by regarding augmented data as new views. A well-known pretraining technique that learns word embeddings through CL is word2vec ~\cite{DBLP:journals/corr/abs-1301-3781}. Later, augmenting different views and conducting CL are used to pre-train state-of-the-art language models. SimCSE ~\cite{gao2021simcse} and ContraCLM ~\cite{DBLP:conf/acl/JainZAWNLTNRBMX23} adopt simple yet effective dropout-based augmentation. Shen \etal ~\cite{DBLP:journals/corr/abs-2009-13818} propose to apply a cutoff for natural language augmentation to boost the model ability on both language understanding and generation. CERT ~\cite{DBLP:journals/corr/abs-2005-12766} and MixText ~\cite{DBLP:conf/acl/ChenYY20} create augmentations of original sentences using back-translation. DeCLUTR ~\cite{DBLP:conf/acl/GiorgiNWB20}, which closely resembles QT ~\cite{DBLP:conf/iclr/LogeswaranL18}, samples textual segments of the anchors up to paragraph length, allowing each sample to be as overlapping view, adjacent view or subsumed view with the anchor. CoDA ~\cite{DBLP:conf/iclr/QuSSSC021} introduces contrast-enhanced and diversity-promoting data augmentation through a combination of back-translation, adversarial training, and label-preserving transformations. CLEAR ~\cite{DBLP:journals/corr/abs-2012-15466} augments sentences by word deletion, span deletion, reordering, and synonym substitution to generate different views. COCO-LM ~\cite{DBLP:conf/nips/MengXBTBHS21} incorporates sequence contrastive loss through sentence cropping. NL-Augmenter ~\cite{DBLP:journals/corr/abs-2112-02721} provides a framework with over a hundred transformations for the users to manually augment natural languages to generate views. Kaushik \etal ~\cite{DBLP:conf/iclr/KaushikHL20} design a human-in-loop system for counterfactually augmenting the documents. EfficientCL ~\cite{DBLP:conf/emnlp/YeKO21} applies a combination of cutoff and PCA jittering, similar to color jittering, for semantic information augmentation. DiffAug \cite{DBLP:conf/emnlp/WangL22} provides an option for sentence representation augmentation that is differentiable.

Besides pre-training, CL with view heterogeneity has been widely used in other tasks. (a) \textit{Fine-tuning}. Contrastive objectives have been used for language model fine-tuning \cite{DBLP:conf/iclr/GunelDCS21, DBLP:journals/corr/abs-2310-18339, DBLP:conf/acl/MaSA21}, and a recent work LM-CPPF ~\cite{DBLP:conf/acl/AbaskohiRY23} proposes to use few-shot paraphrasing for contrastive prompt-based fine-tuning. (b) \textit{Machine Translation}. Different languages are naturally different views. mRASP2 ~\cite{DBLP:conf/acl/PanWWL20} leverages CL with augmentations to align token representations and close the gap among representations of different languages. Li \etal ~\cite{DBLP:conf/acl/Li0CKV22} propose two-stage cross-lingual CL to improve word translation of language models. Mao \etal ~\cite{DBLP:conf/naacl/MaoCDSWK22} introduce a word-level contrastive objective to leverage word alignments for many-to-many neural machine translation. (c) \textit{Other view heterogeneity-related tasks}. CALMS ~\cite{DBLP:conf/acl/WangCZQL21} leverages contrastive sentence ranking and sentence-aligned substitution to conduct multilingual text summarization. Xu \etal ~\cite{DBLP:journals/corr/abs-2310-02263} introduce contrastive post-training techniques for aligning multiple language models of varying strengths. Lee \etal ~\cite{DBLP:conf/iclr/LeeLH21} propose to obtain positive and negative examples for conditional text generation by adding perturbations that optimize the conditional likelihood.

\subsubsection{Multi-modal Foundation Models}
Multi-modal foundation models combine data different modalities.

\textbf{Vision-language model.} The study of the vision-language models is evolving rapidly, and many surveys have provided comprehensive reviews from multiple perspectives. Mogadala \etal ~\cite{DBLP:journals/jair/MogadalaKK21} survey common vision-language tasks, benchmark datasets, and seminal methods. Li \etal ~\cite{DBLP:journals/corr/abs-2203-01922} first summarize the development of task-specific vision-language models, then review vision-language pretraining methods for general vision-language foundation models. Wang \etal ~\cite{DBLP:journals/ijautcomp/WangCQGWWTG23} and Du \etal ~\cite{DBLP:conf/ijcai/DuLLZ22} share recent advances in vision-language model pretraining. Zhang \etal ~\cite{DBLP:journals/corr/abs-2304-00685} review vision-language models specifically for various visual recognition tasks. In this work, we focus on leveraging heterogeneous CL in the pretraining phase of vision-language models.

The seminal work in this group of models that uses heterogeneous CL, specifically cross-modal CL, is the well-known CLIP  (Contrastive Language–Image Pre-training) ~\cite{DBLP:conf/icml/RadfordKHRGASAM21}. By optimizing the contrastive loss from the language and vision views of over 400 million image-caption pairs, CLIP achieves strong performance in few-shot or zero-shot image classification settings. 
% For each batch with a batch size $B$, CLIP minimizes a symmetrical image-text infoNCE loss $\mathcal{L} = \mathcal{L}_{I\rightarrow T} + \mathcal{L}_{T\rightarrow I}$, where $\mathcal{L}_{I\rightarrow T}$ 
% \by{maybe we don't need to show these two losses.}
% \begin{equation}
% \begin{split}
%     \mathcal{L}_{I\rightarrow T} &= -\frac{1}{B} \sum_{i=1}^B \log \frac{\exp(z_i^I \cdot z_i^T / \tau)}{\sum_{j=1}^B\exp(z_i^I \cdot z_j^T / \tau)}\\
%     \mathcal{L}_{T\rightarrow I} &= -\frac{1}{B} \sum_{i=1}^B \log \frac{\exp(z_i^T \cdot z_i^I / \tau)}{\sum_{j=1}^B\exp(z_i^T \cdot z_j^I / \tau)}
% \end{split}
% \end{equation}
% where $z_i^I$ and $z_i^T$ are respectively the embedding of sample $i$ from the image view and text view. $\tau$ is the temperature. 
Another seminal work ALIGN ~\cite{DBLP:conf/icml/JiaYXCPPLSLD21} uses the same heterogeneous contrastive backbone over a larger noisy dataset. Following the impressive success of the image-text CL framework, many CLIP variants have been proposed. KELIP ~\cite{DBLP:journals/corr/abs-2203-14463}, ChineseCLIP ~\cite{DBLP:journals/corr/abs-2211-01335} and AltCLIP ~\cite{DBLP:conf/acl/ChenLZYW23} extend CLIP into other languages by leveraging the heterogeneous CL to fine-tune the CLIP model. DeCLIP ~\cite{DBLP:conf/iclr/LiLZCOSYY22} considers self-supervision together with image-text-pair supervision to achieve data-efficient training. SLIP ~\cite{DBLP:conf/nips/LeeKSKKLK22} introduces image self-supervision to obtain better representations. UniCLIP ~\cite{DBLP:conf/nips/LeeKSKKLK22} further integrates the contrastive loss of both inter-domain pairs and intra-domain pairs into a single universal space. HiCLIP ~\cite{DBLP:conf/iclr/GengYT0Z23} incorporates hierarchy-aware attention with the heterogeneous contrastive pre-training. RA-CLIP ~\cite{DBLP:conf/cvpr/XieSXZZZ23} and LA-CLIP ~\cite{DBLP:conf/nips/FanKIKT23} respectively introduce retrieval-augmented and LLM-augmented heterogeneous CL between images and texts. GrowCLIP ~\cite{DBLP:conf/iccv/DengSHLXHKZZL23} extends the heterogeneous CL into online settings. Some other representative vision-language models also use heterogeneous CL during pretraining, for example, Flamingo ~\cite{DBLP:conf/nips/AlayracDLMBHLMM22}, FLAVA ~\cite{DBLP:conf/cvpr/SinghHGCGRK22}, WenLan ~\cite{DBLP:journals/corr/abs-2103-06561}, ALBEF ~\cite{DBLP:conf/nips/LiSGJXH21}.

\textbf{Audio-language model.} CLAP (Contrastive Language-Audio Pretraining) ~\cite{DBLP:conf/icassp/ElizaldeDIW23} imitates the process of CLIP to build an audio-language model that achieves state-of-the-art performance on multiple downstream tasks, even with much less training data compared to the vision-language domain. LAION CLAP ~\cite{DBLP:conf/icassp/WuCZHBD23} uses more data, as well as feature fusion and keyword-to-caption augmentation, to train the contrastive language-audio model. AudioCLIP ~\cite{DBLP:conf/icassp/GuzhovRHD22} adds the audio modality to the two-modality CLIP through three two-view CL. Wav2CLIP ~\cite{DBLP:conf/icassp/WuSKB22} learns robust audio representations by projecting audio into a shared embedding space with images and text and distilling from CLIP through contrastive loss projection layers. C-MCR ~\cite{DBLP:conf/nips/WangZCHLYTLWZZ23} offers a framework to efficiently train CLIP and CLAP by connecting the representation spaces. CALM ~\cite{DBLP:journals/corr/abs-2202-03587} can efficiently bootstrap high-quality audio embedding by aligning audio representations to pretrained language representations and utilizing contrastive information between acoustic inputs. CLAPSpeech ~\cite{DBLP:conf/acl/YeHRJLHYZ23} learns the prosody variance of the same text token under different contexts through word-level and phoneme-level contrastive pretraining for view heterogeneity.

\textbf{Graph-language model.} On text-attributed graphs, ConGraT ~\cite{DBLP:journals/corr/abs-2305-14321} conducts CLIP-like contrastive pretraining for both language and graph tasks. G2P2 ~\cite{DBLP:conf/sigir/Wen023} extends the CLIP framework to text attributed graphs for zero-shot and few-shot classification. GRENADE ~\cite{DBLP:conf/emnlp/0001DL23} conducts graph-centric CL and knowledge alignment that considers neighborhood-level similarity to learn
expressive and generalized representations. For text-paired graphs ~\cite{DBLP:journals/corr/abs-2312-02783}, MoleculeSTM ~\cite{DBLP:journals/natmi/LiuNWLQLTXA23} and MoMu ~\cite{DBLP:journals/corr/abs-2209-05481}
bridge molecular graphs and text data through contrastive
learning. MolCA ~\cite{DBLP:conf/emnlp/LiuLL00K0C23} adopts a cross-modal projector and uni-modal adapter to practically and efficiently understand molecular contents in both text and graph form. MolFM ~\cite{DBLP:journals/corr/abs-2307-09484} leverages information from the input molecule structure, the input text description, and the auxiliary knowledge graph to build a multimodal molecular foundation model. GIT-Mol~\cite{DBLP:journals/corr/abs-2308-06911} incorporates cross-modal CL to build a multi-modal molecular foundation model with graphs, SMILES (Simplified Molecular Input
Line Entry System), images, and text.

\subsubsection{Other Foundation Models}
% (1){Graph Foundation Models}.
% In this subsection, we briefly review foundation models for other data types due to the limited number of existing works on this topic. 
Inspired by foundation models for language and vision data, recently, some attempts have been made to build foundation models for other data types.
For time series, TimeCLR \cite{yeh2023toward} develops a time series foundation model by leveraging CL to train unlabeled samples from multiple domains. For the graph data, due to the complex nature of graphs, we only find some initial attempts~\cite{DBLP:conf/nips/YouCSCWS20, DBLP:conf/wsdm/YouCWS22, DBLP:journals/nn/LiangDZM0G23, DBLP:journals/tetci/ChenWFML24, zheng2021deeper} to develop graph foundation models and most of them follow the pertaining strategies of large language models~\cite{DBLP:journals/corr/abs-2310-11829, DBLP:journals/corr/abs-2402-02216}. For instance, GraphCL ~\cite{DBLP:conf/nips/YouCSCWS20} studies several intuitive augmentation strategies and proposes the initial framework for graph CL by maximizing the agreement of the augmented graphs in different views. Some later works ~\cite{DBLP:conf/wsdm/YouCWS22, DBLP:conf/bic-ta/SunCD022, DBLP:journals/nn/LiangDZM0G23, DBLP:journals/tetci/ChenWFML24} follow this track and propose other kinds of augmentations. 
% GRACE ~\cite{DBLP:journals/corr/abs-2006-04131} generates heterogeneous views on both structure and attribute levels through graph corruptions. GCC ~\cite{DBLP:conf/kdd/QiuCDZYDWT20} leverages CL with heterogeneous views to pretrain a GNN that learns the intrinsic and transferable structural representations. 
MA-GCL ~\cite{DBLP:conf/aaai/Gong0S23} and SimGRACE ~\cite{DBLP:conf/www/XiaWCHL22} propose that the heterogeneous views can also be generated from the neural architecture instead of the graph instances. 
\vspace{-1mm}
\section{Contrastive Learning for Task Heterogeneity}
\label{task_heterogeneity}
An ultimate goal of CL is to train foundation models to extract useful representations without human annotations.
The foundation models are usually trained through a pre-training and fine-tuning paradigm. 
% There are two types of tasks involved in CL: pre-training tasks and downstream tasks.
During pre-training, various characteristics of the data are injected into the models by pre-training tasks.
After pre-training, the models are fine-tuned to learn task-specific patterns of the downstream tasks.
In this section, we discuss the heterogeneous pre-training tasks and downstream tasks of CL.

\subsection{Pre-training Tasks}
Different pre-training tasks can guide models to capture different aspects of the data.
In general, there are four types of pre-training tasks, including \emph{pretext tasks} \cite{DBLP:conf/iclr/HjelmFLGBTB19,chen2020simple,perozzi2014deepwalk,DBLP:conf/icml/RadfordKHRGASAM21,franceschi2019unsupervised, DBLP:conf/cikm/ZhouZF0H22}, \emph{supervised tasks} \cite{abs-2004-11362,park2022fair,yang2022unified}, \emph{preference tasks} \cite{DBLP:journals/corr/abs-2310-13639, dennler2024using, shen2024improving, DBLP:conf/cikm/DuanFZLW23} and \emph{auxiliary tasks} \cite{pan2022contrastive,yang2022knowledge,ayush2021geography,chen2022learning,liu2023timesurl}.

\subsubsection{Pretext Tasks}
% The major advantage of pretext tasks is that they do not require the expansive and time-consuming human labeling.
The pretext tasks are the pre-training tasks without expensive and time-consuming human labels, and their objective is to discriminate positive and negative instance pairs, which are determined either by \emph{heuristics} \cite{DBLP:conf/iclr/HjelmFLGBTB19,chen2020simple,perozzi2014deepwalk,DBLP:conf/icml/RadfordKHRGASAM21,franceschi2019unsupervised} or \emph{extra models} \cite{li2020prototypical,jing2022x,DBLP:conf/nips/Tian0PKSI20,zhang2022fairness}.

\textbf{Heuristics Based Pair Construction}
% The naive pseudo labels are usually obtained by naive relations between two instances.
The heuristics-based methods construct contrastive pairs based on either simple relationships between instances or heuristically designed data augmentations.
For example, DIM \cite{DBLP:conf/iclr/HjelmFLGBTB19} treats a pair of local and global embeddings from the same image as positive and treats local and global embeddings from different images as negative.
SimCLR \cite{chen2020simple} treats two different augmented views of an image as a positive pair and treats two randomly sampled images as a negative pair.
% BiNE \cite{gao2018bine} treats the connected (user, item) pairs of the bipartite graphs as positive and treats randomly sampled pairs as negative.
DeepWalk \cite{perozzi2014deepwalk} leverages random walks to determine positive node pairs in a graph.
CLIP \cite{DBLP:conf/icml/RadfordKHRGASAM21} treats ground-truth image and text pairs as positive and other image and text pairs as negative.
SRL \cite{franceschi2019unsupervised} regards two adjacent sub-sequences in the same time series as a positive pair and two sub-sequences from different time series as a negative pair.

\textbf{Model Based Pair Construction}
Model-based methods leverage extra models, e.g., \emph{clustering}, \emph{view generation} and \emph{image editing} models, to generate contrastive pairs.
% The negative sampling process of CL inevitably include semantic errors \cite{li2020prototypical}, to address this issue, clustering is used to produce cluster pseudo labels.
For \emph{clustering},
ProtoNCE \cite{li2020prototypical} leverages external clustering methods, e.g., K-Means, to obtain semantic clusters, and uses the cluster centers to reduce the semantic errors of random negative sampling.
X-GOAL \cite{jing2022x} extends ProtoNCE to graphs.
% Sensitive attributes, such as gender and color, exist in real-world datasets.
% Fairness pretext tasks aim to exclude the information of the sensitive attributes in embeddings.
For \emph{view generation}, InfoMin \cite{DBLP:conf/nips/Tian0PKSI20} leverages flow-based models \cite{dinh2016density} to generate augmented views for an input image, and treats these generated views as positive pairs.
AutoGCL \cite{yin2022autogcl} and InfoTS \cite{luo2023time} extend InfoMin to graphs and time series.
For \emph{image editing},
FairCL \cite{zhang2022fairness} trains an image editor \cite{he2019attgan} to generate images with different sensitive labels, e.g., gender. 
The images generated from the same input image but having different sensitive labels are regarded as positive pairs.

\subsubsection{Supervised Tasks}
The data for supervised pre-training tasks is manually labeled before pre-training the models, which incorporates human knowledge.
SupCon \cite{abs-2004-11362} proposes to maximize the similarity of a pair of instances that share the same label.  
Sel-CL \cite{li2022selective} proposes to filter out noisy labels by selecting confident examples based on their representation similarity with their labels.
% FairCL \cite{zhang2022fairness} 
% Sensitive attributes, such as gender and color, exist in real-world datasets.
% Fairness pretext tasks aim to exclude the information of the sensitive attributes in embeddings.
FSCL \cite{park2022fair} introduces a Fair Supervised Contrastive Loss (FSCL) for visual representation learning based on SupCon, which defines the positive and negative pairs based on both class labels, e.g., attractiveness, and sensitive attribute labels, e.g., gender.
UniCL \cite{yang2022unified} unifies (image, label) and (image, text) pairs by expanding the label into a textual description, and then leverages image-to-text and text-to-image contrastive losses to pre-train the model. HeroCon~\cite{DBLP:conf/kdd/ZhengXZH22} proposes the weighted supervised contrastive loss to weight the importance of positive and negative pairs based on the similarity of different label vectors in the multi-label setting.

\subsubsection{Preference Tasks}
In recent years, Human-In-The-Loop (HITL) machine learning has become popular, which induces human prior knowledge into models by including humans in the training process. Different from supervised tasks, where humans first label the data and then the data is used to train the model, in HITL machine learning, humans iteratively evaluate the quality of the prediction made by the model and provide feedback to the model to adjust its learned knowledge. Specifically, ~\cite{DBLP:journals/corr/abs-2310-13639} derives contrastive preference loss for learning optimal behavior from human feedback using the regret-based model of human preferences. ~\cite{dennler2024using} proposes to combine CL loss to model exploratory actions and learn user preferences by utilizing the data collected from an interactive signal design process, where the data collection process can be regarded as the functionality of HITL. ~\cite{shen2024improving} introduces the contrastive rewards to penalize uncertainty and improve robustness based on human feedback. 
% ~\cite{DBLP:conf/prcv/BaiJLW23} proposes a preference CL loss to encode users' interests by contrasting preferences of user-items pairs.
% ~\cite{DBLP:conf/cikm/DuanFZLW23} designs a CL loss to model the long-term and short-term preferences and characterize the users' preferences. 

% \lc{I will add the related work soon.}
% preference-based learning. \url{https://icml.cc/virtual/2023/workshop/21495}

\subsubsection{Auxiliary Tasks}
The auxiliary tasks leverage external or metadata information to improve CL. 
For example,
Knowledge-CLIP \cite{pan2022contrastive} uses knowledge graph to guide CLIP \cite{DBLP:conf/icml/RadfordKHRGASAM21} to encode more precise semantics by tasks related to knowledge graphs e.g., link prediction.
KGCL \cite{yang2022knowledge} introduces a Knowledge Graph CL framework (KGCL) for the recommendation, which leverages knowledge graphs to provide side information of items via a knowledge-aware co-CL task.
GeoCL \cite{ayush2021geography} induces geo-location information into image embeddings by classifying geo-labels.
% MICoL \cite{zhang2022metadata} uses document metadata to build a meta-graph, and two documents are treated as a positive pair if they are reachable in the meta-graph. 
MUSER \cite{chen2022learning} uses text metadata, e.g., lyrics and album description, to learn better music sequence representations by aligning text tokens with music tokens as CLIP \cite{DBLP:conf/icml/RadfordKHRGASAM21}.
TimesURL \cite{liu2023timesurl} uses a reconstruction error to preserve important temporal variation information.
Additionally, other methods directly use downstream tasks as auxiliary tasks \cite{yu2022coca,DBLP:conf/cvpr/0010KBLY21,DBLP:conf/acl/PanWWL20,su2022contrastive,jing2023sterling}, which can also be regarded as \emph{multi-task learning} methods (see Sec. \ref{sec:connection}).

\subsection{Downstream Tasks}
\label{downstream_tasks}
The effectiveness of the CL methods is usually measured by their performance on a variety of downstream tasks.
In this subsection, we first briefly review representative downstream tasks and then discuss how to connect downstream tasks with CL strategies.

\subsubsection{Typical Downstream Tasks}
We briefly review the typical downstream tasks for different fields.

\textbf{Computer Vision.}
% \textit{Computer Vision}
Typical tasks include image classification~\cite{chen2020simple, he2020momentum, DBLP:journals/corr/abs-2003-04297, DBLP:conf/nips/GrillSATRBDPGAP20, DBLP:conf/iccv/CuiZ00J21, DBLP:conf/nips/Tian0PKSI20, DBLP:conf/cvpr/ColeYWAB22,rawat2017deep,zhou2021soda},
image clustering \cite{DBLP:conf/aaai/Li0LPZ021, tsai2020mice, li2020prototypical, wang2021dnb},
objective detection~\cite{DBLP:conf/cvpr/0019CZLCZ22, DBLP:conf/cvpr/ZhangC022, DBLP:conf/cvpr/SunLCYZ21, DBLP:conf/cvpr/BadamdorjR0C22,xie2021detco,wang2021dense}, image generation~\cite{DBLP:conf/cvpr/ZhanYWZLZ22, DBLP:conf/cvpr/DengYCWT20, DBLP:conf/cvpr/0010KBLY21,kang2020contragan}, style transfer~\cite{DBLP:conf/cvpr/HangXYL22, DBLP:conf/nips/ChenZWZZLXL21, DBLP:conf/iccv/YangHY23,park2020contrastive}, etc.

\textbf{Natural Language Processing.} Typical tasks include machine translation~\cite{DBLP:conf/emnlp/0003KLZKJ23, DBLP:conf/emnlp/IndurthiCAT23,  DBLP:conf/acl/OuyangY023, DBLP:conf/acl/Li0CKV22,pan2021contrastive,zhang2022frequency}, text classification~\cite{DBLP:conf/emnlp/Lin022, DBLP:conf/acl/YangYNGX23, DBLP:conf/acl/YuZQSWGWYN23,chen2022contrastnet,pan2022improved,choi2022c2l,wang2022contrastive,wang2021cross}, 
topic modeling \cite{li2022uctopic, nguyen2021contrastive,han2023unified,wang2020coarse,shi2021simple,vayansky2020review,jing2018cross},
text summarization~\cite{DBLP:conf/iconip/0003LZG023, DBLP:conf/ijcnn/ZhangSCXZL22, DBLP:conf/acl/WangCZQL21, DBLP:conf/cvpr/00040QBSW23,jing2021multiplex,xu2022sequence,liu2021simcls}, and information extraction~\cite{DBLP:conf/acl/GuoDCLZQ0YWP23, DBLP:conf/acl/LeeLZDPSZSGWALQ23, DBLP:conf/emnlp/LiZDM0Q23, ye2021contrastive}.

\textbf{Graph Learning.}
Typical tasks include node classification \cite{peng2020graph, wang2021self, jing2021hdmi,DBLP:conf/icml/YouCSW21,DBLP:conf/www/0001XYLWW21}, node clustering \cite{zhang2020commdgi, li2022graph, jing2022x, wang2022clusterscl,liu2023simple,zhao2021graph}, graph classification \cite{velickovic2019deep, sun2019infograph,xu2021infogcl,yin2023coco,luo2022dualgraph}, link prediction \cite{cao2021bipartite,shiao2022link,jing2022coin,he2017neural,kumar2020link}, recommendation \cite{xie2022contrastive,yu2022graph,wei2021contrastive,chen2023heterogeneous,jing2023sterling,yang2023generative,jiang2023adaptive}, knowledge graph reasoning \cite{wang2022simkgc, xu2023temporal, bordes2013translating, sun2018rotate, yan2021dynamic,ji2021survey} and anomaly detection \cite{liu2021anomaly,chen2022gccad,ma2021comprehensive}.

\textbf{Time Series Analysis}
Typical tasks include classification \cite{tonekaboni2020unsupervised,DBLP:conf/ijcai/Eldele0C000G21,zhang2022self,franceschi2019unsupervised,ismail2019deep}, forecasting \cite{DBLP:conf/aaai/YueWDYHTX22,woo2021cost, luo2023time,yang2022unsupervised,lim2021time,jing2022retrieval, zhou2024one,miller2024survey,jing2021network}, anomaly detection \cite{wang2023deep,kim2023contrastive, deldari2021time, yang2023dcdetector, blazquez2021review, ren2019time} and imputation \cite{liu2023timesurl,choi2023conditional,fang2020time, luo2018multivariate,cao2018brits, tashiro2021csdi,jing2024casper}.

\subsubsection{Connecting Downstream Tasks with CL strategies}\label{sec:connection}
Different CL strategies, e.g., different views and pre-training tasks, usually have disparate impact on downstream tasks \cite{DBLP:conf/nips/Tian0PKSI20,DBLP:conf/icml/YouCSW21,luo2023time,zheng2023simts,yeh2023toward}.
Therefore, a fundamental challenge to train foundation models lies in how to construct suitable CL strategies for the desired downstream tasks.
Given a downstream task and a set of available CL strategies, \emph{Automated Machine Learning (AutoML)} \cite{cubuk2018autoaugment,zoph2020learning,DBLP:conf/nips/Tian0PKSI20,yin2022autogcl,luo2023time,tang2023spatio,feng2022adversarial,jin2021automated,jing2024automated} and \emph{prompt learning} \cite{DBLP:conf/mm/ZouHS23,DBLP:conf/wsdm/Xu0QLXHH23, DBLP:conf/acl/AbaskohiRY23, DBLP:conf/acl/WangCZY23, DBLP:conf/acl/ZhengSYLPZXZ23,DBLP:conf/acl/ParanjapeMGHZ21} methods could be used to discover the optimal CL strategies.
Given the optimal CL strategies, one could either use these strategies to pre-train the model and then fine-tune on the downstream tasks, or train the model via \emph{multi-task learning} \cite{yu2022graph,yu2022coca,wang2023characterizing,DBLP:conf/cvpr/0010KBLY21,su2022contrastive,jing2023sterling} by combining the CL strategies with downstream tasks.
Additionally, some works also try to \emph{reformulate} \cite{abs-2004-11362,DBLP:conf/aaai/Li0LPZ021,do2021clustering,sun2018rotate,he2017neural,xie2022contrastive,qiu2021neural,reiss2023mean,eysenbach2022contrastive} the downstream tasks as CL tasks since they are inherently related.

\textbf{Automated Machine Learning.}
Being geared towards automating the procedure of machine learning, AutoML has gained a lot of attention in recent years \cite{he2021automl}.
AutoML methods formulate the problem of searching for the optimal CL strategies as a bi-level optimization problem \cite{cubuk2018autoaugment,DBLP:conf/icml/YouCSW21,jin2021automated}:
\begin{equation}\label{eq:automl}
\begin{split}
    {s}^* &= \arg\max \mathcal{R}(f_{\theta^*}, s)\\
    s.t. \,\theta^* &= \arg\min\mathcal{L}(f_\theta, s)
\end{split}
\end{equation}
where the lower-level problem is to minimize the loss $\mathcal{L}$ (e.g., cross-entropy loss) of the downstream task (e.g., classification) or a surrogate task (e.g., minimization of mutual information \cite{DBLP:conf/nips/Tian0PKSI20}) for the given model $f_\theta$ and the CL strategy $s$;
the upper-level problem is to maximize the validation reward $\mathcal{R}$ (e.g., accuracy) for the pair of the trained model and CL strategy $(f_{\theta^*}, s)$.

Existing AutoML methods can be categorized from two perspectives: 
\emph{search space} and \emph{search algorithms}.
In terms of the search space, existing methods mainly focus on \emph{data augmentations} \cite{cubuk2018autoaugment,li2020differentiable,DBLP:conf/icml/YouCSW21,suresh2021adversarial,reed2021selfaugment}, \emph{view constructions} \cite{DBLP:conf/nips/Tian0PKSI20,yin2022autogcl,luo2023time,tang2023spatio,feng2022adversarial}, \emph{pretext tasks} \cite{jin2021automated} and \emph{overall CL strategies} \cite{jing2024automated}. 
For \emph{data augmentations}, JOAO \cite{DBLP:conf/icml/YouCSW21} could automatically select the most challenging data augmentation pairs for graph data based on the current contrastive loss.
For \emph{view constructions}, InfoMin \cite{DBLP:conf/nips/Tian0PKSI20} argues that good contrastive views should retain information relevant to downstream tasks while minimizing irrelevant nuisances, which constructs the optimal views by minimizing the mutual information between different views.
InfoTS \cite{luo2023time} proposes an information theory-based criteria to evaluate fidelity and variety of the data augmentations for time series, and selects the optimal data augmentations by maximizing the scores derived from the criteria.
For \emph{pretext tasks},
AutoSSL \cite{jin2021automated} automatically searches for the optimal combination of pretext tasks for node clustering and node classification for graphs.
For \emph{overall CL strategies}, AutoCL \cite{jing2024automated} searches for all aspects of CL, including data augmentations, embedding augmentations, contrastive pair construction and loss functions for time series.

In terms of the search algorithms, existing methods are mainly based on \emph{reinforcement learning} \cite{cubuk2018autoaugment,zoph2020learning,jing2024automated}, 
\emph{adversarial learning} \cite{DBLP:conf/nips/Tian0PKSI20,DBLP:conf/icml/YouCSW21,yin2022autogcl,luo2023time,feng2024ariel,suresh2021adversarial,tang2023spatio},
\emph{evolution strategy} \cite{jin2021automated} and \emph{Bayesian optimization}~\cite{reed2021selfaugment}. 
For \emph{reinforcement learning}, AutoCL \cite{jing2024automated} uses a controller network to sample CL strategies and uses the model's performance on the validation set to design reward, where the controller is optimized by maximizing the reward $\mathcal{R}$ via reinforcement learning. 
For \emph{adversarial leanring}
AD-GCL \cite{suresh2021adversarial} measures the similarity of the node embeddings of two different graph views via mutual information, which formulates the lower-level and upper-level problems in Equation \eqref{eq:automl} as maximizing and minimizing the mutual information respectively.
ARIEL \cite{feng2022adversarial} uses the same contrastive loss for both lower-level loss $\mathcal{L}$ and the upper-level reward $\mathcal{R}$, and uses the adversarial attack to maximize $\mathcal{R}$.
For \emph{evolution strategy},
AutoSSL-ES \cite{jin2021automated} has adopted an evolution strategy \cite{hansen2003reducing} to obtain the optimal combination of CL strategies for graphs by optimizing the pseudo-homophily objective.
For \emph{Bayesian optimization},
SelfAugment \cite{reed2021selfaugment} leverages image rotation prediction as the lower-level task, and uses Bayesian optimization \cite{lim2019fast} as the search algorithm to obtain the optimal data augmentations.

\textbf{Prompt Learning.}
% \hh{missing?} \lc{I will add the related work soon.}
The contrastive-based prompt learning methods aim to combine CL with prompt learning for various purposes, such as maximizing the consistency of different representations~\cite{DBLP:conf/mm/ZouHS23}, enabling fine-tuning in few-shot or zero-shot setting~\cite{DBLP:conf/wsdm/Xu0QLXHH23, DBLP:conf/acl/AbaskohiRY23, DBLP:conf/acl/WangCZY23, DBLP:conf/acl/ZhengSYLPZXZ23}, and commonsense reasoning~\cite{DBLP:conf/acl/ParanjapeMGHZ21}. Specifically,~\cite{DBLP:conf/mm/ZouHS23} designs a multimodal prompt transformer to perform cross-modal information fusion and apply CL to maximize the consistency among the fused representation and the representation for each modality for the emotion recognition task. \cite{DBLP:conf/wsdm/Xu0QLXHH23, DBLP:conf/acl/AbaskohiRY23, DBLP:conf/acl/WangCZY23, DBLP:conf/acl/ZhengSYLPZXZ23} combine CL with prompt learning to fine-tune the model in the few-shot or zero-shot setting.
\cite{DBLP:conf/nips/0003KKW23} devises visual prompt-based CL and guided-attention-based prompt ensemble algorithms to task-learn specific state representations from multiple prompted embeddings. \cite{DBLP:conf/acl/ParanjapeMGHZ21} develops a list of contrastive generation prompts for the commonsense reasoning task.

\textbf{Multi-Task Learning.}
When we have prior knowledge about what characteristics of the data can be brought by CL strategies, in addition to the pre-training and fine-tuning paradigm, another simple yet effective way to connect downstream tasks and CL strategies is to combine the pre-training tasks and downstream tasks as a multi-task learning task.
For example, 
CoCa \cite{yu2022coca} combines a contrastive loss with a caption generation loss to train image-text foundation models, where contrastive loss is used to learn global representations and captioning is used to learn fine-grained region-level features.
XMC-GAN \cite{DBLP:conf/cvpr/0010KBLY21} leverages contrastive losses for various pairs, such as (image, sentence) and (generated image, real image), to improve the alignment between them for the text-to-image generation task.
mRASP2 \cite{DBLP:conf/acl/PanWWL20} combines a contrastive loss with cross-entropy for multilingual machine translation, where the contrastive loss is adopted to minimize the representation gap of similar sentences and maximize that of irrelevant sentences.
SimCTG \cite{su2022contrastive} leverages a contrastive loss to encourage language models to learn discriminative and isotropic token representations for neural text generation.
SimGCL \cite{yu2022graph} discovers that InfoNCE loss helps models learn more evenly distributed user and item embeddings, which could mitigate the popularity bias. 
% SimGCL combines InfoNCE with a recommendation loss \cite{rendle2009bpr} to train the recommendation models.
STERLING \cite{jing2023sterling} combines the BYOL loss \cite{DBLP:conf/nips/GrillSATRBDPGAP20} with a co-clustering loss for co-clustering on bipartite graphs.

\textbf{Task Reformulation.}
Certain downstream tasks are inherently related to CL, such as \emph{classification}, \emph{clustering}, \emph{link prediction}, \emph{recommendation}, \emph{anomaly detection} and \emph{reinforcement learning}.
Therefore, the loss functions of these downstream tasks can be reformulated as a contrastive loss.
For example, in terms of \emph{classification}, the previously mentioned SupCon \cite{abs-2004-11362} integrates image class labels into self-supervised contrastive losses and proposes a Supervised Contrastive (SupCon) loss.
CLIP \cite{DBLP:conf/icml/RadfordKHRGASAM21} reformulates the image classification task as an image-text alignment contrastive task.
For \emph{clustering}, CC \cite{DBLP:conf/aaai/Li0LPZ021} introduces a CL-based clustering objective function, called contrastive clustering, by regarding the embedding vector of an instance as the soft cluster labels.
CRLC \cite{do2021clustering} reformulates the objective of clustering as a probability contrastive loss, which trains the parametric clustering classifier by contrasting positive and negative cluster probability pairs.
For \emph{link prediction}, since it is inherently a contrastive task: determining whether a pair of nodes is positive or not, most of the existing methods directly adopt CL losses as the objective functions to train the models \cite{ji2021survey}. For example, RotatE \cite{sun2018rotate} trains knowledge graph link prediction models by the negative sampling loss \cite{mikolov2013distributed}.
For \emph{recommendation}, it can be regarded as a link prediction task with ranking, and thus the training objectives are usually variants of contrastive losses \cite{yan2024reconciling,he2017neural}.
For example, CL4SRec \cite{xie2022contrastive} formulates the objective function of recommendation as a variant of InfoNCE \cite{oord2018representation}.
For \emph{anomaly detection}, 
NeuTraL \cite{qiu2021neural} directly adopts a contrastive loss as the loss function as well as the anomaly score.
MSC \cite{reiss2023mean} introduces a Mean-Shifted Contrastive (MSC) loss for one-class classification, which computes the angles between the image embeddings and the center of the normal embeddings.
For \emph{reinforcement learning}, Contrastive RL \cite{eysenbach2022contrastive} uses CL to directly perform goal-conditioned reinforcement learning by leveraging CL to estimate the Q-function for a certain policy and reward functions.

\section{Future Directions}
The past years have witnessed the rapid development of heterogeneous CL on foundation models. Building upon such progress, it opens the door to many exciting future opportunities to explore in this emerging area. Here, we summarize five promising future directions, focusing on contrastive foundation models. % due to the space limitation, though there are lots of interesting future research directions to pursue.
%\hh{1. can we add a short paragraph, to first briefly summarize the survey and then point out there are a lot of interesting future directions to pursue, such as ...? 2. consider to revise the heading of each future direction so that their connection to CL is clear from the heading. for instace, instead of 'trustworthiness', can we call it 'trustworthy contrastive learning'? in this way, the readers will be clear that we are talking about these topics in the context of CL (instead of a general introduction of e.g., trustworthy ML).} \lc{Sure. I will update this section soon.}

\textbf{Representation Redundancy and Uniqueness for CL Model.}
The current CL models mainly extract the shared representation by maximizing the similarity of two views for the same sample. However, some recent works~\cite{DBLP:conf/nips/LiangDMZMS23, zheng2024multi} have suggested the potential of extracting uniqueness via CL to improve the performance of the downstream tasks. However, the initial methods are only used to deal with the small model, and how to naturally combine it with foundation models remains a great challenge due to the extra computational cost and limited performance improvement at the current stage.

\textbf{Efficiency of CL Foundation Models.}
One major issue of training the foundation models with CL loss is the high GPU memory requirement as discussed in Section~\ref{Large_vision_model}. 
% The time complexity of most CL methods requires $O(n^2)$ to compute the similarity matrix for both positive and negative pairs, where $n$ is the number of samples.
Recently, Zeroth order optimization methods~\cite{malladi2024fine} have shown great potential to alleviate the computational cost by replacing the traditional forward-passing and backward-passing optimization scheme with a forward-passing-only optimization scheme. However, the zeroth order optimizer usually sacrifices the optimization efficiency for lower GPU requirements, as it requires significantly more steps than standard fine-tuning~\cite{malladi2024fine}.
% \hh{do we have a reference for this?}.
Efficiently training or fine-tuning a CL-based foundation model remains a great challenge.

\textbf{Better Multi-view Benchmark Datasets for CL Models.}
Currently, high-quality multi-view benchmark datasets are urgently needed for constructing multi-modal foundation models. While many large-scale text-attributed graphs are collected from social media, e-commerce platforms, and academic domains~\cite{DBLP:journals/corr/abs-2312-02783}, it's essential to acknowledge that real-world graphs span various domains, including finance, healthcare, transportation networks, and local infrastructure networks. Similarly, there is a high demand for large-scale text-image datasets to support the training of vision-language foundational models~\cite{wang2022diffusiondb}. Moreover, concerns have been raised regarding potential biases present in many benchmark datasets. Studies ~\cite{clark-etal-2019-dont, wan2023kelly, oba2023contextual, kotek2023gender} have highlighted the existence of contextual, demographic, and stereotypical biases within benchmark datasets used for large language models
% For instance, most large-scale text-attributed graphs are collected from social media, e-commerce platforms, and academic domains~\cite{DBLP:journals/corr/abs-2312-02783}. However, real-world graphs are ubiquitous, ranging from finance to health, from transportation networks to local infrastructure networks, etc. Similarly, large-scale text-image datasets are in high demand for training the vision language foundation model~\cite{wang2022diffusiondb}. In addition, some paper points out that potential bias exists in many benchmark datasets. For instance,~\cite{clark-etal-2019-dont, wan2023kelly, oba2023contextual, kotek2023gender} show the existence of contextual bias, demographics bias, stereotypical bias in the benchmark dataset for large language model.

\textbf{Trustworthy CL.}
% \hh{the connection between this paragraph and CL seems weak. for instance, we did not mention CL at all in this paragraph.}\lc{I have updated this future direction.}
Trustworthy machine learning refers to the development and deployment of machine learning models with a strong emphasis on interpretability, fairness, transparency, privacy, and robustness. While significant strides have been made in enhancing these aspects by CL-based regularization, inlcuding interpretability~\cite{DBLP:conf/emnlp/JacoviSRECG21, DBLP:journals/corr/abs-2005-12419}, fairness considerations~\cite{zhang2022fairness, zheng2023fairness, wang2022uncovering}, and out-of-distribution robustness~\cite{zheng2024multi, DBLP:conf/acl/MaSA21, DBLP:conf/nips/ZhangR22}, these efforts are still in their nascent stages, \eg, training models on small datasets or failing to consider the view or task heterogeneity.  Despite these early efforts, heterogeneous contrastive foundational models still encounter challenges related to interpretability, fairness, transparency, privacy, and robustness, which persist across multi-modal foundational models as well.

\textbf{Understanding Mechanisms Between CL Strategies and Downstream Tasks.}
As introduced in Section \ref{downstream_tasks}, we present various CL strategies for downstream tasks. However, it remains unclear which CL strategies are good for specific downstream tasks and how can we evaluate the quality of CL strategies. In addition, how different CL strategies compete and cooperate in downstream tasks is expected to be better evaluated and understood by the researchers. Moreover, how to combine CL with other self-supervised methods to further improve the performance of the foundation models deserves great attention.

% \by{
% 1. Connecting CL strategies and downstream tasks, which CL strategy is good for a specific downstream tasks? do we really need CL pre-training for some downstream tasks?\\
% 2. how to evaluate the quality of CL strategies? \\
% 3. further understand the CL strategies: 
% competition and cooperation between different strategies.\\
% 3. how to combine different datasets, how to address data bias, \\
% 4. continual learning for CL, \\
% 5. how to combine CL with other SSL methods, when to use CL and when to use other SSL methods, what are the advantages/disadvantages of CL over other SSL methods? \\
% 6. beyond 0/1 contrastive pairs, add confidences and weights for different pairs, more precise guidance: contrast words within sentences rather than the entire sentences. \\
% 7. active learning: can model actively identify which concepts are well-studied, which needs further clarification.
% }

\section{Conclusion}
This paper provides a thorough exploration of heterogeneous CL for foundation models. We first delve into the traditional CL methods, particularly in addressing view heterogeneity, and elucidate the application of CL techniques in training and fine-tuning multi-view foundation models. Subsequently, we discuss CL methods tailored to tackle task heterogeneity, including pretraining and downstream tasks, and illustrate how CL combines different tasks for various objectives. Finally, we outline potential future research directions in heterogeneous CL for foundation models.

\newpage
\bibliographystyle{ACM-Reference-Format}
\bibliography{reference}


\begin{thebibliography}{291}

%%% the Web address.

\ifx \showCODEN    \undefined \def \showCODEN     #1{\unskip}     \fi
\ifx \showDOI      \undefined \def \showDOI       #1{#1}\fi
\ifx \showISBNx    \undefined \def \showISBNx     #1{\unskip}     \fi
\ifx \showISBNxiii \undefined \def \showISBNxiii  #1{\unskip}     \fi
\ifx \showISSN     \undefined \def \showISSN      #1{\unskip}     \fi
\ifx \showLCCN     \undefined \def \showLCCN      #1{\unskip}     \fi
\ifx \shownote     \undefined \def \shownote      #1{#1}          \fi
\ifx \showarticletitle \undefined \def \showarticletitle #1{#1}   \fi
\ifx \showURL      \undefined \def \showURL       {\relax}        \fi
% invisible to TeX
\providecommand\bibfield[2]{#2}
\providecommand\bibinfo[2]{#2}
\providecommand\natexlab[1]{#1}
\providecommand\showeprint[2][]{arXiv:#2}

\bibitem[Abaskohi et~al\mbox{.}(2023)]%
        {DBLP:conf/acl/AbaskohiRY23}
\bibfield{author}{\bibinfo{person}{Amirhossein Abaskohi},
  \bibinfo{person}{Sascha Rothe}, {and} \bibinfo{person}{Yadollah
  Yaghoobzadeh}.} \bibinfo{year}{2023}\natexlab{}.
\newblock \showarticletitle{{LM-CPPF:} Paraphrasing-Guided Data Augmentation
  for Contrastive Prompt-Based Few-Shot Fine-Tuning}. In
  \bibinfo{booktitle}{\emph{Proceedings of the 61st Annual Meeting of the
  Association for Computational Linguistics (Volume 2: Short Papers), {ACL}
  2023, Toronto, Canada, July 9-14, 2023}}. \bibinfo{publisher}{Association for
  Computational Linguistics}, \bibinfo{pages}{670--681}.
\newblock


\bibitem[Alayrac et~al\mbox{.}(2022)]%
        {DBLP:conf/nips/AlayracDLMBHLMM22}
\bibfield{author}{\bibinfo{person}{Jean{-}Baptiste Alayrac},
  \bibinfo{person}{Jeff Donahue}, \bibinfo{person}{Pauline Luc},
  \bibinfo{person}{Antoine Miech}, \bibinfo{person}{Iain Barr},
  \bibinfo{person}{Yana Hasson}, \bibinfo{person}{Karel Lenc},
  \bibinfo{person}{Arthur Mensch}, \bibinfo{person}{Katherine Millican},
  \bibinfo{person}{Malcolm Reynolds}, \bibinfo{person}{Roman Ring},
  \bibinfo{person}{Eliza Rutherford}, \bibinfo{person}{Serkan Cabi},
  \bibinfo{person}{Tengda Han}, \bibinfo{person}{Zhitao Gong},
  \bibinfo{person}{Sina Samangooei}, \bibinfo{person}{Marianne Monteiro},
  \bibinfo{person}{Jacob~L. Menick}, \bibinfo{person}{Sebastian Borgeaud},
  \bibinfo{person}{Andy Brock}, \bibinfo{person}{Aida Nematzadeh},
  \bibinfo{person}{Sahand Sharifzadeh}, \bibinfo{person}{Mikolaj Binkowski},
  \bibinfo{person}{Ricardo Barreira}, \bibinfo{person}{Oriol Vinyals},
  \bibinfo{person}{Andrew Zisserman}, {and} \bibinfo{person}{Kar{\'{e}}n
  Simonyan}.} \bibinfo{year}{2022}\natexlab{}.
\newblock \showarticletitle{Flamingo: a Visual Language Model for Few-Shot
  Learning}. In \bibinfo{booktitle}{\emph{Advances in Neural Information
  Processing Systems 35: Annual Conference on Neural Information Processing
  Systems 2022, NeurIPS 2022, New Orleans, LA, USA, November 28 - December 9,
  2022}}.
\newblock


\bibitem[Albelwi(2022)]%
        {DBLP:journals/entropy/Albelwi22}
\bibfield{author}{\bibinfo{person}{Saleh Albelwi}.}
  \bibinfo{year}{2022}\natexlab{}.
\newblock \showarticletitle{Survey on Self-Supervised Learning: Auxiliary
  Pretext Tasks and Contrastive Learning Methods in Imaging}.
\newblock \bibinfo{journal}{\emph{Entropy}} \bibinfo{volume}{24},
  \bibinfo{number}{4} (\bibinfo{year}{2022}), \bibinfo{pages}{551}.
\newblock


\bibitem[Ayush et~al\mbox{.}(2021)]%
        {ayush2021geography}
\bibfield{author}{\bibinfo{person}{Kumar Ayush}, \bibinfo{person}{Burak
  Uzkent}, \bibinfo{person}{Chenlin Meng}, \bibinfo{person}{Kumar Tanmay},
  \bibinfo{person}{Marshall Burke}, \bibinfo{person}{David Lobell}, {and}
  \bibinfo{person}{Stefano Ermon}.} \bibinfo{year}{2021}\natexlab{}.
\newblock \showarticletitle{Geography-aware self-supervised learning}. In
  \bibinfo{booktitle}{\emph{Proceedings of the IEEE/CVF International
  Conference on Computer Vision}}. \bibinfo{pages}{10181--10190}.
\newblock


\bibitem[Badamdorj et~al\mbox{.}(2022)]%
        {DBLP:conf/cvpr/BadamdorjR0C22}
\bibfield{author}{\bibinfo{person}{Taivanbat Badamdorj},
  \bibinfo{person}{Mrigank Rochan}, \bibinfo{person}{Yang Wang}, {and}
  \bibinfo{person}{Li Cheng}.} \bibinfo{year}{2022}\natexlab{}.
\newblock \showarticletitle{Contrastive Learning for Unsupervised Video
  Highlight Detection}. In \bibinfo{booktitle}{\emph{{IEEE/CVF} Conference on
  Computer Vision and Pattern Recognition, {CVPR} 2022, New Orleans, LA, USA,
  June 18-24, 2022}}. \bibinfo{publisher}{{IEEE}},
  \bibinfo{pages}{14022--14032}.
\newblock


\bibitem[Bl{\'a}zquez-Garc{\'\i}a et~al\mbox{.}(2021)]%
        {blazquez2021review}
\bibfield{author}{\bibinfo{person}{Ane Bl{\'a}zquez-Garc{\'\i}a},
  \bibinfo{person}{Angel Conde}, \bibinfo{person}{Usue Mori}, {and}
  \bibinfo{person}{Jose~A Lozano}.} \bibinfo{year}{2021}\natexlab{}.
\newblock \showarticletitle{A review on outlier/anomaly detection in time
  series data}.
\newblock \bibinfo{journal}{\emph{ACM Computing Surveys (CSUR)}}
  \bibinfo{volume}{54}, \bibinfo{number}{3} (\bibinfo{year}{2021}),
  \bibinfo{pages}{1--33}.
\newblock


\bibitem[Bordes et~al\mbox{.}(2013)]%
        {bordes2013translating}
\bibfield{author}{\bibinfo{person}{Antoine Bordes}, \bibinfo{person}{Nicolas
  Usunier}, \bibinfo{person}{Alberto Garcia-Duran}, \bibinfo{person}{Jason
  Weston}, {and} \bibinfo{person}{Oksana Yakhnenko}.}
  \bibinfo{year}{2013}\natexlab{}.
\newblock \showarticletitle{Translating embeddings for modeling
  multi-relational data}.
\newblock \bibinfo{journal}{\emph{Advances in neural information processing
  systems}}  \bibinfo{volume}{26} (\bibinfo{year}{2013}).
\newblock


\bibitem[Brannon et~al\mbox{.}(2023)]%
        {DBLP:journals/corr/abs-2305-14321}
\bibfield{author}{\bibinfo{person}{William Brannon}, \bibinfo{person}{Suyash
  Fulay}, \bibinfo{person}{Hang Jiang}, \bibinfo{person}{Wonjune Kang},
  \bibinfo{person}{Brandon Roy}, \bibinfo{person}{Jad Kabbara}, {and}
  \bibinfo{person}{Deb Roy}.} \bibinfo{year}{2023}\natexlab{}.
\newblock \showarticletitle{ConGraT: Self-Supervised Contrastive Pretraining
  for Joint Graph and Text Embeddings}.
\newblock \bibinfo{journal}{\emph{CoRR}}  \bibinfo{volume}{abs/2305.14321}
  (\bibinfo{year}{2023}).
\newblock


\bibitem[Cao et~al\mbox{.}(2021)]%
        {cao2021bipartite}
\bibfield{author}{\bibinfo{person}{Jiangxia Cao}, \bibinfo{person}{Xixun Lin},
  \bibinfo{person}{Shu Guo}, \bibinfo{person}{Luchen Liu},
  \bibinfo{person}{Tingwen Liu}, {and} \bibinfo{person}{Bin Wang}.}
  \bibinfo{year}{2021}\natexlab{}.
\newblock \showarticletitle{Bipartite graph embedding via mutual information
  maximization}. In \bibinfo{booktitle}{\emph{Proceedings of the 14th ACM
  international conference on web search and data mining}}.
  \bibinfo{pages}{635--643}.
\newblock


\bibitem[Cao et~al\mbox{.}(2018)]%
        {cao2018brits}
\bibfield{author}{\bibinfo{person}{Wei Cao}, \bibinfo{person}{Dong Wang},
  \bibinfo{person}{Jian Li}, \bibinfo{person}{Hao Zhou}, \bibinfo{person}{Lei
  Li}, {and} \bibinfo{person}{Yitan Li}.} \bibinfo{year}{2018}\natexlab{}.
\newblock \showarticletitle{Brits: Bidirectional recurrent imputation for time
  series}.
\newblock \bibinfo{journal}{\emph{Advances in neural information processing
  systems}}  \bibinfo{volume}{31} (\bibinfo{year}{2018}).
\newblock


\bibitem[Chen et~al\mbox{.}(2022c)]%
        {chen2022gccad}
\bibfield{author}{\bibinfo{person}{Bo Chen}, \bibinfo{person}{Jing Zhang},
  \bibinfo{person}{Xiaokang Zhang}, \bibinfo{person}{Yuxiao Dong},
  \bibinfo{person}{Jian Song}, \bibinfo{person}{Peng Zhang},
  \bibinfo{person}{Kaibo Xu}, \bibinfo{person}{Evgeny Kharlamov}, {and}
  \bibinfo{person}{Jie Tang}.} \bibinfo{year}{2022}\natexlab{c}.
\newblock \showarticletitle{Gccad: Graph contrastive learning for anomaly
  detection}.
\newblock \bibinfo{journal}{\emph{IEEE Transactions on Knowledge and Data
  Engineering}} (\bibinfo{year}{2022}).
\newblock


\bibitem[Chen et~al\mbox{.}(2021)]%
        {DBLP:conf/nips/ChenZWZZLXL21}
\bibfield{author}{\bibinfo{person}{Haibo Chen}, \bibinfo{person}{Lei Zhao},
  \bibinfo{person}{Zhizhong Wang}, \bibinfo{person}{Huiming Zhang},
  \bibinfo{person}{Zhiwen Zuo}, \bibinfo{person}{Ailin Li},
  \bibinfo{person}{Wei Xing}, {and} \bibinfo{person}{Dongming Lu}.}
  \bibinfo{year}{2021}\natexlab{}.
\newblock \showarticletitle{Artistic Style Transfer with Internal-external
  Learning and Contrastive Learning}. In \bibinfo{booktitle}{\emph{Advances in
  Neural Information Processing Systems 34: Annual Conference on Neural
  Information Processing Systems 2021, NeurIPS 2021, December 6-14, 2021,
  virtual}}. \bibinfo{pages}{26561--26573}.
\newblock


\bibitem[Chen et~al\mbox{.}(2020c)]%
        {DBLP:conf/acl/ChenYY20}
\bibfield{author}{\bibinfo{person}{Jiaao Chen}, \bibinfo{person}{Zichao Yang},
  {and} \bibinfo{person}{Diyi Yang}.} \bibinfo{year}{2020}\natexlab{c}.
\newblock \showarticletitle{MixText: Linguistically-Informed Interpolation of
  Hidden Space for Semi-Supervised Text Classification}. In
  \bibinfo{booktitle}{\emph{Proceedings of the 58th Annual Meeting of the
  Association for Computational Linguistics, {ACL} 2020, Online, July 5-10,
  2020}}, \bibfield{editor}{\bibinfo{person}{Dan Jurafsky},
  \bibinfo{person}{Joyce Chai}, \bibinfo{person}{Natalie Schluter}, {and}
  \bibinfo{person}{Joel~R. Tetreault}} (Eds.). \bibinfo{publisher}{Association
  for Computational Linguistics}, \bibinfo{pages}{2147--2157}.
\newblock


\bibitem[Chen et~al\mbox{.}(2022b)]%
        {chen2022contrastnet}
\bibfield{author}{\bibinfo{person}{Junfan Chen}, \bibinfo{person}{Richong
  Zhang}, \bibinfo{person}{Yongyi Mao}, {and} \bibinfo{person}{Jie Xu}.}
  \bibinfo{year}{2022}\natexlab{b}.
\newblock \showarticletitle{Contrastnet: A contrastive learning framework for
  few-shot text classification}. In \bibinfo{booktitle}{\emph{Proceedings of
  the AAAI Conference on Artificial Intelligence}}, Vol.~\bibinfo{volume}{36}.
  \bibinfo{pages}{10492--10500}.
\newblock


\bibitem[Chen et~al\mbox{.}(2023a)]%
        {chen2023heterogeneous}
\bibfield{author}{\bibinfo{person}{Mengru Chen}, \bibinfo{person}{Chao Huang},
  \bibinfo{person}{Lianghao Xia}, \bibinfo{person}{Wei Wei},
  \bibinfo{person}{Yong Xu}, {and} \bibinfo{person}{Ronghua Luo}.}
  \bibinfo{year}{2023}\natexlab{a}.
\newblock \showarticletitle{Heterogeneous graph contrastive learning for
  recommendation}. In \bibinfo{booktitle}{\emph{Proceedings of the sixteenth
  ACM international conference on web search and data mining}}.
  \bibinfo{pages}{544--552}.
\newblock


\bibitem[Chen et~al\mbox{.}(2020b)]%
        {chen2020simple}
\bibfield{author}{\bibinfo{person}{Ting Chen}, \bibinfo{person}{Simon
  Kornblith}, \bibinfo{person}{Mohammad Norouzi}, {and}
  \bibinfo{person}{Geoffrey Hinton}.} \bibinfo{year}{2020}\natexlab{b}.
\newblock \showarticletitle{A simple framework for contrastive learning of
  visual representations}. In \bibinfo{booktitle}{\emph{International
  conference on machine learning}}. PMLR, \bibinfo{pages}{1597--1607}.
\newblock


\bibitem[Chen et~al\mbox{.}(2022a)]%
        {chen2022learning}
\bibfield{author}{\bibinfo{person}{Tianyu Chen}, \bibinfo{person}{Yuan Xie},
  \bibinfo{person}{Shuai Zhang}, \bibinfo{person}{Shaohan Huang},
  \bibinfo{person}{Haoyi Zhou}, {and} \bibinfo{person}{Jianxin Li}.}
  \bibinfo{year}{2022}\natexlab{a}.
\newblock \showarticletitle{Learning music sequence representation from text
  supervision}. In \bibinfo{booktitle}{\emph{ICASSP 2022-2022 IEEE
  International Conference on Acoustics, Speech and Signal Processing
  (ICASSP)}}. IEEE, \bibinfo{pages}{4583--4587}.
\newblock


\bibitem[Chen et~al\mbox{.}(2020a)]%
        {DBLP:journals/corr/abs-2003-04297}
\bibfield{author}{\bibinfo{person}{Xinlei Chen}, \bibinfo{person}{Haoqi Fan},
  \bibinfo{person}{Ross~B. Girshick}, {and} \bibinfo{person}{Kaiming He}.}
  \bibinfo{year}{2020}\natexlab{a}.
\newblock \showarticletitle{Improved Baselines with Momentum Contrastive
  Learning}.
\newblock \bibinfo{journal}{\emph{CoRR}}  \bibinfo{volume}{abs/2003.04297}
  (\bibinfo{year}{2020}).
\newblock


\bibitem[Chen et~al\mbox{.}(2024)]%
        {DBLP:journals/tetci/ChenWFML24}
\bibfield{author}{\bibinfo{person}{Xiaoru Chen}, \bibinfo{person}{Yingxu Wang},
  \bibinfo{person}{Jinyuan Fang}, \bibinfo{person}{Zaiqiao Meng}, {and}
  \bibinfo{person}{Shangsong Liang}.} \bibinfo{year}{2024}\natexlab{}.
\newblock \showarticletitle{Heterogeneous Graph Contrastive Learning With
  Metapath-Based Augmentations}.
\newblock \bibinfo{journal}{\emph{{IEEE} Trans. Emerg. Top. Comput. Intell.}}
  \bibinfo{volume}{8}, \bibinfo{number}{1} (\bibinfo{year}{2024}),
  \bibinfo{pages}{1003--1014}.
\newblock


\bibitem[Chen et~al\mbox{.}(2023b)]%
        {DBLP:conf/acl/ChenLZYW23}
\bibfield{author}{\bibinfo{person}{Zhongzhi Chen}, \bibinfo{person}{Guang Liu},
  \bibinfo{person}{Bo{-}Wen Zhang}, \bibinfo{person}{Qinghong Yang}, {and}
  \bibinfo{person}{Ledell Wu}.} \bibinfo{year}{2023}\natexlab{b}.
\newblock \showarticletitle{AltCLIP: Altering the Language Encoder in {CLIP}
  for Extended Language Capabilities}. In \bibinfo{booktitle}{\emph{Findings of
  the Association for Computational Linguistics: {ACL} 2023, Toronto, Canada,
  July 9-14, 2023}}, \bibfield{editor}{\bibinfo{person}{Anna Rogers},
  \bibinfo{person}{Jordan~L. Boyd{-}Graber}, {and} \bibinfo{person}{Naoaki
  Okazaki}} (Eds.). \bibinfo{publisher}{Association for Computational
  Linguistics}, \bibinfo{pages}{8666--8682}.
\newblock


\bibitem[Choi and Lee(2023)]%
        {choi2023conditional}
\bibfield{author}{\bibinfo{person}{MinGyu Choi} {and} \bibinfo{person}{Changhee
  Lee}.} \bibinfo{year}{2023}\natexlab{}.
\newblock \showarticletitle{Conditional Information Bottleneck Approach for
  Time Series Imputation}. In \bibinfo{booktitle}{\emph{The Twelfth
  International Conference on Learning Representations}}.
\newblock


\bibitem[Choi et~al\mbox{.}(2022)]%
        {choi2022c2l}
\bibfield{author}{\bibinfo{person}{Seungtaek Choi}, \bibinfo{person}{Myeongho
  Jeong}, \bibinfo{person}{Hojae Han}, {and} \bibinfo{person}{Seung-won
  Hwang}.} \bibinfo{year}{2022}\natexlab{}.
\newblock \showarticletitle{C2l: Causally contrastive learning for robust text
  classification}. In \bibinfo{booktitle}{\emph{Proceedings of the AAAI
  Conference on Artificial Intelligence}}, Vol.~\bibinfo{volume}{36}.
  \bibinfo{pages}{10526--10534}.
\newblock


\bibitem[Choi et~al\mbox{.}(2023)]%
        {DBLP:conf/nips/0003KKW23}
\bibfield{author}{\bibinfo{person}{Wonje Choi}, \bibinfo{person}{Woo~Kyung
  Kim}, \bibinfo{person}{Seunghyun Kim}, {and} \bibinfo{person}{Honguk Woo}.}
  \bibinfo{year}{2023}\natexlab{}.
\newblock \showarticletitle{Efficient Policy Adaptation with Contrastive Prompt
  Ensemble for Embodied Agents}. In \bibinfo{booktitle}{\emph{Advances in
  Neural Information Processing Systems 36: Annual Conference on Neural
  Information Processing Systems 2023, NeurIPS 2023, New Orleans, LA, USA,
  December 10 - 16, 2023}}, \bibfield{editor}{\bibinfo{person}{Alice Oh},
  \bibinfo{person}{Tristan Naumann}, \bibinfo{person}{Amir Globerson},
  \bibinfo{person}{Kate Saenko}, \bibinfo{person}{Moritz Hardt}, {and}
  \bibinfo{person}{Sergey Levine}} (Eds.).
\newblock


\bibitem[Clark et~al\mbox{.}(2019)]%
        {clark-etal-2019-dont}
\bibfield{author}{\bibinfo{person}{Christopher Clark}, \bibinfo{person}{Mark
  Yatskar}, {and} \bibinfo{person}{Luke Zettlemoyer}.}
  \bibinfo{year}{2019}\natexlab{}.
\newblock \showarticletitle{Don{'}t Take the Easy Way Out: Ensemble Based
  Methods for Avoiding Known Dataset Biases}. In
  \bibinfo{booktitle}{\emph{Proceedings of the 2019 Conference on Empirical
  Methods in Natural Language Processing and the 9th International Joint
  Conference on Natural Language Processing (EMNLP-IJCNLP)}},
  \bibfield{editor}{\bibinfo{person}{Kentaro Inui}, \bibinfo{person}{Jing
  Jiang}, \bibinfo{person}{Vincent Ng}, {and} \bibinfo{person}{Xiaojun Wan}}
  (Eds.). \bibinfo{publisher}{Association for Computational Linguistics},
  \bibinfo{address}{Hong Kong, China}.
\newblock


\bibitem[Cole et~al\mbox{.}(2022)]%
        {DBLP:conf/cvpr/ColeYWAB22}
\bibfield{author}{\bibinfo{person}{Elijah Cole}, \bibinfo{person}{Xuan Yang},
  \bibinfo{person}{Kimberly Wilber}, \bibinfo{person}{Oisin~Mac Aodha}, {and}
  \bibinfo{person}{Serge~J. Belongie}.} \bibinfo{year}{2022}\natexlab{}.
\newblock \showarticletitle{When Does Contrastive Visual Representation
  Learning Work?}. In \bibinfo{booktitle}{\emph{{IEEE/CVF} Conference on
  Computer Vision and Pattern Recognition, {CVPR} 2022, New Orleans, LA, USA,
  June 18-24, 2022}}. \bibinfo{publisher}{{IEEE}}, \bibinfo{pages}{1--10}.
\newblock


\bibitem[Cubuk et~al\mbox{.}(2018)]%
        {cubuk2018autoaugment}
\bibfield{author}{\bibinfo{person}{Ekin~D Cubuk}, \bibinfo{person}{Barret
  Zoph}, \bibinfo{person}{Dandelion Mane}, \bibinfo{person}{Vijay Vasudevan},
  {and} \bibinfo{person}{Quoc~V Le}.} \bibinfo{year}{2018}\natexlab{}.
\newblock \showarticletitle{Autoaugment: Learning augmentation policies from
  data}.
\newblock \bibinfo{journal}{\emph{arXiv preprint arXiv:1805.09501}}
  (\bibinfo{year}{2018}).
\newblock


\bibitem[Cui et~al\mbox{.}(2021)]%
        {DBLP:conf/iccv/CuiZ00J21}
\bibfield{author}{\bibinfo{person}{Jiequan Cui}, \bibinfo{person}{Zhisheng
  Zhong}, \bibinfo{person}{Shu Liu}, \bibinfo{person}{Bei Yu}, {and}
  \bibinfo{person}{Jiaya Jia}.} \bibinfo{year}{2021}\natexlab{}.
\newblock \showarticletitle{Parametric Contrastive Learning}. In
  \bibinfo{booktitle}{\emph{2021 {IEEE/CVF} International Conference on
  Computer Vision, {ICCV} 2021, Montreal, QC, Canada, October 10-17, 2021}}.
  \bibinfo{publisher}{{IEEE}}, \bibinfo{pages}{695--704}.
\newblock


\bibitem[Deldari et~al\mbox{.}(2021)]%
        {deldari2021time}
\bibfield{author}{\bibinfo{person}{Shohreh Deldari}, \bibinfo{person}{Daniel~V
  Smith}, \bibinfo{person}{Hao Xue}, {and} \bibinfo{person}{Flora~D Salim}.}
  \bibinfo{year}{2021}\natexlab{}.
\newblock \showarticletitle{Time series change point detection with
  self-supervised contrastive predictive coding}. In
  \bibinfo{booktitle}{\emph{Proceedings of the Web Conference 2021}}.
  \bibinfo{pages}{3124--3135}.
\newblock


\bibitem[Deng et~al\mbox{.}(2023)]%
        {DBLP:conf/iccv/DengSHLXHKZZL23}
\bibfield{author}{\bibinfo{person}{Xinchi Deng}, \bibinfo{person}{Han Shi},
  \bibinfo{person}{Runhui Huang}, \bibinfo{person}{Changlin Li},
  \bibinfo{person}{Hang Xu}, \bibinfo{person}{Jianhua Han},
  \bibinfo{person}{James~T. Kwok}, \bibinfo{person}{Shen Zhao},
  \bibinfo{person}{Wei Zhang}, {and} \bibinfo{person}{Xiaodan Liang}.}
  \bibinfo{year}{2023}\natexlab{}.
\newblock \showarticletitle{GrowCLIP: Data-aware Automatic Model Growing for
  Large-scale Contrastive Language-Image Pre-training}. In
  \bibinfo{booktitle}{\emph{{IEEE/CVF} International Conference on Computer
  Vision, {ICCV} 2023, Paris, France, October 1-6, 2023}}.
  \bibinfo{publisher}{{IEEE}}, \bibinfo{pages}{22121--22132}.
\newblock


\bibitem[Deng et~al\mbox{.}(2020)]%
        {DBLP:conf/cvpr/DengYCWT20}
\bibfield{author}{\bibinfo{person}{Yu Deng}, \bibinfo{person}{Jiaolong Yang},
  \bibinfo{person}{Dong Chen}, \bibinfo{person}{Fang Wen}, {and}
  \bibinfo{person}{Xin Tong}.} \bibinfo{year}{2020}\natexlab{}.
\newblock \showarticletitle{Disentangled and Controllable Face Image Generation
  via 3D Imitative-Contrastive Learning}. In \bibinfo{booktitle}{\emph{2020
  {IEEE/CVF} Conference on Computer Vision and Pattern Recognition, {CVPR}
  2020, Seattle, WA, USA, June 13-19, 2020}}. \bibinfo{publisher}{Computer
  Vision Foundation / {IEEE}}, \bibinfo{pages}{5153--5162}.
\newblock


\bibitem[Dennler et~al\mbox{.}(2024)]%
        {dennler2024using}
\bibfield{author}{\bibinfo{person}{Nathaniel~Steele Dennler},
  \bibinfo{person}{Stefanos Nikolaidis}, {and} \bibinfo{person}{Maja Mataric}.}
  \bibinfo{year}{2024}\natexlab{}.
\newblock \showarticletitle{Using Exploratory Search to Learn Representations
  for Human Preferences}. In \bibinfo{booktitle}{\emph{Companion of the 2024
  ACM/IEEE International Conference on Human-Robot Interaction}}.
  \bibinfo{pages}{392--396}.
\newblock


\bibitem[Dhole et~al\mbox{.}(2021)]%
        {DBLP:journals/corr/abs-2112-02721}
\bibfield{author}{\bibinfo{person}{Kaustubh~D. Dhole}, \bibinfo{person}{Varun
  Gangal}, \bibinfo{person}{Sebastian Gehrmann}, \bibinfo{person}{Aadesh
  Gupta}, \bibinfo{person}{Zhenhao Li}, \bibinfo{person}{Saad Mahamood},
  \bibinfo{person}{Abinaya Mahendiran}, \bibinfo{person}{Simon Mille},
  \bibinfo{person}{Ashish Srivastava}, \bibinfo{person}{Samson Tan},
  \bibinfo{person}{Tongshuang Wu}, \bibinfo{person}{Jascha Sohl{-}Dickstein},
  \bibinfo{person}{Jinho~D. Choi}, \bibinfo{person}{Eduard~H. Hovy},
  \bibinfo{person}{Ondrej Dusek}, \bibinfo{person}{Sebastian Ruder},
  \bibinfo{person}{Sajant Anand}, \bibinfo{person}{Nagender Aneja},
  \bibinfo{person}{Rabin Banjade}, \bibinfo{person}{Lisa Barthe},
  \bibinfo{person}{Hanna Behnke}, \bibinfo{person}{Ian Berlot{-}Attwell},
  \bibinfo{person}{Connor Boyle}, \bibinfo{person}{Caroline Brun},
  \bibinfo{person}{Marco Antonio~Sobrevilla Cabezudo}, \bibinfo{person}{Samuel
  Cahyawijaya}, \bibinfo{person}{Emile Chapuis}, \bibinfo{person}{Wanxiang
  Che}, \bibinfo{person}{Mukund Choudhary}, \bibinfo{person}{Christian Clauss},
  \bibinfo{person}{Pierre Colombo}, \bibinfo{person}{Filip Cornell},
  \bibinfo{person}{Gautier Dagan}, \bibinfo{person}{Mayukh Das},
  \bibinfo{person}{Tanay Dixit}, \bibinfo{person}{Thomas Dopierre},
  \bibinfo{person}{Paul{-}Alexis Dray}, \bibinfo{person}{Suchitra Dubey},
  \bibinfo{person}{Tatiana Ekeinhor}, \bibinfo{person}{Marco~Di Giovanni},
  \bibinfo{person}{Rishabh Gupta}, \bibinfo{person}{Rishabh Gupta},
  \bibinfo{person}{Louanes Hamla}, \bibinfo{person}{Sang Han},
  \bibinfo{person}{Fabrice Harel{-}Canada}, \bibinfo{person}{Antoine Honore},
  \bibinfo{person}{Ishan Jindal}, \bibinfo{person}{Przemyslaw~K. Joniak},
  \bibinfo{person}{Denis Kleyko}, \bibinfo{person}{Venelin Kovatchev}, {and}
  \bibinfo{person}{et al.}} \bibinfo{year}{2021}\natexlab{}.
\newblock \showarticletitle{NL-Augmenter: {A} Framework for Task-Sensitive
  Natural Language Augmentation}.
\newblock \bibinfo{journal}{\emph{CoRR}}  \bibinfo{volume}{abs/2112.02721}
  (\bibinfo{year}{2021}).
\newblock


\bibitem[Dinh et~al\mbox{.}(2016)]%
        {dinh2016density}
\bibfield{author}{\bibinfo{person}{Laurent Dinh}, \bibinfo{person}{Jascha
  Sohl-Dickstein}, {and} \bibinfo{person}{Samy Bengio}.}
  \bibinfo{year}{2016}\natexlab{}.
\newblock \showarticletitle{Density estimation using Real NVP}. In
  \bibinfo{booktitle}{\emph{International Conference on Learning
  Representations}}.
\newblock


\bibitem[Do et~al\mbox{.}(2021)]%
        {do2021clustering}
\bibfield{author}{\bibinfo{person}{Kien Do}, \bibinfo{person}{Truyen Tran},
  {and} \bibinfo{person}{Svetha Venkatesh}.} \bibinfo{year}{2021}\natexlab{}.
\newblock \showarticletitle{Clustering by maximizing mutual information across
  views}. In \bibinfo{booktitle}{\emph{Proceedings of the IEEE/CVF
  international conference on computer vision}}. \bibinfo{pages}{9928--9938}.
\newblock


\bibitem[Dong et~al\mbox{.}(2022)]%
        {DBLP:conf/cvpr/DongZWWKWLWL22}
\bibfield{author}{\bibinfo{person}{Xiao Dong}, \bibinfo{person}{Xunlin Zhan},
  \bibinfo{person}{Yangxin Wu}, \bibinfo{person}{Yunchao Wei},
  \bibinfo{person}{Michael~C. Kampffmeyer}, \bibinfo{person}{Xiaoyong Wei},
  \bibinfo{person}{Minlong Lu}, \bibinfo{person}{Yaowei Wang}, {and}
  \bibinfo{person}{Xiaodan Liang}.} \bibinfo{year}{2022}\natexlab{}.
\newblock \showarticletitle{M5Product: Self-harmonized Contrastive Learning for
  E-commercial Multi-modal Pretraining}. In
  \bibinfo{booktitle}{\emph{{IEEE/CVF} Conference on Computer Vision and
  Pattern Recognition, {CVPR} 2022, New Orleans, LA, USA, June 18-24, 2022}}.
  \bibinfo{publisher}{{IEEE}}, \bibinfo{pages}{21220--21230}.
\newblock


\bibitem[Du et~al\mbox{.}(2022)]%
        {DBLP:conf/ijcai/DuLLZ22}
\bibfield{author}{\bibinfo{person}{Yifan Du}, \bibinfo{person}{Zikang Liu},
  \bibinfo{person}{Junyi Li}, {and} \bibinfo{person}{Wayne~Xin Zhao}.}
  \bibinfo{year}{2022}\natexlab{}.
\newblock \showarticletitle{A Survey of Vision-Language Pre-Trained Models}. In
  \bibinfo{booktitle}{\emph{Proceedings of the Thirty-First International Joint
  Conference on Artificial Intelligence, {IJCAI} 2022, Vienna, Austria, 23-29
  July 2022}}. \bibinfo{publisher}{ijcai.org}, \bibinfo{pages}{5436--5443}.
\newblock


\bibitem[Duan et~al\mbox{.}(2023)]%
        {DBLP:conf/cikm/DuanFZLW23}
\bibfield{author}{\bibinfo{person}{Chenghua Duan}, \bibinfo{person}{Wei Fan},
  \bibinfo{person}{Wei Zhou}, \bibinfo{person}{Hu Liu}, {and}
  \bibinfo{person}{Junhao Wen}.} \bibinfo{year}{2023}\natexlab{}.
\newblock \showarticletitle{CLSPRec: Contrastive Learning of Long and
  Short-term Preferences for Next {POI} Recommendation}. In
  \bibinfo{booktitle}{\emph{Proceedings of the 32nd {ACM} International
  Conference on Information and Knowledge Management, {CIKM} 2023, Birmingham,
  United Kingdom, October 21-25, 2023}},
  \bibfield{editor}{\bibinfo{person}{Ingo Frommholz}, \bibinfo{person}{Frank
  Hopfgartner}, \bibinfo{person}{Mark Lee}, \bibinfo{person}{Michael Oakes},
  \bibinfo{person}{Mounia Lalmas}, \bibinfo{person}{Min Zhang}, {and}
  \bibinfo{person}{Rodrygo L.~T. Santos}} (Eds.). \bibinfo{publisher}{{ACM}},
  \bibinfo{pages}{473--482}.
\newblock


\bibitem[Dwibedi et~al\mbox{.}(2021)]%
        {DBLP:conf/iccv/DwibediATSZ21}
\bibfield{author}{\bibinfo{person}{Debidatta Dwibedi}, \bibinfo{person}{Yusuf
  Aytar}, \bibinfo{person}{Jonathan Tompson}, \bibinfo{person}{Pierre
  Sermanet}, {and} \bibinfo{person}{Andrew Zisserman}.}
  \bibinfo{year}{2021}\natexlab{}.
\newblock \showarticletitle{With a Little Help from My Friends:
  Nearest-Neighbor Contrastive Learning of Visual Representations}. In
  \bibinfo{booktitle}{\emph{2021 {IEEE/CVF} International Conference on
  Computer Vision, {ICCV} 2021, Montreal, QC, Canada, October 10-17, 2021}}.
  \bibinfo{publisher}{{IEEE}}, \bibinfo{pages}{9568--9577}.
\newblock


\bibitem[Eldele et~al\mbox{.}(2021)]%
        {DBLP:conf/ijcai/Eldele0C000G21}
\bibfield{author}{\bibinfo{person}{Emadeldeen Eldele}, \bibinfo{person}{Mohamed
  Ragab}, \bibinfo{person}{Zhenghua Chen}, \bibinfo{person}{Min Wu},
  \bibinfo{person}{Chee~Keong Kwoh}, \bibinfo{person}{Xiaoli Li}, {and}
  \bibinfo{person}{Cuntai Guan}.} \bibinfo{year}{2021}\natexlab{}.
\newblock \showarticletitle{Time-Series Representation Learning via Temporal
  and Contextual Contrasting}. In \bibinfo{booktitle}{\emph{Proceedings of the
  Thirtieth International Joint Conference on Artificial Intelligence, {IJCAI}
  2021, Virtual Event / Montreal, Canada, 19-27 August 2021}},
  \bibfield{editor}{\bibinfo{person}{Zhi{-}Hua Zhou}} (Ed.).
  \bibinfo{publisher}{ijcai.org}, \bibinfo{pages}{2352--2359}.
\newblock


\bibitem[Elizalde et~al\mbox{.}(2023)]%
        {DBLP:conf/icassp/ElizaldeDIW23}
\bibfield{author}{\bibinfo{person}{Benjamin Elizalde}, \bibinfo{person}{Soham
  Deshmukh}, \bibinfo{person}{Mahmoud~Al Ismail}, {and}
  \bibinfo{person}{Huaming Wang}.} \bibinfo{year}{2023}\natexlab{}.
\newblock \showarticletitle{{CLAP} Learning Audio Concepts from Natural
  Language Supervision}. In \bibinfo{booktitle}{\emph{{IEEE} International
  Conference on Acoustics, Speech and Signal Processing {ICASSP} 2023, Rhodes
  Island, Greece, June 4-10, 2023}}. \bibinfo{publisher}{{IEEE}},
  \bibinfo{pages}{1--5}.
\newblock


\bibitem[Eysenbach et~al\mbox{.}(2022)]%
        {eysenbach2022contrastive}
\bibfield{author}{\bibinfo{person}{Benjamin Eysenbach},
  \bibinfo{person}{Tianjun Zhang}, \bibinfo{person}{Sergey Levine}, {and}
  \bibinfo{person}{Russ~R Salakhutdinov}.} \bibinfo{year}{2022}\natexlab{}.
\newblock \showarticletitle{Contrastive learning as goal-conditioned
  reinforcement learning}.
\newblock \bibinfo{journal}{\emph{Advances in Neural Information Processing
  Systems}}  \bibinfo{volume}{35} (\bibinfo{year}{2022}),
  \bibinfo{pages}{35603--35620}.
\newblock


\bibitem[Fan et~al\mbox{.}(2023)]%
        {DBLP:conf/nips/FanKIKT23}
\bibfield{author}{\bibinfo{person}{Lijie Fan}, \bibinfo{person}{Dilip
  Krishnan}, \bibinfo{person}{Phillip Isola}, \bibinfo{person}{Dina Katabi},
  {and} \bibinfo{person}{Yonglong Tian}.} \bibinfo{year}{2023}\natexlab{}.
\newblock \showarticletitle{Improving {CLIP} Training with Language Rewrites}.
  In \bibinfo{booktitle}{\emph{Advances in Neural Information Processing
  Systems 36: Annual Conference on Neural Information Processing Systems 2023,
  NeurIPS 2023, New Orleans, LA, USA, December 10 - 16, 2023}},
  \bibfield{editor}{\bibinfo{person}{Alice Oh}, \bibinfo{person}{Tristan
  Naumann}, \bibinfo{person}{Amir Globerson}, \bibinfo{person}{Kate Saenko},
  \bibinfo{person}{Moritz Hardt}, {and} \bibinfo{person}{Sergey Levine}}
  (Eds.).
\newblock


\bibitem[Fang and Wang(2020)]%
        {fang2020time}
\bibfield{author}{\bibinfo{person}{Chenguang Fang} {and} \bibinfo{person}{Chen
  Wang}.} \bibinfo{year}{2020}\natexlab{}.
\newblock \showarticletitle{Time series data imputation: A survey on deep
  learning approaches}.
\newblock \bibinfo{journal}{\emph{arXiv preprint arXiv:2011.11347}}
  (\bibinfo{year}{2020}).
\newblock


\bibitem[Fang and Xie(2020)]%
        {DBLP:journals/corr/abs-2005-12766}
\bibfield{author}{\bibinfo{person}{Hongchao Fang} {and}
  \bibinfo{person}{Pengtao Xie}.} \bibinfo{year}{2020}\natexlab{}.
\newblock \showarticletitle{{CERT:} Contrastive Self-supervised Learning for
  Language Understanding}.
\newblock \bibinfo{journal}{\emph{CoRR}}  \bibinfo{volume}{abs/2005.12766}
  (\bibinfo{year}{2020}).
\newblock


\bibitem[Feng et~al\mbox{.}(2022)]%
        {feng2022adversarial}
\bibfield{author}{\bibinfo{person}{Shengyu Feng}, \bibinfo{person}{Baoyu Jing},
  \bibinfo{person}{Yada Zhu}, {and} \bibinfo{person}{Hanghang Tong}.}
  \bibinfo{year}{2022}\natexlab{}.
\newblock \showarticletitle{Adversarial graph contrastive learning with
  information regularization}. In \bibinfo{booktitle}{\emph{Proceedings of the
  ACM Web Conference 2022}}. \bibinfo{pages}{1362--1371}.
\newblock


\bibitem[Feng et~al\mbox{.}(2024)]%
        {feng2024ariel}
\bibfield{author}{\bibinfo{person}{Shengyu Feng}, \bibinfo{person}{Baoyu Jing},
  \bibinfo{person}{Yada Zhu}, {and} \bibinfo{person}{Hanghang Tong}.}
  \bibinfo{year}{2024}\natexlab{}.
\newblock \showarticletitle{Ariel: Adversarial graph contrastive learning}.
\newblock \bibinfo{journal}{\emph{ACM Transactions on Knowledge Discovery from
  Data}} \bibinfo{volume}{18}, \bibinfo{number}{4} (\bibinfo{year}{2024}),
  \bibinfo{pages}{1--22}.
\newblock


\bibitem[Franceschi et~al\mbox{.}(2019)]%
        {franceschi2019unsupervised}
\bibfield{author}{\bibinfo{person}{Jean-Yves Franceschi},
  \bibinfo{person}{Aymeric Dieuleveut}, {and} \bibinfo{person}{Martin Jaggi}.}
  \bibinfo{year}{2019}\natexlab{}.
\newblock \showarticletitle{Unsupervised scalable representation learning for
  multivariate time series}.
\newblock \bibinfo{journal}{\emph{Advances in neural information processing
  systems}}  \bibinfo{volume}{32} (\bibinfo{year}{2019}).
\newblock


\bibitem[Fujiwara et~al\mbox{.}(2020)]%
        {DBLP:journals/corr/abs-2005-12419}
\bibfield{author}{\bibinfo{person}{Takanori Fujiwara}, \bibinfo{person}{Jian
  Zhao}, \bibinfo{person}{Francine Chen}, \bibinfo{person}{Yaoliang Yu}, {and}
  \bibinfo{person}{Kwan{-}Liu Ma}.} \bibinfo{year}{2020}\natexlab{}.
\newblock \showarticletitle{Interpretable Contrastive Learning for Networks}.
\newblock \bibinfo{journal}{\emph{CoRR}}  \bibinfo{volume}{abs/2005.12419}
  (\bibinfo{year}{2020}).
\newblock


\bibitem[Gao et~al\mbox{.}(2021)]%
        {gao2021simcse}
\bibfield{author}{\bibinfo{person}{Tianyu Gao}, \bibinfo{person}{Xingcheng
  Yao}, {and} \bibinfo{person}{Danqi Chen}.} \bibinfo{year}{2021}\natexlab{}.
\newblock \showarticletitle{SimCSE: Simple Contrastive Learning of Sentence
  Embeddings}. In \bibinfo{booktitle}{\emph{Proceedings of the 2021 Conference
  on Empirical Methods in Natural Language Processing}}.
  \bibinfo{pages}{6894--6910}.
\newblock


\bibitem[Geng et~al\mbox{.}(2023)]%
        {DBLP:conf/iclr/GengYT0Z23}
\bibfield{author}{\bibinfo{person}{Shijie Geng}, \bibinfo{person}{Jianbo Yuan},
  \bibinfo{person}{Yu Tian}, \bibinfo{person}{Yuxiao Chen}, {and}
  \bibinfo{person}{Yongfeng Zhang}.} \bibinfo{year}{2023}\natexlab{}.
\newblock \showarticletitle{HiCLIP: Contrastive Language-Image Pretraining with
  Hierarchy-aware Attention}. In \bibinfo{booktitle}{\emph{The Eleventh
  International Conference on Learning Representations, {ICLR} 2023, Kigali,
  Rwanda, May 1-5, 2023}}. \bibinfo{publisher}{OpenReview.net}.
\newblock


\bibitem[Giorgi et~al\mbox{.}(2021)]%
        {DBLP:conf/acl/GiorgiNWB20}
\bibfield{author}{\bibinfo{person}{John~M. Giorgi}, \bibinfo{person}{Osvald
  Nitski}, \bibinfo{person}{Bo Wang}, {and} \bibinfo{person}{Gary~D. Bader}.}
  \bibinfo{year}{2021}\natexlab{}.
\newblock \showarticletitle{DeCLUTR: Deep Contrastive Learning for Unsupervised
  Textual Representations}. In \bibinfo{booktitle}{\emph{Proceedings of the
  59th Annual Meeting of the Association for Computational Linguistics and the
  11th International Joint Conference on Natural Language Processing,
  {ACL/IJCNLP} 2021, (Volume 1: Long Papers), Virtual Event, August 1-6,
  2021}}, \bibfield{editor}{\bibinfo{person}{Chengqing Zong},
  \bibinfo{person}{Fei Xia}, \bibinfo{person}{Wenjie Li}, {and}
  \bibinfo{person}{Roberto Navigli}} (Eds.). \bibinfo{publisher}{Association
  for Computational Linguistics}, \bibinfo{pages}{879--895}.
\newblock


\bibitem[Gong et~al\mbox{.}(2023)]%
        {DBLP:conf/aaai/Gong0S23}
\bibfield{author}{\bibinfo{person}{Xumeng Gong}, \bibinfo{person}{Cheng Yang},
  {and} \bibinfo{person}{Chuan Shi}.} \bibinfo{year}{2023}\natexlab{}.
\newblock \showarticletitle{{MA-GCL:} Model Augmentation Tricks for Graph
  Contrastive Learning}. In \bibinfo{booktitle}{\emph{Thirty-Seventh {AAAI}
  Conference on Artificial Intelligence, {AAAI} 2023, Thirty-Fifth Conference
  on Innovative Applications of Artificial Intelligence, {IAAI} 2023,
  Thirteenth Symposium on Educational Advances in Artificial Intelligence,
  {EAAI} 2023, Washington, DC, USA, February 7-14, 2023}},
  \bibfield{editor}{\bibinfo{person}{Brian Williams}, \bibinfo{person}{Yiling
  Chen}, {and} \bibinfo{person}{Jennifer Neville}} (Eds.).
  \bibinfo{publisher}{{AAAI} Press}, \bibinfo{pages}{4284--4292}.
\newblock


\bibitem[Grill et~al\mbox{.}(2020)]%
        {DBLP:conf/nips/GrillSATRBDPGAP20}
\bibfield{author}{\bibinfo{person}{Jean{-}Bastien Grill},
  \bibinfo{person}{Florian Strub}, \bibinfo{person}{Florent Altch{\'{e}}},
  \bibinfo{person}{Corentin Tallec}, \bibinfo{person}{Pierre~H. Richemond},
  \bibinfo{person}{Elena Buchatskaya}, \bibinfo{person}{Carl Doersch},
  \bibinfo{person}{Bernardo~{\'{A}}vila Pires}, \bibinfo{person}{Zhaohan Guo},
  \bibinfo{person}{Mohammad~Gheshlaghi Azar}, \bibinfo{person}{Bilal Piot},
  \bibinfo{person}{Koray Kavukcuoglu}, \bibinfo{person}{R{\'{e}}mi Munos},
  {and} \bibinfo{person}{Michal Valko}.} \bibinfo{year}{2020}\natexlab{}.
\newblock \showarticletitle{Bootstrap Your Own Latent - {A} New Approach to
  Self-Supervised Learning}. In \bibinfo{booktitle}{\emph{Advances in Neural
  Information Processing Systems 33: Annual Conference on Neural Information
  Processing Systems 2020, NeurIPS 2020, December 6-12, 2020, virtual}}.
\newblock


\bibitem[Gunel et~al\mbox{.}(2021)]%
        {DBLP:conf/iclr/GunelDCS21}
\bibfield{author}{\bibinfo{person}{Beliz Gunel}, \bibinfo{person}{Jingfei Du},
  \bibinfo{person}{Alexis Conneau}, {and} \bibinfo{person}{Veselin Stoyanov}.}
  \bibinfo{year}{2021}\natexlab{}.
\newblock \showarticletitle{Supervised Contrastive Learning for Pre-trained
  Language Model Fine-tuning}. In \bibinfo{booktitle}{\emph{9th International
  Conference on Learning Representations, {ICLR} 2021, Virtual Event, Austria,
  May 3-7, 2021}}. \bibinfo{publisher}{OpenReview.net}.
\newblock


\bibitem[Guo et~al\mbox{.}(2023)]%
        {DBLP:conf/acl/GuoDCLZQ0YWP23}
\bibfield{author}{\bibinfo{person}{Xinnan Guo}, \bibinfo{person}{Wentao Deng},
  \bibinfo{person}{Yongrui Chen}, \bibinfo{person}{Yang Li},
  \bibinfo{person}{Mengdi Zhou}, \bibinfo{person}{Guilin Qi},
  \bibinfo{person}{Tianxing Wu}, \bibinfo{person}{Dong Yang},
  \bibinfo{person}{Liubin Wang}, {and} \bibinfo{person}{Yong Pan}.}
  \bibinfo{year}{2023}\natexlab{}.
\newblock \showarticletitle{CoMave: Contrastive Pre-training with Multi-scale
  Masking for Attribute Value Extraction}. In
  \bibinfo{booktitle}{\emph{Findings of the Association for Computational
  Linguistics: {ACL} 2023, Toronto, Canada, July 9-14, 2023}},
  \bibfield{editor}{\bibinfo{person}{Anna Rogers}, \bibinfo{person}{Jordan~L.
  Boyd{-}Graber}, {and} \bibinfo{person}{Naoaki Okazaki}} (Eds.).
  \bibinfo{publisher}{Association for Computational Linguistics},
  \bibinfo{pages}{6007--6018}.
\newblock


\bibitem[Gutmann and Hyv{\"{a}}rinen(2010)]%
        {DBLP:journals/jmlr/GutmannH10}
\bibfield{author}{\bibinfo{person}{Michael Gutmann} {and} \bibinfo{person}{Aapo
  Hyv{\"{a}}rinen}.} \bibinfo{year}{2010}\natexlab{}.
\newblock \showarticletitle{Noise-contrastive estimation: {A} new estimation
  principle for unnormalized statistical models}. In
  \bibinfo{booktitle}{\emph{Proceedings of the Thirteenth International
  Conference on Artificial Intelligence and Statistics, {AISTATS} 2010, Chia
  Laguna Resort, Sardinia, Italy, May 13-15, 2010}}
  \emph{(\bibinfo{series}{{JMLR} Proceedings}, Vol.~\bibinfo{volume}{9})}.
  \bibinfo{publisher}{JMLR.org}, \bibinfo{pages}{297--304}.
\newblock


\bibitem[Guzhov et~al\mbox{.}(2022)]%
        {DBLP:conf/icassp/GuzhovRHD22}
\bibfield{author}{\bibinfo{person}{Andrey Guzhov}, \bibinfo{person}{Federico
  Raue}, \bibinfo{person}{J{\"{o}}rn Hees}, {and} \bibinfo{person}{Andreas
  Dengel}.} \bibinfo{year}{2022}\natexlab{}.
\newblock \showarticletitle{Audioclip: Extending Clip to Image, Text and
  Audio}. In \bibinfo{booktitle}{\emph{{IEEE} International Conference on
  Acoustics, Speech and Signal Processing, {ICASSP} 2022, Virtual and
  Singapore, 23-27 May 2022}}. \bibinfo{publisher}{{IEEE}},
  \bibinfo{pages}{976--980}.
\newblock


\bibitem[Han et~al\mbox{.}(2023)]%
        {han2023unified}
\bibfield{author}{\bibinfo{person}{Sungwon Han}, \bibinfo{person}{Mingi Shin},
  \bibinfo{person}{Sungkyu Park}, \bibinfo{person}{Changwook Jung}, {and}
  \bibinfo{person}{Meeyoung Cha}.} \bibinfo{year}{2023}\natexlab{}.
\newblock \showarticletitle{Unified neural topic model via contrastive learning
  and term weighting}. In \bibinfo{booktitle}{\emph{Proceedings of the 17th
  Conference of the European Chapter of the Association for Computational
  Linguistics}}. \bibinfo{pages}{1802--1817}.
\newblock


\bibitem[Hang et~al\mbox{.}(2022)]%
        {DBLP:conf/cvpr/HangXYL22}
\bibfield{author}{\bibinfo{person}{Yucheng Hang}, \bibinfo{person}{Bin Xia},
  \bibinfo{person}{Wenming Yang}, {and} \bibinfo{person}{Qingmin Liao}.}
  \bibinfo{year}{2022}\natexlab{}.
\newblock \showarticletitle{SCS-Co: Self-Consistent Style Contrastive Learning
  for Image Harmonization}. In \bibinfo{booktitle}{\emph{{IEEE/CVF} Conference
  on Computer Vision and Pattern Recognition, {CVPR} 2022, New Orleans, LA,
  USA, June 18-24, 2022}}. \bibinfo{publisher}{{IEEE}},
  \bibinfo{pages}{19678--19687}.
\newblock


\bibitem[Hansen et~al\mbox{.}(2003)]%
        {hansen2003reducing}
\bibfield{author}{\bibinfo{person}{Nikolaus Hansen}, \bibinfo{person}{Sibylle~D
  M{\"u}ller}, {and} \bibinfo{person}{Petros Koumoutsakos}.}
  \bibinfo{year}{2003}\natexlab{}.
\newblock \showarticletitle{Reducing the time complexity of the derandomized
  evolution strategy with covariance matrix adaptation (CMA-ES)}.
\newblock \bibinfo{journal}{\emph{Evolutionary computation}}
  \bibinfo{volume}{11}, \bibinfo{number}{1} (\bibinfo{year}{2003}),
  \bibinfo{pages}{1--18}.
\newblock


\bibitem[Hassani and Ahmadi(2020)]%
        {DBLP:conf/icml/HassaniA20}
\bibfield{author}{\bibinfo{person}{Kaveh Hassani} {and} \bibinfo{person}{Amir
  Hosein~Khas Ahmadi}.} \bibinfo{year}{2020}\natexlab{}.
\newblock \showarticletitle{Contrastive Multi-View Representation Learning on
  Graphs}. In \bibinfo{booktitle}{\emph{Proceedings of the 37th International
  Conference on Machine Learning, {ICML} 2020, 13-18 July 2020, Virtual Event}}
  \emph{(\bibinfo{series}{Proceedings of Machine Learning Research},
  Vol.~\bibinfo{volume}{119})}. \bibinfo{publisher}{{PMLR}},
  \bibinfo{pages}{4116--4126}.
\newblock


\bibitem[He et~al\mbox{.}(2023)]%
        {DBLP:conf/cvpr/00040QBSW23}
\bibfield{author}{\bibinfo{person}{Bo He}, \bibinfo{person}{Jun Wang},
  \bibinfo{person}{Jielin Qiu}, \bibinfo{person}{Trung Bui},
  \bibinfo{person}{Abhinav Shrivastava}, {and} \bibinfo{person}{Zhaowen Wang}.}
  \bibinfo{year}{2023}\natexlab{}.
\newblock \showarticletitle{Align and Attend: Multimodal Summarization with
  Dual Contrastive Losses}. In \bibinfo{booktitle}{\emph{{IEEE/CVF} Conference
  on Computer Vision and Pattern Recognition, {CVPR} 2023, Vancouver, BC,
  Canada, June 17-24, 2023}}. \bibinfo{publisher}{{IEEE}},
  \bibinfo{pages}{14867--14878}.
\newblock


\bibitem[He et~al\mbox{.}(2020)]%
        {he2020momentum}
\bibfield{author}{\bibinfo{person}{Kaiming He}, \bibinfo{person}{Haoqi Fan},
  \bibinfo{person}{Yuxin Wu}, \bibinfo{person}{Saining Xie}, {and}
  \bibinfo{person}{Ross Girshick}.} \bibinfo{year}{2020}\natexlab{}.
\newblock \showarticletitle{Momentum contrast for unsupervised visual
  representation learning}. In \bibinfo{booktitle}{\emph{Proceedings of the
  IEEE/CVF conference on computer vision and pattern recognition}}.
  \bibinfo{pages}{9729--9738}.
\newblock


\bibitem[He et~al\mbox{.}(2017)]%
        {he2017neural}
\bibfield{author}{\bibinfo{person}{Xiangnan He}, \bibinfo{person}{Lizi Liao},
  \bibinfo{person}{Hanwang Zhang}, \bibinfo{person}{Liqiang Nie},
  \bibinfo{person}{Xia Hu}, {and} \bibinfo{person}{Tat-Seng Chua}.}
  \bibinfo{year}{2017}\natexlab{}.
\newblock \showarticletitle{Neural collaborative filtering}. In
  \bibinfo{booktitle}{\emph{Proceedings of the 26th international conference on
  world wide web}}. \bibinfo{pages}{173--182}.
\newblock


\bibitem[He et~al\mbox{.}(2021)]%
        {he2021automl}
\bibfield{author}{\bibinfo{person}{Xin He}, \bibinfo{person}{Kaiyong Zhao},
  {and} \bibinfo{person}{Xiaowen Chu}.} \bibinfo{year}{2021}\natexlab{}.
\newblock \showarticletitle{AutoML: A survey of the state-of-the-art}.
\newblock \bibinfo{journal}{\emph{Knowledge-based systems}}
  \bibinfo{volume}{212} (\bibinfo{year}{2021}), \bibinfo{pages}{106622}.
\newblock


\bibitem[He et~al\mbox{.}(2019)]%
        {he2019attgan}
\bibfield{author}{\bibinfo{person}{Zhenliang He}, \bibinfo{person}{Wangmeng
  Zuo}, \bibinfo{person}{Meina Kan}, \bibinfo{person}{Shiguang Shan}, {and}
  \bibinfo{person}{Xilin Chen}.} \bibinfo{year}{2019}\natexlab{}.
\newblock \showarticletitle{Attgan: Facial attribute editing by only changing
  what you want}.
\newblock \bibinfo{journal}{\emph{IEEE transactions on image processing}}
  \bibinfo{volume}{28}, \bibinfo{number}{11} (\bibinfo{year}{2019}),
  \bibinfo{pages}{5464--5478}.
\newblock


\bibitem[Hejna et~al\mbox{.}(2023)]%
        {DBLP:journals/corr/abs-2310-13639}
\bibfield{author}{\bibinfo{person}{Joey Hejna}, \bibinfo{person}{Rafael
  Rafailov}, \bibinfo{person}{Harshit Sikchi}, \bibinfo{person}{Chelsea Finn},
  \bibinfo{person}{Scott Niekum}, \bibinfo{person}{W.~Bradley Knox}, {and}
  \bibinfo{person}{Dorsa Sadigh}.} \bibinfo{year}{2023}\natexlab{}.
\newblock \showarticletitle{Contrastive Preference Learning: Learning from
  Human Feedback without {RL}}.
\newblock \bibinfo{journal}{\emph{CoRR}}  \bibinfo{volume}{abs/2310.13639}
  (\bibinfo{year}{2023}).
\newblock


\bibitem[Hjelm et~al\mbox{.}(2019)]%
        {DBLP:conf/iclr/HjelmFLGBTB19}
\bibfield{author}{\bibinfo{person}{R.~Devon Hjelm}, \bibinfo{person}{Alex
  Fedorov}, \bibinfo{person}{Samuel Lavoie{-}Marchildon},
  \bibinfo{person}{Karan Grewal}, \bibinfo{person}{Philip Bachman},
  \bibinfo{person}{Adam Trischler}, {and} \bibinfo{person}{Yoshua Bengio}.}
  \bibinfo{year}{2019}\natexlab{}.
\newblock \showarticletitle{Learning deep representations by mutual information
  estimation and maximization}. In \bibinfo{booktitle}{\emph{7th International
  Conference on Learning Representations, {ICLR} 2019, New Orleans, LA, USA,
  May 6-9, 2019}}. \bibinfo{publisher}{OpenReview.net}.
\newblock


\bibitem[Huo et~al\mbox{.}(2021)]%
        {DBLP:journals/corr/abs-2103-06561}
\bibfield{author}{\bibinfo{person}{Yuqi Huo}, \bibinfo{person}{Manli Zhang},
  \bibinfo{person}{Guangzhen Liu}, \bibinfo{person}{Haoyu Lu},
  \bibinfo{person}{Yizhao Gao}, \bibinfo{person}{Guoxing Yang},
  \bibinfo{person}{Jingyuan Wen}, \bibinfo{person}{Heng Zhang},
  \bibinfo{person}{Baogui Xu}, \bibinfo{person}{Weihao Zheng},
  \bibinfo{person}{Zongzheng Xi}, \bibinfo{person}{Yueqian Yang},
  \bibinfo{person}{Anwen Hu}, \bibinfo{person}{Jinming Zhao},
  \bibinfo{person}{Ruichen Li}, \bibinfo{person}{Yida Zhao},
  \bibinfo{person}{Liang Zhang}, \bibinfo{person}{Yuqing Song},
  \bibinfo{person}{Xin Hong}, \bibinfo{person}{Wanqing Cui},
  \bibinfo{person}{Dan~Yang Hou}, \bibinfo{person}{Yingyan Li},
  \bibinfo{person}{Junyi Li}, \bibinfo{person}{Peiyu Liu},
  \bibinfo{person}{Zheng Gong}, \bibinfo{person}{Chuhao Jin},
  \bibinfo{person}{Yuchong Sun}, \bibinfo{person}{Shizhe Chen},
  \bibinfo{person}{Zhiwu Lu}, \bibinfo{person}{Zhicheng Dou},
  \bibinfo{person}{Qin Jin}, \bibinfo{person}{Yanyan Lan},
  \bibinfo{person}{Wayne~Xin Zhao}, \bibinfo{person}{Ruihua Song}, {and}
  \bibinfo{person}{Ji{-}Rong Wen}.} \bibinfo{year}{2021}\natexlab{}.
\newblock \showarticletitle{WenLan: Bridging Vision and Language by Large-Scale
  Multi-Modal Pre-Training}.
\newblock \bibinfo{journal}{\emph{CoRR}}  \bibinfo{volume}{abs/2103.06561}
  (\bibinfo{year}{2021}).
\newblock


\bibitem[Indurthi et~al\mbox{.}(2023)]%
        {DBLP:conf/emnlp/IndurthiCAT23}
\bibfield{author}{\bibinfo{person}{Sathish Indurthi}, \bibinfo{person}{Shamil
  Chollampatt}, \bibinfo{person}{Ravi Agrawal}, {and} \bibinfo{person}{Marco
  Turchi}.} \bibinfo{year}{2023}\natexlab{}.
\newblock \showarticletitle{{CLAD-ST:} Contrastive Learning with Adversarial
  Data for Robust Speech Translation}. In \bibinfo{booktitle}{\emph{Proceedings
  of the 2023 Conference on Empirical Methods in Natural Language Processing,
  {EMNLP} 2023, Singapore, December 6-10, 2023}},
  \bibfield{editor}{\bibinfo{person}{Houda Bouamor}, \bibinfo{person}{Juan
  Pino}, {and} \bibinfo{person}{Kalika Bali}} (Eds.).
  \bibinfo{publisher}{Association for Computational Linguistics},
  \bibinfo{pages}{9049--9056}.
\newblock


\bibitem[Ismail~Fawaz et~al\mbox{.}(2019)]%
        {ismail2019deep}
\bibfield{author}{\bibinfo{person}{Hassan Ismail~Fawaz},
  \bibinfo{person}{Germain Forestier}, \bibinfo{person}{Jonathan Weber},
  \bibinfo{person}{Lhassane Idoumghar}, {and} \bibinfo{person}{Pierre-Alain
  Muller}.} \bibinfo{year}{2019}\natexlab{}.
\newblock \showarticletitle{Deep learning for time series classification: a
  review}.
\newblock \bibinfo{journal}{\emph{Data mining and knowledge discovery}}
  \bibinfo{volume}{33}, \bibinfo{number}{4} (\bibinfo{year}{2019}),
  \bibinfo{pages}{917--963}.
\newblock


\bibitem[Jacovi et~al\mbox{.}(2021)]%
        {DBLP:conf/emnlp/JacoviSRECG21}
\bibfield{author}{\bibinfo{person}{Alon Jacovi}, \bibinfo{person}{Swabha
  Swayamdipta}, \bibinfo{person}{Shauli Ravfogel}, \bibinfo{person}{Yanai
  Elazar}, \bibinfo{person}{Yejin Choi}, {and} \bibinfo{person}{Yoav
  Goldberg}.} \bibinfo{year}{2021}\natexlab{}.
\newblock \showarticletitle{Contrastive Explanations for Model
  Interpretability}. In \bibinfo{booktitle}{\emph{Proceedings of the 2021
  Conference on Empirical Methods in Natural Language Processing, {EMNLP} 2021,
  Virtual Event / Punta Cana, Dominican Republic, 7-11 November, 2021}},
  \bibfield{editor}{\bibinfo{person}{Marie{-}Francine Moens},
  \bibinfo{person}{Xuanjing Huang}, \bibinfo{person}{Lucia Specia}, {and}
  \bibinfo{person}{Scott~Wen{-}tau Yih}} (Eds.).
  \bibinfo{publisher}{Association for Computational Linguistics},
  \bibinfo{pages}{1597--1611}.
\newblock


\bibitem[Jain et~al\mbox{.}(2023)]%
        {DBLP:conf/acl/JainZAWNLTNRBMX23}
\bibfield{author}{\bibinfo{person}{Nihal Jain}, \bibinfo{person}{Dejiao Zhang},
  \bibinfo{person}{Wasi~Uddin Ahmad}, \bibinfo{person}{Zijian Wang},
  \bibinfo{person}{Feng Nan}, \bibinfo{person}{Xiaopeng Li},
  \bibinfo{person}{Ming Tan}, \bibinfo{person}{Ramesh Nallapati},
  \bibinfo{person}{Baishakhi Ray}, \bibinfo{person}{Parminder Bhatia},
  \bibinfo{person}{Xiaofei Ma}, {and} \bibinfo{person}{Bing Xiang}.}
  \bibinfo{year}{2023}\natexlab{}.
\newblock \showarticletitle{ContraCLM: Contrastive Learning For Causal Language
  Model}. In \bibinfo{booktitle}{\emph{Proceedings of the 61st Annual Meeting
  of the Association for Computational Linguistics (Volume 1: Long Papers),
  {ACL} 2023, Toronto, Canada, July 9-14, 2023}},
  \bibfield{editor}{\bibinfo{person}{Anna Rogers}, \bibinfo{person}{Jordan~L.
  Boyd{-}Graber}, {and} \bibinfo{person}{Naoaki Okazaki}} (Eds.).
  \bibinfo{publisher}{Association for Computational Linguistics},
  \bibinfo{pages}{6436--6459}.
\newblock


\bibitem[Jaiswal et~al\mbox{.}(2020)]%
        {jaiswal2020survey}
\bibfield{author}{\bibinfo{person}{Ashish Jaiswal},
  \bibinfo{person}{Ashwin~Ramesh Babu}, \bibinfo{person}{Mohammad~Zaki Zadeh},
  \bibinfo{person}{Debapriya Banerjee}, {and} \bibinfo{person}{Fillia
  Makedon}.} \bibinfo{year}{2020}\natexlab{}.
\newblock \showarticletitle{A survey on contrastive self-supervised learning}.
\newblock \bibinfo{journal}{\emph{Technologies}} \bibinfo{volume}{9},
  \bibinfo{number}{1} (\bibinfo{year}{2020}), \bibinfo{pages}{2}.
\newblock


\bibitem[Ji et~al\mbox{.}(2021)]%
        {ji2021survey}
\bibfield{author}{\bibinfo{person}{Shaoxiong Ji}, \bibinfo{person}{Shirui Pan},
  \bibinfo{person}{Erik Cambria}, \bibinfo{person}{Pekka Marttinen}, {and}
  \bibinfo{person}{S~Yu Philip}.} \bibinfo{year}{2021}\natexlab{}.
\newblock \showarticletitle{A survey on knowledge graphs: Representation,
  acquisition, and applications}.
\newblock \bibinfo{journal}{\emph{IEEE transactions on neural networks and
  learning systems}} \bibinfo{volume}{33}, \bibinfo{number}{2}
  (\bibinfo{year}{2021}), \bibinfo{pages}{494--514}.
\newblock


\bibitem[Jia et~al\mbox{.}(2021)]%
        {DBLP:conf/icml/JiaYXCPPLSLD21}
\bibfield{author}{\bibinfo{person}{Chao Jia}, \bibinfo{person}{Yinfei Yang},
  \bibinfo{person}{Ye Xia}, \bibinfo{person}{Yi{-}Ting Chen},
  \bibinfo{person}{Zarana Parekh}, \bibinfo{person}{Hieu Pham},
  \bibinfo{person}{Quoc~V. Le}, \bibinfo{person}{Yun{-}Hsuan Sung},
  \bibinfo{person}{Zhen Li}, {and} \bibinfo{person}{Tom Duerig}.}
  \bibinfo{year}{2021}\natexlab{}.
\newblock \showarticletitle{Scaling Up Visual and Vision-Language
  Representation Learning With Noisy Text Supervision}. In
  \bibinfo{booktitle}{\emph{Proceedings of the 38th International Conference on
  Machine Learning, {ICML} 2021, 18-24 July 2021, Virtual Event}}
  \emph{(\bibinfo{series}{Proceedings of Machine Learning Research},
  Vol.~\bibinfo{volume}{139})}. \bibinfo{publisher}{{PMLR}},
  \bibinfo{pages}{4904--4916}.
\newblock


\bibitem[Jiang et~al\mbox{.}(2023)]%
        {jiang2023adaptive}
\bibfield{author}{\bibinfo{person}{Yangqin Jiang}, \bibinfo{person}{Chao
  Huang}, {and} \bibinfo{person}{Lianghao Huang}.}
  \bibinfo{year}{2023}\natexlab{}.
\newblock \showarticletitle{Adaptive graph contrastive learning for
  recommendation}. In \bibinfo{booktitle}{\emph{Proceedings of the 29th ACM
  SIGKDD conference on knowledge discovery and data mining}}.
  \bibinfo{pages}{4252--4261}.
\newblock


\bibitem[Jin et~al\mbox{.}(2023)]%
        {DBLP:journals/corr/abs-2312-02783}
\bibfield{author}{\bibinfo{person}{Bowen Jin}, \bibinfo{person}{Gang Liu},
  \bibinfo{person}{Chi Han}, \bibinfo{person}{Meng Jiang},
  \bibinfo{person}{Heng Ji}, {and} \bibinfo{person}{Jiawei Han}.}
  \bibinfo{year}{2023}\natexlab{}.
\newblock \showarticletitle{Large Language Models on Graphs: {A} Comprehensive
  Survey}.
\newblock \bibinfo{journal}{\emph{CoRR}}  \bibinfo{volume}{abs/2312.02783}
  (\bibinfo{year}{2023}).
\newblock


\bibitem[Jin et~al\mbox{.}(2021b)]%
        {DBLP:conf/ijcai/JinZL00P21}
\bibfield{author}{\bibinfo{person}{Ming Jin}, \bibinfo{person}{Yizhen Zheng},
  \bibinfo{person}{Yuan{-}Fang Li}, \bibinfo{person}{Chen Gong},
  \bibinfo{person}{Chuan Zhou}, {and} \bibinfo{person}{Shirui Pan}.}
  \bibinfo{year}{2021}\natexlab{b}.
\newblock \showarticletitle{Multi-Scale Contrastive Siamese Networks for
  Self-Supervised Graph Representation Learning}. In
  \bibinfo{booktitle}{\emph{Proceedings of the Thirtieth International Joint
  Conference on Artificial Intelligence, {IJCAI} 2021, Virtual Event /
  Montreal, Canada, 19-27 August 2021}},
  \bibfield{editor}{\bibinfo{person}{Zhi{-}Hua Zhou}} (Ed.).
  \bibinfo{publisher}{ijcai.org}, \bibinfo{pages}{1477--1483}.
\newblock


\bibitem[Jin et~al\mbox{.}(2021a)]%
        {jin2021automated}
\bibfield{author}{\bibinfo{person}{Wei Jin}, \bibinfo{person}{Xiaorui Liu},
  \bibinfo{person}{Xiangyu Zhao}, \bibinfo{person}{Yao Ma},
  \bibinfo{person}{Neil Shah}, {and} \bibinfo{person}{Jiliang Tang}.}
  \bibinfo{year}{2021}\natexlab{a}.
\newblock \showarticletitle{Automated Self-Supervised Learning for Graphs}. In
  \bibinfo{booktitle}{\emph{International Conference on Learning
  Representations}}.
\newblock


\bibitem[Jing et~al\mbox{.}(2022a)]%
        {jing2022x}
\bibfield{author}{\bibinfo{person}{Baoyu Jing}, \bibinfo{person}{Shengyu Feng},
  \bibinfo{person}{Yuejia Xiang}, \bibinfo{person}{Xi Chen},
  \bibinfo{person}{Yu Chen}, {and} \bibinfo{person}{Hanghang Tong}.}
  \bibinfo{year}{2022}\natexlab{a}.
\newblock \showarticletitle{X-GOAL: Multiplex heterogeneous graph prototypical
  contrastive learning}. In \bibinfo{booktitle}{\emph{Proceedings of the 31st
  ACM International Conference on Information \& Knowledge Management}}.
  \bibinfo{pages}{894--904}.
\newblock


\bibitem[Jing et~al\mbox{.}(2018)]%
        {jing2018cross}
\bibfield{author}{\bibinfo{person}{Baoyu Jing}, \bibinfo{person}{Chenwei Lu},
  \bibinfo{person}{Deqing Wang}, \bibinfo{person}{Fuzhen Zhuang}, {and}
  \bibinfo{person}{Cheng Niu}.} \bibinfo{year}{2018}\natexlab{}.
\newblock \showarticletitle{Cross-domain labeled LDA for cross-domain text
  classification}. In \bibinfo{booktitle}{\emph{2018 IEEE International
  Conference on Data Mining (ICDM)}}. IEEE, \bibinfo{pages}{187--196}.
\newblock


\bibitem[Jing et~al\mbox{.}(2021a)]%
        {jing2021hdmi}
\bibfield{author}{\bibinfo{person}{Baoyu Jing}, \bibinfo{person}{Chanyoung
  Park}, {and} \bibinfo{person}{Hanghang Tong}.}
  \bibinfo{year}{2021}\natexlab{a}.
\newblock \showarticletitle{Hdmi: High-order deep multiplex infomax}. In
  \bibinfo{booktitle}{\emph{Proceedings of the Web Conference 2021}}.
  \bibinfo{pages}{2414--2424}.
\newblock


\bibitem[Jing et~al\mbox{.}(2021b)]%
        {jing2021network}
\bibfield{author}{\bibinfo{person}{Baoyu Jing}, \bibinfo{person}{Hanghang
  Tong}, {and} \bibinfo{person}{Yada Zhu}.} \bibinfo{year}{2021}\natexlab{b}.
\newblock \showarticletitle{Network of tensor time series}. In
  \bibinfo{booktitle}{\emph{Proceedings of the Web Conference 2021}}.
  \bibinfo{pages}{2425--2437}.
\newblock


\bibitem[Jing et~al\mbox{.}(2024a)]%
        {jing2024automated}
\bibfield{author}{\bibinfo{person}{Baoyu Jing}, \bibinfo{person}{Yansen Wang},
  \bibinfo{person}{Guoxin Sui}, \bibinfo{person}{Jing Hong},
  \bibinfo{person}{Jingrui He}, \bibinfo{person}{Yuqing Yang},
  \bibinfo{person}{Dongsheng Li}, {and} \bibinfo{person}{Kan Ren}.}
  \bibinfo{year}{2024}\natexlab{a}.
\newblock \bibinfo{title}{Automated Contrastive Learning Strategy Search for
  Time Series}.
\newblock
\newblock
\showeprint[arxiv]{2403.12641}


\bibitem[Jing et~al\mbox{.}(2023)]%
        {jing2023sterling}
\bibfield{author}{\bibinfo{person}{Baoyu Jing}, \bibinfo{person}{Yuchen Yan},
  \bibinfo{person}{Kaize Ding}, \bibinfo{person}{Chanyoung Park},
  \bibinfo{person}{Yada Zhu}, \bibinfo{person}{Huan Liu}, {and}
  \bibinfo{person}{Hanghang Tong}.} \bibinfo{year}{2023}\natexlab{}.
\newblock \showarticletitle{Sterling: Synergistic representation learning on
  bipartite graphs}.
\newblock \bibinfo{journal}{\emph{arXiv preprint arXiv:2302.05428}}
  (\bibinfo{year}{2023}).
\newblock


\bibitem[Jing et~al\mbox{.}(2022b)]%
        {jing2022coin}
\bibfield{author}{\bibinfo{person}{Baoyu Jing}, \bibinfo{person}{Yuchen Yan},
  \bibinfo{person}{Yada Zhu}, {and} \bibinfo{person}{Hanghang Tong}.}
  \bibinfo{year}{2022}\natexlab{b}.
\newblock \showarticletitle{Coin: Co-cluster infomax for bipartite graphs}. In
  \bibinfo{booktitle}{\emph{NeurIPS 2022 Workshop: New Frontiers in Graph
  Learning}}.
\newblock


\bibitem[Jing et~al\mbox{.}(2021c)]%
        {jing2021multiplex}
\bibfield{author}{\bibinfo{person}{Baoyu Jing}, \bibinfo{person}{Zeyu You},
  \bibinfo{person}{Tao Yang}, \bibinfo{person}{Wei Fan}, {and}
  \bibinfo{person}{Hanghang Tong}.} \bibinfo{year}{2021}\natexlab{c}.
\newblock \showarticletitle{Multiplex Graph Neural Network for Extractive Text
  Summarization}. In \bibinfo{booktitle}{\emph{Proceedings of the 2021
  Conference on Empirical Methods in Natural Language Processing}}.
  \bibinfo{pages}{133--139}.
\newblock


\bibitem[Jing et~al\mbox{.}(2022c)]%
        {jing2022retrieval}
\bibfield{author}{\bibinfo{person}{Baoyu Jing}, \bibinfo{person}{Si Zhang},
  \bibinfo{person}{Yada Zhu}, \bibinfo{person}{Bin Peng},
  \bibinfo{person}{Kaiyu Guan}, \bibinfo{person}{Andrew Margenot}, {and}
  \bibinfo{person}{Hanghang Tong}.} \bibinfo{year}{2022}\natexlab{c}.
\newblock \showarticletitle{Retrieval based time series forecasting}.
\newblock \bibinfo{journal}{\emph{arXiv preprint arXiv:2209.13525}}
  (\bibinfo{year}{2022}).
\newblock


\bibitem[Jing et~al\mbox{.}(2024b)]%
        {jing2024casper}
\bibfield{author}{\bibinfo{person}{Baoyu Jing}, \bibinfo{person}{Dawei Zhou},
  \bibinfo{person}{Kan Ren}, {and} \bibinfo{person}{Carl Yang}.}
  \bibinfo{year}{2024}\natexlab{b}.
\newblock \showarticletitle{CASPER: Causality-Aware Spatiotemporal Graph Neural
  Networks for Spatiotemporal Time Series Imputation}.
\newblock \bibinfo{journal}{\emph{arXiv preprint arXiv:2403.11960}}
  (\bibinfo{year}{2024}).
\newblock


\bibitem[Kang and Park(2020)]%
        {kang2020contragan}
\bibfield{author}{\bibinfo{person}{Minguk Kang} {and} \bibinfo{person}{Jaesik
  Park}.} \bibinfo{year}{2020}\natexlab{}.
\newblock \showarticletitle{Contragan: Contrastive learning for conditional
  image generation}.
\newblock \bibinfo{journal}{\emph{Advances in Neural Information Processing
  Systems}}  \bibinfo{volume}{33} (\bibinfo{year}{2020}),
  \bibinfo{pages}{21357--21369}.
\newblock


\bibitem[Kaushik et~al\mbox{.}(2020)]%
        {DBLP:conf/iclr/KaushikHL20}
\bibfield{author}{\bibinfo{person}{Divyansh Kaushik},
  \bibinfo{person}{Eduard~H. Hovy}, {and} \bibinfo{person}{Zachary~Chase
  Lipton}.} \bibinfo{year}{2020}\natexlab{}.
\newblock \showarticletitle{Learning The Difference That Makes {A} Difference
  With Counterfactually-Augmented Data}. In \bibinfo{booktitle}{\emph{8th
  International Conference on Learning Representations, {ICLR} 2020, Addis
  Ababa, Ethiopia, April 26-30, 2020}}. \bibinfo{publisher}{OpenReview.net}.
\newblock


\bibitem[Khosla et~al\mbox{.}(2020)]%
        {abs-2004-11362}
\bibfield{author}{\bibinfo{person}{Prannay Khosla}, \bibinfo{person}{Piotr
  Teterwak}, \bibinfo{person}{Chen Wang}, \bibinfo{person}{Aaron Sarna},
  \bibinfo{person}{Yonglong Tian}, \bibinfo{person}{Phillip Isola},
  \bibinfo{person}{Aaron Maschinot}, \bibinfo{person}{Ce Liu}, {and}
  \bibinfo{person}{Dilip Krishnan}.} \bibinfo{year}{2020}\natexlab{}.
\newblock \showarticletitle{Supervised Contrastive Learning}.
\newblock \bibinfo{journal}{\emph{CoRR}}  \bibinfo{volume}{abs/2004.11362}
  (\bibinfo{year}{2020}).
\newblock


\bibitem[Kim et~al\mbox{.}(2023)]%
        {kim2023contrastive}
\bibfield{author}{\bibinfo{person}{HyunGi Kim}, \bibinfo{person}{Siwon Kim},
  \bibinfo{person}{Seonwoo Min}, {and} \bibinfo{person}{Byunghan Lee}.}
  \bibinfo{year}{2023}\natexlab{}.
\newblock \showarticletitle{Contrastive Time-Series Anomaly Detection}.
\newblock \bibinfo{journal}{\emph{IEEE Transactions on Knowledge and Data
  Engineering}} (\bibinfo{year}{2023}).
\newblock


\bibitem[Ko and Gu(2022)]%
        {DBLP:journals/corr/abs-2203-14463}
\bibfield{author}{\bibinfo{person}{ByungSoo Ko} {and} \bibinfo{person}{Geonmo
  Gu}.} \bibinfo{year}{2022}\natexlab{}.
\newblock \showarticletitle{Large-scale Bilingual Language-Image Contrastive
  Learning}.
\newblock \bibinfo{journal}{\emph{CoRR}}  \bibinfo{volume}{abs/2203.14463}
  (\bibinfo{year}{2022}).
\newblock


\bibitem[Kotek et~al\mbox{.}(2023)]%
        {kotek2023gender}
\bibfield{author}{\bibinfo{person}{Hadas Kotek}, \bibinfo{person}{Rikker
  Dockum}, {and} \bibinfo{person}{David Sun}.} \bibinfo{year}{2023}\natexlab{}.
\newblock \showarticletitle{Gender bias and stereotypes in large language
  models}. In \bibinfo{booktitle}{\emph{Proceedings of The ACM Collective
  Intelligence Conference}}. \bibinfo{pages}{12--24}.
\newblock


\bibitem[Kumar et~al\mbox{.}(2020)]%
        {kumar2020link}
\bibfield{author}{\bibinfo{person}{Ajay Kumar},
  \bibinfo{person}{Shashank~Sheshar Singh}, \bibinfo{person}{Kuldeep Singh},
  {and} \bibinfo{person}{Bhaskar Biswas}.} \bibinfo{year}{2020}\natexlab{}.
\newblock \showarticletitle{Link prediction techniques, applications, and
  performance: A survey}.
\newblock \bibinfo{journal}{\emph{Physica A: Statistical Mechanics and its
  Applications}}  \bibinfo{volume}{553} (\bibinfo{year}{2020}),
  \bibinfo{pages}{124289}.
\newblock


\bibitem[Le-Khac et~al\mbox{.}(2020)]%
        {le2020contrastive}
\bibfield{author}{\bibinfo{person}{Phuc~H Le-Khac}, \bibinfo{person}{Graham
  Healy}, {and} \bibinfo{person}{Alan~F Smeaton}.}
  \bibinfo{year}{2020}\natexlab{}.
\newblock \showarticletitle{Contrastive representation learning: A framework
  and review}.
\newblock \bibinfo{journal}{\emph{IEEE Access}}  \bibinfo{volume}{8}
  (\bibinfo{year}{2020}), \bibinfo{pages}{193907--193934}.
\newblock


\bibitem[Lee et~al\mbox{.}(2023b)]%
        {DBLP:conf/acl/LeeLZDPSZSGWALQ23}
\bibfield{author}{\bibinfo{person}{Chen{-}Yu Lee},
  \bibinfo{person}{Chun{-}Liang Li}, \bibinfo{person}{Hao Zhang},
  \bibinfo{person}{Timothy Dozat}, \bibinfo{person}{Vincent Perot},
  \bibinfo{person}{Guolong Su}, \bibinfo{person}{Xiang Zhang},
  \bibinfo{person}{Kihyuk Sohn}, \bibinfo{person}{Nikolay Glushnev},
  \bibinfo{person}{Renshen Wang}, \bibinfo{person}{Joshua Ainslie},
  \bibinfo{person}{Shangbang Long}, \bibinfo{person}{Siyang Qin},
  \bibinfo{person}{Yasuhisa Fujii}, \bibinfo{person}{Nan Hua}, {and}
  \bibinfo{person}{Tomas Pfister}.} \bibinfo{year}{2023}\natexlab{b}.
\newblock \showarticletitle{FormNetV2: Multimodal Graph Contrastive Learning
  for Form Document Information Extraction}. In
  \bibinfo{booktitle}{\emph{Proceedings of the 61st Annual Meeting of the
  Association for Computational Linguistics (Volume 1: Long Papers), {ACL}
  2023, Toronto, Canada, July 9-14, 2023}}. \bibinfo{publisher}{Association for
  Computational Linguistics}, \bibinfo{pages}{9011--9026}.
\newblock


\bibitem[Lee et~al\mbox{.}(2022)]%
        {DBLP:conf/nips/LeeKSKKLK22}
\bibfield{author}{\bibinfo{person}{Janghyeon Lee}, \bibinfo{person}{Jongsuk
  Kim}, \bibinfo{person}{Hyounguk Shon}, \bibinfo{person}{Bumsoo Kim},
  \bibinfo{person}{Seung~Hwan Kim}, \bibinfo{person}{Honglak Lee}, {and}
  \bibinfo{person}{Junmo Kim}.} \bibinfo{year}{2022}\natexlab{}.
\newblock \showarticletitle{UniCLIP: Unified Framework for Contrastive
  Language-Image Pre-training}. In \bibinfo{booktitle}{\emph{Advances in Neural
  Information Processing Systems 35: Annual Conference on Neural Information
  Processing Systems 2022, NeurIPS 2022, New Orleans, LA, USA, November 28 -
  December 9, 2022}}.
\newblock


\bibitem[Lee et~al\mbox{.}(2023a)]%
        {DBLP:conf/emnlp/0003KLZKJ23}
\bibfield{author}{\bibinfo{person}{Minwoo Lee}, \bibinfo{person}{Hyukhun Koh},
  \bibinfo{person}{Kang{-}il Lee}, \bibinfo{person}{Dongdong Zhang},
  \bibinfo{person}{Minsung Kim}, {and} \bibinfo{person}{Kyomin Jung}.}
  \bibinfo{year}{2023}\natexlab{a}.
\newblock \showarticletitle{Target-Agnostic Gender-Aware Contrastive Learning
  for Mitigating Bias in Multilingual Machine Translation}. In
  \bibinfo{booktitle}{\emph{Proceedings of the 2023 Conference on Empirical
  Methods in Natural Language Processing, {EMNLP} 2023, Singapore, December
  6-10, 2023}}, \bibfield{editor}{\bibinfo{person}{Houda Bouamor},
  \bibinfo{person}{Juan Pino}, {and} \bibinfo{person}{Kalika Bali}} (Eds.).
  \bibinfo{publisher}{Association for Computational Linguistics},
  \bibinfo{pages}{16825--16839}.
\newblock


\bibitem[Lee et~al\mbox{.}(2021)]%
        {DBLP:conf/iclr/LeeLH21}
\bibfield{author}{\bibinfo{person}{Seanie Lee}, \bibinfo{person}{Dong~Bok Lee},
  {and} \bibinfo{person}{Sung~Ju Hwang}.} \bibinfo{year}{2021}\natexlab{}.
\newblock \showarticletitle{Contrastive Learning with Adversarial Perturbations
  for Conditional Text Generation}. In \bibinfo{booktitle}{\emph{9th
  International Conference on Learning Representations, {ICLR} 2021, Virtual
  Event, Austria, May 3-7, 2021}}. \bibinfo{publisher}{OpenReview.net}.
\newblock


\bibitem[Li et~al\mbox{.}(2022a)]%
        {li2022graph}
\bibfield{author}{\bibinfo{person}{Bolian Li}, \bibinfo{person}{Baoyu Jing},
  {and} \bibinfo{person}{Hanghang Tong}.} \bibinfo{year}{2022}\natexlab{a}.
\newblock \showarticletitle{Graph communal contrastive learning}. In
  \bibinfo{booktitle}{\emph{Proceedings of the ACM web conference 2022}}.
  \bibinfo{pages}{1203--1213}.
\newblock


\bibitem[Li et~al\mbox{.}(2023b)]%
        {DBLP:conf/emnlp/LiWFO23}
\bibfield{author}{\bibinfo{person}{Dongyuan Li}, \bibinfo{person}{Yusong Wang},
  \bibinfo{person}{Kotaro Funakoshi}, {and} \bibinfo{person}{Manabu Okumura}.}
  \bibinfo{year}{2023}\natexlab{b}.
\newblock \showarticletitle{Joyful: Joint Modality Fusion and Graph Contrastive
  Learning for Multimoda Emotion Recognition}. In
  \bibinfo{booktitle}{\emph{Proceedings of the 2023 Conference on Empirical
  Methods in Natural Language Processing, {EMNLP} 2023, Singapore, December
  6-10, 2023}}. \bibinfo{publisher}{Association for Computational Linguistics},
  \bibinfo{pages}{16051--16069}.
\newblock


\bibitem[Li et~al\mbox{.}(2022f)]%
        {DBLP:journals/corr/abs-2203-01922}
\bibfield{author}{\bibinfo{person}{Feng Li}, \bibinfo{person}{Hao Zhang},
  \bibinfo{person}{Yi{-}Fan Zhang}, \bibinfo{person}{Shilong Liu},
  \bibinfo{person}{Jian Guo}, \bibinfo{person}{Lionel~M. Ni},
  \bibinfo{person}{PengChuan Zhang}, {and} \bibinfo{person}{Lei Zhang}.}
  \bibinfo{year}{2022}\natexlab{f}.
\newblock \showarticletitle{Vision-Language Intelligence: Tasks, Representation
  Learning, and Large Models}.
\newblock \bibinfo{journal}{\emph{CoRR}}  \bibinfo{volume}{abs/2203.01922}
  (\bibinfo{year}{2022}).
\newblock


\bibitem[Li et~al\mbox{.}(2021b)]%
        {DBLP:conf/nips/LiSGJXH21}
\bibfield{author}{\bibinfo{person}{Junnan Li}, \bibinfo{person}{Ramprasaath~R.
  Selvaraju}, \bibinfo{person}{Akhilesh Gotmare}, \bibinfo{person}{Shafiq~R.
  Joty}, \bibinfo{person}{Caiming Xiong}, {and}
  \bibinfo{person}{Steven~Chu{-}Hong Hoi}.} \bibinfo{year}{2021}\natexlab{b}.
\newblock \showarticletitle{Align before Fuse: Vision and Language
  Representation Learning with Momentum Distillation}. In
  \bibinfo{booktitle}{\emph{Advances in Neural Information Processing Systems
  34: Annual Conference on Neural Information Processing Systems 2021, NeurIPS
  2021, December 6-14, 2021, virtual}}. \bibinfo{pages}{9694--9705}.
\newblock


\bibitem[Li et~al\mbox{.}(2022d)]%
        {li2022uctopic}
\bibfield{author}{\bibinfo{person}{Jiacheng Li}, \bibinfo{person}{Jingbo
  Shang}, {and} \bibinfo{person}{Julian McAuley}.}
  \bibinfo{year}{2022}\natexlab{d}.
\newblock \showarticletitle{UCTopic: Unsupervised Contrastive Learning for
  Phrase Representations and Topic Mining}. In
  \bibinfo{booktitle}{\emph{Proceedings of the 60th Annual Meeting of the
  Association for Computational Linguistics (Volume 1: Long Papers)}}.
  \bibinfo{pages}{6159--6169}.
\newblock


\bibitem[Li et~al\mbox{.}(2023c)]%
        {DBLP:conf/emnlp/LiZDM0Q23}
\bibfield{author}{\bibinfo{person}{Jiaqi Li}, \bibinfo{person}{Chuanyi Zhang},
  \bibinfo{person}{Miaozeng Du}, \bibinfo{person}{Dehai Min},
  \bibinfo{person}{Yongrui Chen}, {and} \bibinfo{person}{Guilin Qi}.}
  \bibinfo{year}{2023}\natexlab{c}.
\newblock \showarticletitle{Three Stream Based Multi-level Event Contrastive
  Learning for Text-Video Event Extraction}. In
  \bibinfo{booktitle}{\emph{Proceedings of the 2023 Conference on Empirical
  Methods in Natural Language Processing, {EMNLP} 2023, Singapore, December
  6-10, 2023}}, \bibfield{editor}{\bibinfo{person}{Houda Bouamor},
  \bibinfo{person}{Juan Pino}, {and} \bibinfo{person}{Kalika Bali}} (Eds.).
  \bibinfo{publisher}{Association for Computational Linguistics},
  \bibinfo{pages}{1666--1676}.
\newblock


\bibitem[Li et~al\mbox{.}(2020b)]%
        {li2020prototypical}
\bibfield{author}{\bibinfo{person}{Junnan Li}, \bibinfo{person}{Pan Zhou},
  \bibinfo{person}{Caiming Xiong}, {and} \bibinfo{person}{Steven Hoi}.}
  \bibinfo{year}{2020}\natexlab{b}.
\newblock \showarticletitle{Prototypical Contrastive Learning of Unsupervised
  Representations}. In \bibinfo{booktitle}{\emph{International Conference on
  Learning Representations}}.
\newblock


\bibitem[Li et~al\mbox{.}(2022e)]%
        {li2022selective}
\bibfield{author}{\bibinfo{person}{Shikun Li}, \bibinfo{person}{Xiaobo Xia},
  \bibinfo{person}{Shiming Ge}, {and} \bibinfo{person}{Tongliang Liu}.}
  \bibinfo{year}{2022}\natexlab{e}.
\newblock \showarticletitle{Selective-supervised contrastive learning with
  noisy labels}. In \bibinfo{booktitle}{\emph{Proceedings of the IEEE/CVF
  conference on computer vision and pattern recognition}}.
  \bibinfo{pages}{316--325}.
\newblock


\bibitem[Li et~al\mbox{.}(2023a)]%
        {DBLP:conf/emnlp/0001DL23}
\bibfield{author}{\bibinfo{person}{Yichuan Li}, \bibinfo{person}{Kaize Ding},
  {and} \bibinfo{person}{Kyumin Lee}.} \bibinfo{year}{2023}\natexlab{a}.
\newblock \showarticletitle{{GRENADE:} Graph-Centric Language Model for
  Self-Supervised Representation Learning on Text-Attributed Graphs}. In
  \bibinfo{booktitle}{\emph{Findings of the Association for Computational
  Linguistics: {EMNLP} 2023, Singapore, December 6-10, 2023}},
  \bibfield{editor}{\bibinfo{person}{Houda Bouamor}, \bibinfo{person}{Juan
  Pino}, {and} \bibinfo{person}{Kalika Bali}} (Eds.).
  \bibinfo{publisher}{Association for Computational Linguistics},
  \bibinfo{pages}{2745--2757}.
\newblock


\bibitem[Li et~al\mbox{.}(2020a)]%
        {li2020differentiable}
\bibfield{author}{\bibinfo{person}{Yonggang Li}, \bibinfo{person}{Guosheng Hu},
  \bibinfo{person}{Yongtao Wang}, \bibinfo{person}{Timothy Hospedales},
  \bibinfo{person}{Neil~M Robertson}, {and} \bibinfo{person}{Yongxin Yang}.}
  \bibinfo{year}{2020}\natexlab{a}.
\newblock \showarticletitle{Differentiable automatic data augmentation}. In
  \bibinfo{booktitle}{\emph{Computer Vision--ECCV 2020: 16th European
  Conference, Glasgow, UK, August 23--28, 2020, Proceedings, Part XXII 16}}.
  Springer, \bibinfo{pages}{580--595}.
\newblock


\bibitem[Li et~al\mbox{.}(2021a)]%
        {DBLP:conf/aaai/Li0LPZ021}
\bibfield{author}{\bibinfo{person}{Yunfan Li}, \bibinfo{person}{Peng Hu},
  \bibinfo{person}{Jerry~Zitao Liu}, \bibinfo{person}{Dezhong Peng},
  \bibinfo{person}{Joey~Tianyi Zhou}, {and} \bibinfo{person}{Xi Peng}.}
  \bibinfo{year}{2021}\natexlab{a}.
\newblock \showarticletitle{Contrastive Clustering}. In
  \bibinfo{booktitle}{\emph{Thirty-Fifth {AAAI} Conference on Artificial
  Intelligence, {AAAI} 2021, Thirty-Third Conference on Innovative Applications
  of Artificial Intelligence, {IAAI} 2021, The Eleventh Symposium on
  Educational Advances in Artificial Intelligence, {EAAI} 2021, Virtual Event,
  February 2-9, 2021}}. \bibinfo{publisher}{{AAAI} Press},
  \bibinfo{pages}{8547--8555}.
\newblock


\bibitem[Li et~al\mbox{.}(2022b)]%
        {DBLP:conf/iclr/LiLZCOSYY22}
\bibfield{author}{\bibinfo{person}{Yangguang Li}, \bibinfo{person}{Feng Liang},
  \bibinfo{person}{Lichen Zhao}, \bibinfo{person}{Yufeng Cui},
  \bibinfo{person}{Wanli Ouyang}, \bibinfo{person}{Jing Shao},
  \bibinfo{person}{Fengwei Yu}, {and} \bibinfo{person}{Junjie Yan}.}
  \bibinfo{year}{2022}\natexlab{b}.
\newblock \showarticletitle{Supervision Exists Everywhere: {A} Data Efficient
  Contrastive Language-Image Pre-training Paradigm}. In
  \bibinfo{booktitle}{\emph{The Tenth International Conference on Learning
  Representations, {ICLR} 2022, Virtual Event, April 25-29, 2022}}.
  \bibinfo{publisher}{OpenReview.net}.
\newblock


\bibitem[Li et~al\mbox{.}(2022c)]%
        {DBLP:conf/acl/Li0CKV22}
\bibfield{author}{\bibinfo{person}{Yaoyiran Li}, \bibinfo{person}{Fangyu Liu},
  \bibinfo{person}{Nigel Collier}, \bibinfo{person}{Anna Korhonen}, {and}
  \bibinfo{person}{Ivan Vulic}.} \bibinfo{year}{2022}\natexlab{c}.
\newblock \showarticletitle{Improving Word Translation via Two-Stage
  Contrastive Learning}. In \bibinfo{booktitle}{\emph{Proceedings of the 60th
  Annual Meeting of the Association for Computational Linguistics (Volume 1:
  Long Papers), {ACL} 2022, Dublin, Ireland, May 22-27, 2022}},
  \bibfield{editor}{\bibinfo{person}{Smaranda Muresan},
  \bibinfo{person}{Preslav Nakov}, {and} \bibinfo{person}{Aline Villavicencio}}
  (Eds.). \bibinfo{publisher}{Association for Computational Linguistics},
  \bibinfo{pages}{4353--4374}.
\newblock


\bibitem[Li et~al\mbox{.}(2018)]%
        {li2018survey}
\bibfield{author}{\bibinfo{person}{Yingming Li}, \bibinfo{person}{Ming Yang},
  {and} \bibinfo{person}{Zhongfei Zhang}.} \bibinfo{year}{2018}\natexlab{}.
\newblock \showarticletitle{A survey of multi-view representation learning}.
\newblock \bibinfo{journal}{\emph{IEEE transactions on knowledge and data
  engineering}} \bibinfo{volume}{31}, \bibinfo{number}{10}
  (\bibinfo{year}{2018}), \bibinfo{pages}{1863--1883}.
\newblock


\bibitem[Liang et~al\mbox{.}(2023b)]%
        {DBLP:journals/nn/LiangDZM0G23}
\bibfield{author}{\bibinfo{person}{Huidong Liang}, \bibinfo{person}{Xingjian
  Du}, \bibinfo{person}{Bilei Zhu}, \bibinfo{person}{Zejun Ma},
  \bibinfo{person}{Ke Chen}, {and} \bibinfo{person}{Junbin Gao}.}
  \bibinfo{year}{2023}\natexlab{b}.
\newblock \showarticletitle{Graph contrastive learning with implicit
  augmentations}.
\newblock \bibinfo{journal}{\emph{Neural Networks}}  \bibinfo{volume}{163}
  (\bibinfo{year}{2023}), \bibinfo{pages}{156--164}.
\newblock


\bibitem[Liang et~al\mbox{.}(2023a)]%
        {DBLP:conf/nips/LiangDMZMS23}
\bibfield{author}{\bibinfo{person}{Paul~Pu Liang}, \bibinfo{person}{Zihao
  Deng}, \bibinfo{person}{Martin~Q. Ma}, \bibinfo{person}{James~Y. Zou},
  \bibinfo{person}{Louis{-}Philippe Morency}, {and} \bibinfo{person}{Ruslan
  Salakhutdinov}.} \bibinfo{year}{2023}\natexlab{a}.
\newblock \showarticletitle{Factorized Contrastive Learning: Going Beyond
  Multi-view Redundancy}. In \bibinfo{booktitle}{\emph{Advances in Neural
  Information Processing Systems 36: Annual Conference on Neural Information
  Processing Systems 2023, NeurIPS 2023, New Orleans, LA, USA, December 10 -
  16, 2023}}, \bibfield{editor}{\bibinfo{person}{Alice Oh},
  \bibinfo{person}{Tristan Naumann}, \bibinfo{person}{Amir Globerson},
  \bibinfo{person}{Kate Saenko}, \bibinfo{person}{Moritz Hardt}, {and}
  \bibinfo{person}{Sergey Levine}} (Eds.).
\newblock


\bibitem[Lim and Zohren(2021)]%
        {lim2021time}
\bibfield{author}{\bibinfo{person}{Bryan Lim} {and} \bibinfo{person}{Stefan
  Zohren}.} \bibinfo{year}{2021}\natexlab{}.
\newblock \showarticletitle{Time-series forecasting with deep learning: a
  survey}.
\newblock \bibinfo{journal}{\emph{Philosophical Transactions of the Royal
  Society A}} \bibinfo{volume}{379}, \bibinfo{number}{2194}
  (\bibinfo{year}{2021}), \bibinfo{pages}{20200209}.
\newblock


\bibitem[Lim et~al\mbox{.}(2019)]%
        {lim2019fast}
\bibfield{author}{\bibinfo{person}{Sungbin Lim}, \bibinfo{person}{Ildoo Kim},
  \bibinfo{person}{Taesup Kim}, \bibinfo{person}{Chiheon Kim}, {and}
  \bibinfo{person}{Sungwoong Kim}.} \bibinfo{year}{2019}\natexlab{}.
\newblock \showarticletitle{Fast autoaugment}.
\newblock \bibinfo{journal}{\emph{Advances in Neural Information Processing
  Systems}}  \bibinfo{volume}{32} (\bibinfo{year}{2019}).
\newblock


\bibitem[Lin et~al\mbox{.}(2022)]%
        {DBLP:conf/ijcai/LinBB0ZX22}
\bibfield{author}{\bibinfo{person}{Fangfei Lin}, \bibinfo{person}{Bing Bai},
  \bibinfo{person}{Kun Bai}, \bibinfo{person}{Yazhou Ren},
  \bibinfo{person}{Peng Zhao}, {and} \bibinfo{person}{Zenglin Xu}.}
  \bibinfo{year}{2022}\natexlab{}.
\newblock \showarticletitle{Contrastive Multi-view Hyperbolic Hierarchical
  Clustering}. In \bibinfo{booktitle}{\emph{Proceedings of the Thirty-First
  International Joint Conference on Artificial Intelligence, {IJCAI} 2022,
  Vienna, Austria, 23-29 July 2022}}. \bibinfo{publisher}{ijcai.org},
  \bibinfo{pages}{3250--3256}.
\newblock


\bibitem[Lin and Hu(2022)]%
        {DBLP:conf/emnlp/Lin022}
\bibfield{author}{\bibinfo{person}{Ronghao Lin} {and} \bibinfo{person}{Haifeng
  Hu}.} \bibinfo{year}{2022}\natexlab{}.
\newblock \showarticletitle{Multimodal Contrastive Learning via Uni-Modal
  Coding and Cross-Modal Prediction for Multimodal Sentiment Analysis}. In
  \bibinfo{booktitle}{\emph{Findings of the Association for Computational
  Linguistics: {EMNLP} 2022, Abu Dhabi, United Arab Emirates, December 7-11,
  2022}}. \bibinfo{publisher}{Association for Computational Linguistics},
  \bibinfo{pages}{511--523}.
\newblock


\bibitem[Lin et~al\mbox{.}(2021)]%
        {DBLP:conf/cvpr/0001GLL0021}
\bibfield{author}{\bibinfo{person}{Yijie Lin}, \bibinfo{person}{Yuanbiao Gou},
  \bibinfo{person}{Zitao Liu}, \bibinfo{person}{Boyun Li},
  \bibinfo{person}{Jiancheng Lv}, {and} \bibinfo{person}{Xi Peng}.}
  \bibinfo{year}{2021}\natexlab{}.
\newblock \showarticletitle{{COMPLETER:} Incomplete Multi-View Clustering via
  Contrastive Prediction}. In \bibinfo{booktitle}{\emph{{IEEE} Conference on
  Computer Vision and Pattern Recognition, {CVPR} 2021, virtual, June 19-25,
  2021}}. \bibinfo{publisher}{Computer Vision Foundation / {IEEE}},
  \bibinfo{pages}{11174--11183}.
\newblock


\bibitem[Liu and Chen(2023)]%
        {liu2023timesurl}
\bibfield{author}{\bibinfo{person}{Jiexi Liu} {and} \bibinfo{person}{Songcan
  Chen}.} \bibinfo{year}{2023}\natexlab{}.
\newblock \showarticletitle{TimesURL: Self-supervised Contrastive Learning for
  Universal Time Series Representation Learning}.
\newblock \bibinfo{journal}{\emph{arXiv preprint arXiv:2312.15709}}
  (\bibinfo{year}{2023}).
\newblock


\bibitem[Liu et~al\mbox{.}(2023e)]%
        {DBLP:journals/corr/abs-2310-11829}
\bibfield{author}{\bibinfo{person}{Jiawei Liu}, \bibinfo{person}{Cheng Yang},
  \bibinfo{person}{Zhiyuan Lu}, \bibinfo{person}{Junze Chen},
  \bibinfo{person}{Yibo Li}, \bibinfo{person}{Mengmei Zhang},
  \bibinfo{person}{Ting Bai}, \bibinfo{person}{Yuan Fang},
  \bibinfo{person}{Lichao Sun}, \bibinfo{person}{Philip~S. Yu}, {and}
  \bibinfo{person}{Chuan Shi}.} \bibinfo{year}{2023}\natexlab{e}.
\newblock \showarticletitle{Towards Graph Foundation Models: {A} Survey and
  Beyond}.
\newblock \bibinfo{journal}{\emph{CoRR}}  \bibinfo{volume}{abs/2310.11829}
  (\bibinfo{year}{2023}).
\newblock


\bibitem[Liu et~al\mbox{.}(2023c)]%
        {DBLP:journals/corr/abs-2308-06911}
\bibfield{author}{\bibinfo{person}{Pengfei Liu}, \bibinfo{person}{Yiming Ren},
  {and} \bibinfo{person}{Zhixiang Ren}.} \bibinfo{year}{2023}\natexlab{c}.
\newblock \showarticletitle{GIT-Mol: {A} Multi-modal Large Language Model for
  Molecular Science with Graph, Image, and Text}.
\newblock \bibinfo{journal}{\emph{CoRR}}  \bibinfo{volume}{abs/2308.06911}
  (\bibinfo{year}{2023}).
\newblock


\bibitem[Liu et~al\mbox{.}(2023d)]%
        {DBLP:journals/corr/abs-2310-18339}
\bibfield{author}{\bibinfo{person}{Qidong Liu}, \bibinfo{person}{Xian Wu},
  \bibinfo{person}{Xiangyu Zhao}, \bibinfo{person}{Yuanshao Zhu},
  \bibinfo{person}{Derong Xu}, \bibinfo{person}{Feng Tian}, {and}
  \bibinfo{person}{Yefeng Zheng}.} \bibinfo{year}{2023}\natexlab{d}.
\newblock \showarticletitle{MOELoRA: An MOE-based Parameter Efficient
  Fine-Tuning Method for Multi-task Medical Applications}.
\newblock \bibinfo{journal}{\emph{CoRR}}  \bibinfo{volume}{abs/2310.18339}
  (\bibinfo{year}{2023}).
\newblock


\bibitem[Liu et~al\mbox{.}(2023b)]%
        {DBLP:journals/natmi/LiuNWLQLTXA23}
\bibfield{author}{\bibinfo{person}{Shengchao Liu}, \bibinfo{person}{Weili Nie},
  \bibinfo{person}{Chengpeng Wang}, \bibinfo{person}{Jiarui Lu},
  \bibinfo{person}{Zhuoran Qiao}, \bibinfo{person}{Ling Liu},
  \bibinfo{person}{Jian Tang}, \bibinfo{person}{Chaowei Xiao}, {and}
  \bibinfo{person}{Animashree Anandkumar}.} \bibinfo{year}{2023}\natexlab{b}.
\newblock \showarticletitle{Multi-modal molecule structure-text model for
  text-based retrieval and editing}.
\newblock \bibinfo{journal}{\emph{Nat. Mac. Intell.}} \bibinfo{volume}{5},
  \bibinfo{number}{12} (\bibinfo{year}{2023}), \bibinfo{pages}{1447--1457}.
\newblock


\bibitem[Liu et~al\mbox{.}(2023g)]%
        {DBLP:journals/tkde/LiuZHMWZT23}
\bibfield{author}{\bibinfo{person}{Xiao Liu}, \bibinfo{person}{Fanjin Zhang},
  \bibinfo{person}{Zhenyu Hou}, \bibinfo{person}{Li Mian},
  \bibinfo{person}{Zhaoyu Wang}, \bibinfo{person}{Jing Zhang}, {and}
  \bibinfo{person}{Jie Tang}.} \bibinfo{year}{2023}\natexlab{g}.
\newblock \showarticletitle{Self-Supervised Learning: Generative or
  Contrastive}.
\newblock \bibinfo{journal}{\emph{{IEEE} Trans. Knowl. Data Eng.}}
  \bibinfo{volume}{35}, \bibinfo{number}{1} (\bibinfo{year}{2023}),
  \bibinfo{pages}{857--876}.
\newblock


\bibitem[Liu et~al\mbox{.}(2021)]%
        {liu2021anomaly}
\bibfield{author}{\bibinfo{person}{Yixin Liu}, \bibinfo{person}{Zhao Li},
  \bibinfo{person}{Shirui Pan}, \bibinfo{person}{Chen Gong},
  \bibinfo{person}{Chuan Zhou}, {and} \bibinfo{person}{George Karypis}.}
  \bibinfo{year}{2021}\natexlab{}.
\newblock \showarticletitle{Anomaly detection on attributed networks via
  contrastive self-supervised learning}.
\newblock \bibinfo{journal}{\emph{IEEE transactions on neural networks and
  learning systems}} \bibinfo{volume}{33}, \bibinfo{number}{6}
  (\bibinfo{year}{2021}), \bibinfo{pages}{2378--2392}.
\newblock


\bibitem[Liu and Liu(2021)]%
        {liu2021simcls}
\bibfield{author}{\bibinfo{person}{Yixin Liu} {and} \bibinfo{person}{Pengfei
  Liu}.} \bibinfo{year}{2021}\natexlab{}.
\newblock \showarticletitle{SimCLS: A Simple Framework for Contrastive Learning
  of Abstractive Summarization}. In \bibinfo{booktitle}{\emph{Proceedings of
  the 59th Annual Meeting of the Association for Computational Linguistics and
  the 11th International Joint Conference on Natural Language Processing
  (Volume 2: Short Papers)}}. \bibinfo{pages}{1065--1072}.
\newblock


\bibitem[Liu et~al\mbox{.}(2023f)]%
        {liu2023simple}
\bibfield{author}{\bibinfo{person}{Yue Liu}, \bibinfo{person}{Xihong Yang},
  \bibinfo{person}{Sihang Zhou}, \bibinfo{person}{Xinwang Liu},
  \bibinfo{person}{Siwei Wang}, \bibinfo{person}{Ke Liang},
  \bibinfo{person}{Wenxuan Tu}, {and} \bibinfo{person}{Liang Li}.}
  \bibinfo{year}{2023}\natexlab{f}.
\newblock \showarticletitle{Simple contrastive graph clustering}.
\newblock \bibinfo{journal}{\emph{IEEE Transactions on Neural Networks and
  Learning Systems}} (\bibinfo{year}{2023}).
\newblock


\bibitem[Liu et~al\mbox{.}(2023a)]%
        {DBLP:conf/emnlp/LiuLL00K0C23}
\bibfield{author}{\bibinfo{person}{Zhiyuan Liu}, \bibinfo{person}{Sihang Li},
  \bibinfo{person}{Yanchen Luo}, \bibinfo{person}{Hao Fei},
  \bibinfo{person}{Yixin Cao}, \bibinfo{person}{Kenji Kawaguchi},
  \bibinfo{person}{Xiang Wang}, {and} \bibinfo{person}{Tat{-}Seng Chua}.}
  \bibinfo{year}{2023}\natexlab{a}.
\newblock \showarticletitle{MolCA: Molecular Graph-Language Modeling with
  Cross-Modal Projector and Uni-Modal Adapter}. In
  \bibinfo{booktitle}{\emph{Proceedings of the 2023 Conference on Empirical
  Methods in Natural Language Processing, {EMNLP} 2023, Singapore, December
  6-10, 2023}}, \bibfield{editor}{\bibinfo{person}{Houda Bouamor},
  \bibinfo{person}{Juan Pino}, {and} \bibinfo{person}{Kalika Bali}} (Eds.).
  \bibinfo{publisher}{Association for Computational Linguistics},
  \bibinfo{pages}{15623--15638}.
\newblock


\bibitem[Logeswaran and Lee(2018)]%
        {DBLP:conf/iclr/LogeswaranL18}
\bibfield{author}{\bibinfo{person}{Lajanugen Logeswaran} {and}
  \bibinfo{person}{Honglak Lee}.} \bibinfo{year}{2018}\natexlab{}.
\newblock \showarticletitle{An efficient framework for learning sentence
  representations}. In \bibinfo{booktitle}{\emph{6th International Conference
  on Learning Representations, {ICLR} 2018, Vancouver, BC, Canada, April 30 -
  May 3, 2018, Conference Track Proceedings}}.
  \bibinfo{publisher}{OpenReview.net}.
\newblock


\bibitem[Luo et~al\mbox{.}(2023a)]%
        {luo2023time}
\bibfield{author}{\bibinfo{person}{Dongsheng Luo}, \bibinfo{person}{Wei Cheng},
  \bibinfo{person}{Yingheng Wang}, \bibinfo{person}{Dongkuan Xu},
  \bibinfo{person}{Jingchao Ni}, \bibinfo{person}{Wenchao Yu},
  \bibinfo{person}{Xuchao Zhang}, \bibinfo{person}{Yanchi Liu},
  \bibinfo{person}{Yuncong Chen}, \bibinfo{person}{Haifeng Chen},
  {et~al\mbox{.}}} \bibinfo{year}{2023}\natexlab{a}.
\newblock \showarticletitle{Time series contrastive learning with
  information-aware augmentations}. In \bibinfo{booktitle}{\emph{Proceedings of
  the AAAI Conference on Artificial Intelligence}}, Vol.~\bibinfo{volume}{37}.
  \bibinfo{pages}{4534--4542}.
\newblock


\bibitem[Luo et~al\mbox{.}(2022)]%
        {luo2022dualgraph}
\bibfield{author}{\bibinfo{person}{Xiao Luo}, \bibinfo{person}{Wei Ju},
  \bibinfo{person}{Meng Qu}, \bibinfo{person}{Chong Chen},
  \bibinfo{person}{Minghua Deng}, \bibinfo{person}{Xian-Sheng Hua}, {and}
  \bibinfo{person}{Ming Zhang}.} \bibinfo{year}{2022}\natexlab{}.
\newblock \showarticletitle{Dualgraph: Improving semi-supervised graph
  classification via dual contrastive learning}. In
  \bibinfo{booktitle}{\emph{2022 IEEE 38th International Conference on Data
  Engineering (ICDE)}}. IEEE, \bibinfo{pages}{699--712}.
\newblock


\bibitem[Luo et~al\mbox{.}(2018)]%
        {luo2018multivariate}
\bibfield{author}{\bibinfo{person}{Yonghong Luo}, \bibinfo{person}{Xiangrui
  Cai}, \bibinfo{person}{Ying Zhang}, \bibinfo{person}{Jun Xu},
  {et~al\mbox{.}}} \bibinfo{year}{2018}\natexlab{}.
\newblock \showarticletitle{Multivariate time series imputation with generative
  adversarial networks}.
\newblock \bibinfo{journal}{\emph{Advances in neural information processing
  systems}}  \bibinfo{volume}{31} (\bibinfo{year}{2018}).
\newblock


\bibitem[Luo et~al\mbox{.}(2023b)]%
        {DBLP:journals/corr/abs-2307-09484}
\bibfield{author}{\bibinfo{person}{Yizhen Luo}, \bibinfo{person}{Kai Yang},
  \bibinfo{person}{Massimo Hong}, \bibinfo{person}{Xing~Yi Liu}, {and}
  \bibinfo{person}{Zaiqing Nie}.} \bibinfo{year}{2023}\natexlab{b}.
\newblock \showarticletitle{MolFM: {A} Multimodal Molecular Foundation Model}.
\newblock \bibinfo{journal}{\emph{CoRR}}  \bibinfo{volume}{abs/2307.09484}
  (\bibinfo{year}{2023}).
\newblock


\bibitem[Ma et~al\mbox{.}(2021a)]%
        {DBLP:conf/acl/MaSA21}
\bibfield{author}{\bibinfo{person}{Xiaofei Ma},
  \bibinfo{person}{C{\'{\i}}cero~Nogueira dos Santos}, {and}
  \bibinfo{person}{Andrew~O. Arnold}.} \bibinfo{year}{2021}\natexlab{a}.
\newblock \showarticletitle{Contrastive Fine-tuning Improves Robustness for
  Neural Rankers}. In \bibinfo{booktitle}{\emph{Findings of the Association for
  Computational Linguistics: {ACL/IJCNLP} 2021, Online Event, August 1-6,
  2021}} \emph{(\bibinfo{series}{Findings of {ACL}},
  Vol.~\bibinfo{volume}{{ACL/IJCNLP} 2021})},
  \bibfield{editor}{\bibinfo{person}{Chengqing Zong}, \bibinfo{person}{Fei
  Xia}, \bibinfo{person}{Wenjie Li}, {and} \bibinfo{person}{Roberto Navigli}}
  (Eds.). \bibinfo{publisher}{Association for Computational Linguistics},
  \bibinfo{pages}{570--582}.
\newblock


\bibitem[Ma et~al\mbox{.}(2021b)]%
        {ma2021comprehensive}
\bibfield{author}{\bibinfo{person}{Xiaoxiao Ma}, \bibinfo{person}{Jia Wu},
  \bibinfo{person}{Shan Xue}, \bibinfo{person}{Jian Yang},
  \bibinfo{person}{Chuan Zhou}, \bibinfo{person}{Quan~Z Sheng},
  \bibinfo{person}{Hui Xiong}, {and} \bibinfo{person}{Leman Akoglu}.}
  \bibinfo{year}{2021}\natexlab{b}.
\newblock \showarticletitle{A comprehensive survey on graph anomaly detection
  with deep learning}.
\newblock \bibinfo{journal}{\emph{IEEE Transactions on Knowledge and Data
  Engineering}} \bibinfo{volume}{35}, \bibinfo{number}{12}
  (\bibinfo{year}{2021}), \bibinfo{pages}{12012--12038}.
\newblock


\bibitem[Malladi et~al\mbox{.}(2024)]%
        {malladi2024fine}
\bibfield{author}{\bibinfo{person}{Sadhika Malladi}, \bibinfo{person}{Tianyu
  Gao}, \bibinfo{person}{Eshaan Nichani}, \bibinfo{person}{Alex Damian},
  \bibinfo{person}{Jason~D Lee}, \bibinfo{person}{Danqi Chen}, {and}
  \bibinfo{person}{Sanjeev Arora}.} \bibinfo{year}{2024}\natexlab{}.
\newblock \showarticletitle{Fine-tuning language models with just forward
  passes}.
\newblock \bibinfo{journal}{\emph{Advances in Neural Information Processing
  Systems}}  \bibinfo{volume}{36} (\bibinfo{year}{2024}).
\newblock


\bibitem[Mao et~al\mbox{.}(2024)]%
        {DBLP:journals/corr/abs-2402-02216}
\bibfield{author}{\bibinfo{person}{Haitao Mao}, \bibinfo{person}{Zhikai Chen},
  \bibinfo{person}{Wenzhuo Tang}, \bibinfo{person}{Jianan Zhao},
  \bibinfo{person}{Yao Ma}, \bibinfo{person}{Tong Zhao}, \bibinfo{person}{Neil
  Shah}, \bibinfo{person}{Mikhail Galkin}, {and} \bibinfo{person}{Jiliang
  Tang}.} \bibinfo{year}{2024}\natexlab{}.
\newblock \showarticletitle{Graph Foundation Models}.
\newblock \bibinfo{journal}{\emph{CoRR}}  \bibinfo{volume}{abs/2402.02216}
  (\bibinfo{year}{2024}).
\newblock


\bibitem[Mao et~al\mbox{.}(2022)]%
        {DBLP:conf/naacl/MaoCDSWK22}
\bibfield{author}{\bibinfo{person}{Zhuoyuan Mao}, \bibinfo{person}{Chenhui
  Chu}, \bibinfo{person}{Raj Dabre}, \bibinfo{person}{Haiyue Song},
  \bibinfo{person}{Zhen Wan}, {and} \bibinfo{person}{Sadao Kurohashi}.}
  \bibinfo{year}{2022}\natexlab{}.
\newblock \showarticletitle{When do Contrastive Word Alignments Improve
  Many-to-many Neural Machine Translation?}. In
  \bibinfo{booktitle}{\emph{Findings of the Association for Computational
  Linguistics: {NAACL} 2022, Seattle, WA, United States, July 10-15, 2022}},
  \bibfield{editor}{\bibinfo{person}{Marine Carpuat},
  \bibinfo{person}{Marie{-}Catherine de~Marneffe}, {and}
  \bibinfo{person}{Iv{\'{a}}n Vladimir~Meza Ru{\'{\i}}z}} (Eds.).
  \bibinfo{publisher}{Association for Computational Linguistics},
  \bibinfo{pages}{1766--1775}.
\newblock


\bibitem[Meng et~al\mbox{.}(2021)]%
        {DBLP:conf/nips/MengXBTBHS21}
\bibfield{author}{\bibinfo{person}{Yu Meng}, \bibinfo{person}{Chenyan Xiong},
  \bibinfo{person}{Payal Bajaj}, \bibinfo{person}{Saurabh Tiwary},
  \bibinfo{person}{Paul Bennett}, \bibinfo{person}{Jiawei Han}, {and}
  \bibinfo{person}{Xia Song}.} \bibinfo{year}{2021}\natexlab{}.
\newblock \showarticletitle{{COCO-LM:} Correcting and Contrasting Text
  Sequences for Language Model Pretraining}. In
  \bibinfo{booktitle}{\emph{Advances in Neural Information Processing Systems
  34: Annual Conference on Neural Information Processing Systems 2021, NeurIPS
  2021, December 6-14, 2021, virtual}}. \bibinfo{pages}{23102--23114}.
\newblock


\bibitem[Mikolov et~al\mbox{.}(2013a)]%
        {DBLP:journals/corr/abs-1301-3781}
\bibfield{author}{\bibinfo{person}{Tom{\'{a}}s Mikolov}, \bibinfo{person}{Kai
  Chen}, \bibinfo{person}{Greg Corrado}, {and} \bibinfo{person}{Jeffrey Dean}.}
  \bibinfo{year}{2013}\natexlab{a}.
\newblock \showarticletitle{Efficient Estimation of Word Representations in
  Vector Space}. In \bibinfo{booktitle}{\emph{1st International Conference on
  Learning Representations, {ICLR} 2013, Scottsdale, Arizona, USA, May 2-4,
  2013, Workshop Track Proceedings}}, \bibfield{editor}{\bibinfo{person}{Yoshua
  Bengio} {and} \bibinfo{person}{Yann LeCun}} (Eds.).
\newblock


\bibitem[Mikolov et~al\mbox{.}(2013b)]%
        {mikolov2013distributed}
\bibfield{author}{\bibinfo{person}{Tomas Mikolov}, \bibinfo{person}{Ilya
  Sutskever}, \bibinfo{person}{Kai Chen}, \bibinfo{person}{Greg~S Corrado},
  {and} \bibinfo{person}{Jeff Dean}.} \bibinfo{year}{2013}\natexlab{b}.
\newblock \showarticletitle{Distributed representations of words and phrases
  and their compositionality}.
\newblock \bibinfo{journal}{\emph{Advances in neural information processing
  systems}}  \bibinfo{volume}{26} (\bibinfo{year}{2013}).
\newblock


\bibitem[Miller et~al\mbox{.}(2024)]%
        {miller2024survey}
\bibfield{author}{\bibinfo{person}{John~A Miller}, \bibinfo{person}{Mohammed
  Aldosari}, \bibinfo{person}{Farah Saeed}, \bibinfo{person}{Nasid~Habib
  Barna}, \bibinfo{person}{Subas Rana}, \bibinfo{person}{I~Budak Arpinar},
  {and} \bibinfo{person}{Ninghao Liu}.} \bibinfo{year}{2024}\natexlab{}.
\newblock \showarticletitle{A survey of deep learning and foundation models for
  time series forecasting}.
\newblock \bibinfo{journal}{\emph{arXiv preprint arXiv:2401.13912}}
  (\bibinfo{year}{2024}).
\newblock


\bibitem[Mogadala et~al\mbox{.}(2021)]%
        {DBLP:journals/jair/MogadalaKK21}
\bibfield{author}{\bibinfo{person}{Aditya Mogadala}, \bibinfo{person}{Marimuthu
  Kalimuthu}, {and} \bibinfo{person}{Dietrich Klakow}.}
  \bibinfo{year}{2021}\natexlab{}.
\newblock \showarticletitle{Trends in Integration of Vision and Language
  Research: {A} Survey of Tasks, Datasets, and Methods}.
\newblock \bibinfo{journal}{\emph{J. Artif. Intell. Res.}}
  \bibinfo{volume}{71} (\bibinfo{year}{2021}), \bibinfo{pages}{1183--1317}.
\newblock


\bibitem[Nguyen and Luu(2021)]%
        {nguyen2021contrastive}
\bibfield{author}{\bibinfo{person}{Thong Nguyen} {and}
  \bibinfo{person}{Anh~Tuan Luu}.} \bibinfo{year}{2021}\natexlab{}.
\newblock \showarticletitle{Contrastive learning for neural topic model}.
\newblock \bibinfo{journal}{\emph{Advances in neural information processing
  systems}}  \bibinfo{volume}{34} (\bibinfo{year}{2021}),
  \bibinfo{pages}{11974--11986}.
\newblock


\bibitem[Oba et~al\mbox{.}(2023)]%
        {oba2023contextual}
\bibfield{author}{\bibinfo{person}{Daisuke Oba}, \bibinfo{person}{Masahiro
  Kaneko}, {and} \bibinfo{person}{Danushka Bollegala}.}
  \bibinfo{year}{2023}\natexlab{}.
\newblock \showarticletitle{In-contextual bias suppression for large language
  models}.
\newblock \bibinfo{journal}{\emph{arXiv preprint arXiv:2309.07251}}
  (\bibinfo{year}{2023}).
\newblock


\bibitem[Oord et~al\mbox{.}(2018)]%
        {oord2018representation}
\bibfield{author}{\bibinfo{person}{Aaron van~den Oord}, \bibinfo{person}{Yazhe
  Li}, {and} \bibinfo{person}{Oriol Vinyals}.} \bibinfo{year}{2018}\natexlab{}.
\newblock \showarticletitle{Representation learning with contrastive predictive
  coding}.
\newblock \bibinfo{journal}{\emph{arXiv preprint arXiv:1807.03748}}
  (\bibinfo{year}{2018}).
\newblock


\bibitem[Ouyang et~al\mbox{.}(2023)]%
        {DBLP:conf/acl/OuyangY023}
\bibfield{author}{\bibinfo{person}{Siqi Ouyang}, \bibinfo{person}{Rong Ye},
  {and} \bibinfo{person}{Lei Li}.} \bibinfo{year}{2023}\natexlab{}.
\newblock \showarticletitle{{WACO:} Word-Aligned Contrastive Learning for
  Speech Translation}. In \bibinfo{booktitle}{\emph{Proceedings of the 61st
  Annual Meeting of the Association for Computational Linguistics (Volume 1:
  Long Papers), {ACL} 2023, Toronto, Canada, July 9-14, 2023}}.
  \bibinfo{publisher}{Association for Computational Linguistics},
  \bibinfo{pages}{3891--3907}.
\newblock


\bibitem[Pan and Kang(2021)]%
        {DBLP:conf/nips/PanK21}
\bibfield{author}{\bibinfo{person}{Erlin Pan} {and} \bibinfo{person}{Zhao
  Kang}.} \bibinfo{year}{2021}\natexlab{}.
\newblock \showarticletitle{Multi-view Contrastive Graph Clustering}. In
  \bibinfo{booktitle}{\emph{Advances in Neural Information Processing Systems
  34: Annual Conference on Neural Information Processing Systems 2021, NeurIPS
  2021, December 6-14, 2021, virtual}},
  \bibfield{editor}{\bibinfo{person}{Marc'Aurelio Ranzato},
  \bibinfo{person}{Alina Beygelzimer}, \bibinfo{person}{Yann~N. Dauphin},
  \bibinfo{person}{Percy Liang}, {and} \bibinfo{person}{Jennifer~Wortman
  Vaughan}} (Eds.). \bibinfo{pages}{2148--2159}.
\newblock


\bibitem[Pan et~al\mbox{.}(2022a)]%
        {pan2022improved}
\bibfield{author}{\bibinfo{person}{Lin Pan}, \bibinfo{person}{Chung-Wei Hang},
  \bibinfo{person}{Avirup Sil}, {and} \bibinfo{person}{Saloni Potdar}.}
  \bibinfo{year}{2022}\natexlab{a}.
\newblock \showarticletitle{Improved text classification via contrastive
  adversarial training}. In \bibinfo{booktitle}{\emph{Proceedings of the AAAI
  Conference on Artificial Intelligence}}, Vol.~\bibinfo{volume}{36}.
  \bibinfo{pages}{11130--11138}.
\newblock


\bibitem[Pan et~al\mbox{.}(2021a)]%
        {DBLP:conf/acl/PanWWL20}
\bibfield{author}{\bibinfo{person}{Xiao Pan}, \bibinfo{person}{Mingxuan Wang},
  \bibinfo{person}{Liwei Wu}, {and} \bibinfo{person}{Lei Li}.}
  \bibinfo{year}{2021}\natexlab{a}.
\newblock \showarticletitle{Contrastive Learning for Many-to-many Multilingual
  Neural Machine Translation}. In \bibinfo{booktitle}{\emph{Proceedings of the
  59th Annual Meeting of the Association for Computational Linguistics and the
  11th International Joint Conference on Natural Language Processing,
  {ACL/IJCNLP} 2021, (Volume 1: Long Papers), Virtual Event, August 1-6,
  2021}}, \bibfield{editor}{\bibinfo{person}{Chengqing Zong},
  \bibinfo{person}{Fei Xia}, \bibinfo{person}{Wenjie Li}, {and}
  \bibinfo{person}{Roberto Navigli}} (Eds.). \bibinfo{publisher}{Association
  for Computational Linguistics}, \bibinfo{pages}{244--258}.
\newblock


\bibitem[Pan et~al\mbox{.}(2021b)]%
        {pan2021contrastive}
\bibfield{author}{\bibinfo{person}{Xiao Pan}, \bibinfo{person}{Mingxuan Wang},
  \bibinfo{person}{Liwei Wu}, {and} \bibinfo{person}{Lei Li}.}
  \bibinfo{year}{2021}\natexlab{b}.
\newblock \showarticletitle{Contrastive learning for many-to-many multilingual
  neural machine translation}.
\newblock \bibinfo{journal}{\emph{arXiv preprint arXiv:2105.09501}}
  (\bibinfo{year}{2021}).
\newblock


\bibitem[Pan et~al\mbox{.}(2022b)]%
        {pan2022contrastive}
\bibfield{author}{\bibinfo{person}{Xuran Pan}, \bibinfo{person}{Tianzhu Ye},
  \bibinfo{person}{Dongchen Han}, \bibinfo{person}{Shiji Song}, {and}
  \bibinfo{person}{Gao Huang}.} \bibinfo{year}{2022}\natexlab{b}.
\newblock \showarticletitle{Contrastive language-image pre-training with
  knowledge graphs}.
\newblock \bibinfo{journal}{\emph{Advances in Neural Information Processing
  Systems}}  \bibinfo{volume}{35} (\bibinfo{year}{2022}),
  \bibinfo{pages}{22895--22910}.
\newblock


\bibitem[Paranjape et~al\mbox{.}(2021)]%
        {DBLP:conf/acl/ParanjapeMGHZ21}
\bibfield{author}{\bibinfo{person}{Bhargavi Paranjape}, \bibinfo{person}{Julian
  Michael}, \bibinfo{person}{Marjan Ghazvininejad}, \bibinfo{person}{Hannaneh
  Hajishirzi}, {and} \bibinfo{person}{Luke Zettlemoyer}.}
  \bibinfo{year}{2021}\natexlab{}.
\newblock \showarticletitle{Prompting Contrastive Explanations for Commonsense
  Reasoning Tasks}. In \bibinfo{booktitle}{\emph{Findings of the Association
  for Computational Linguistics: {ACL/IJCNLP} 2021, Online Event, August 1-6,
  2021}} \emph{(\bibinfo{series}{Findings of {ACL}},
  Vol.~\bibinfo{volume}{{ACL/IJCNLP} 2021})},
  \bibfield{editor}{\bibinfo{person}{Chengqing Zong}, \bibinfo{person}{Fei
  Xia}, \bibinfo{person}{Wenjie Li}, {and} \bibinfo{person}{Roberto Navigli}}
  (Eds.). \bibinfo{publisher}{Association for Computational Linguistics},
  \bibinfo{pages}{4179--4192}.
\newblock


\bibitem[Park et~al\mbox{.}(2022)]%
        {park2022fair}
\bibfield{author}{\bibinfo{person}{Sungho Park}, \bibinfo{person}{Jewook Lee},
  \bibinfo{person}{Pilhyeon Lee}, \bibinfo{person}{Sunhee Hwang},
  \bibinfo{person}{Dohyung Kim}, {and} \bibinfo{person}{Hyeran Byun}.}
  \bibinfo{year}{2022}\natexlab{}.
\newblock \showarticletitle{Fair contrastive learning for facial attribute
  classification}. In \bibinfo{booktitle}{\emph{Proceedings of the IEEE/CVF
  Conference on Computer Vision and Pattern Recognition}}.
  \bibinfo{pages}{10389--10398}.
\newblock


\bibitem[Park et~al\mbox{.}(2020)]%
        {park2020contrastive}
\bibfield{author}{\bibinfo{person}{Taesung Park}, \bibinfo{person}{Alexei~A
  Efros}, \bibinfo{person}{Richard Zhang}, {and} \bibinfo{person}{Jun-Yan
  Zhu}.} \bibinfo{year}{2020}\natexlab{}.
\newblock \showarticletitle{Contrastive learning for unpaired image-to-image
  translation}. In \bibinfo{booktitle}{\emph{Computer Vision--ECCV 2020: 16th
  European Conference, Glasgow, UK, August 23--28, 2020, Proceedings, Part IX
  16}}. Springer, \bibinfo{pages}{319--345}.
\newblock


\bibitem[Peng et~al\mbox{.}(2020)]%
        {peng2020graph}
\bibfield{author}{\bibinfo{person}{Zhen Peng}, \bibinfo{person}{Wenbing Huang},
  \bibinfo{person}{Minnan Luo}, \bibinfo{person}{Qinghua Zheng},
  \bibinfo{person}{Yu Rong}, \bibinfo{person}{Tingyang Xu}, {and}
  \bibinfo{person}{Junzhou Huang}.} \bibinfo{year}{2020}\natexlab{}.
\newblock \showarticletitle{Graph representation learning via graphical mutual
  information maximization}. In \bibinfo{booktitle}{\emph{Proceedings of The
  Web Conference 2020}}. \bibinfo{pages}{259--270}.
\newblock


\bibitem[Perozzi et~al\mbox{.}(2014)]%
        {perozzi2014deepwalk}
\bibfield{author}{\bibinfo{person}{Bryan Perozzi}, \bibinfo{person}{Rami
  Al-Rfou}, {and} \bibinfo{person}{Steven Skiena}.}
  \bibinfo{year}{2014}\natexlab{}.
\newblock \showarticletitle{Deepwalk: Online learning of social
  representations}. In \bibinfo{booktitle}{\emph{Proceedings of the 20th ACM
  SIGKDD international conference on Knowledge discovery and data mining}}.
  \bibinfo{pages}{701--710}.
\newblock


\bibitem[Qiu et~al\mbox{.}(2021)]%
        {qiu2021neural}
\bibfield{author}{\bibinfo{person}{Chen Qiu}, \bibinfo{person}{Timo Pfrommer},
  \bibinfo{person}{Marius Kloft}, \bibinfo{person}{Stephan Mandt}, {and}
  \bibinfo{person}{Maja Rudolph}.} \bibinfo{year}{2021}\natexlab{}.
\newblock \showarticletitle{Neural transformation learning for deep anomaly
  detection beyond images}. In \bibinfo{booktitle}{\emph{International
  conference on machine learning}}. PMLR, \bibinfo{pages}{8703--8714}.
\newblock


\bibitem[Qu et~al\mbox{.}(2021)]%
        {DBLP:conf/iclr/QuSSSC021}
\bibfield{author}{\bibinfo{person}{Yanru Qu}, \bibinfo{person}{Dinghan Shen},
  \bibinfo{person}{Yelong Shen}, \bibinfo{person}{Sandra Sajeev},
  \bibinfo{person}{Weizhu Chen}, {and} \bibinfo{person}{Jiawei Han}.}
  \bibinfo{year}{2021}\natexlab{}.
\newblock \showarticletitle{CoDA: Contrast-enhanced and Diversity-promoting
  Data Augmentation for Natural Language Understanding}. In
  \bibinfo{booktitle}{\emph{9th International Conference on Learning
  Representations, {ICLR} 2021, Virtual Event, Austria, May 3-7, 2021}}.
  \bibinfo{publisher}{OpenReview.net}.
\newblock


\bibitem[Radford et~al\mbox{.}(2021)]%
        {DBLP:conf/icml/RadfordKHRGASAM21}
\bibfield{author}{\bibinfo{person}{Alec Radford}, \bibinfo{person}{Jong~Wook
  Kim}, \bibinfo{person}{Chris Hallacy}, \bibinfo{person}{Aditya Ramesh},
  \bibinfo{person}{Gabriel Goh}, \bibinfo{person}{Sandhini Agarwal},
  \bibinfo{person}{Girish Sastry}, \bibinfo{person}{Amanda Askell},
  \bibinfo{person}{Pamela Mishkin}, \bibinfo{person}{Jack Clark},
  \bibinfo{person}{Gretchen Krueger}, {and} \bibinfo{person}{Ilya Sutskever}.}
  \bibinfo{year}{2021}\natexlab{}.
\newblock \showarticletitle{Learning Transferable Visual Models From Natural
  Language Supervision}. In \bibinfo{booktitle}{\emph{Proceedings of the 38th
  International Conference on Machine Learning, {ICML} 2021, 18-24 July 2021,
  Virtual Event}} \emph{(\bibinfo{series}{Proceedings of Machine Learning
  Research}, Vol.~\bibinfo{volume}{139})},
  \bibfield{editor}{\bibinfo{person}{Marina Meila} {and} \bibinfo{person}{Tong
  Zhang}} (Eds.). \bibinfo{publisher}{{PMLR}}, \bibinfo{pages}{8748--8763}.
\newblock


\bibitem[Rawat and Wang(2017)]%
        {rawat2017deep}
\bibfield{author}{\bibinfo{person}{Waseem Rawat} {and} \bibinfo{person}{Zenghui
  Wang}.} \bibinfo{year}{2017}\natexlab{}.
\newblock \showarticletitle{Deep convolutional neural networks for image
  classification: A comprehensive review}.
\newblock \bibinfo{journal}{\emph{Neural computation}} \bibinfo{volume}{29},
  \bibinfo{number}{9} (\bibinfo{year}{2017}), \bibinfo{pages}{2352--2449}.
\newblock


\bibitem[Reed et~al\mbox{.}(2021)]%
        {reed2021selfaugment}
\bibfield{author}{\bibinfo{person}{Colorado~J Reed}, \bibinfo{person}{Sean
  Metzger}, \bibinfo{person}{Aravind Srinivas}, \bibinfo{person}{Trevor
  Darrell}, {and} \bibinfo{person}{Kurt Keutzer}.}
  \bibinfo{year}{2021}\natexlab{}.
\newblock \showarticletitle{Selfaugment: Automatic augmentation policies for
  self-supervised learning}. In \bibinfo{booktitle}{\emph{Proceedings of the
  IEEE/CVF conference on computer vision and pattern recognition}}.
  \bibinfo{pages}{2674--2683}.
\newblock


\bibitem[Reiss and Hoshen(2023)]%
        {reiss2023mean}
\bibfield{author}{\bibinfo{person}{Tal Reiss} {and} \bibinfo{person}{Yedid
  Hoshen}.} \bibinfo{year}{2023}\natexlab{}.
\newblock \showarticletitle{Mean-shifted contrastive loss for anomaly
  detection}. In \bibinfo{booktitle}{\emph{Proceedings of the AAAI Conference
  on Artificial Intelligence}}, Vol.~\bibinfo{volume}{37}.
  \bibinfo{pages}{2155--2162}.
\newblock


\bibitem[Ren et~al\mbox{.}(2019)]%
        {ren2019time}
\bibfield{author}{\bibinfo{person}{Hansheng Ren}, \bibinfo{person}{Bixiong Xu},
  \bibinfo{person}{Yujing Wang}, \bibinfo{person}{Chao Yi},
  \bibinfo{person}{Congrui Huang}, \bibinfo{person}{Xiaoyu Kou},
  \bibinfo{person}{Tony Xing}, \bibinfo{person}{Mao Yang}, \bibinfo{person}{Jie
  Tong}, {and} \bibinfo{person}{Qi Zhang}.} \bibinfo{year}{2019}\natexlab{}.
\newblock \showarticletitle{Time-series anomaly detection service at
  microsoft}. In \bibinfo{booktitle}{\emph{Proceedings of the 25th ACM SIGKDD
  international conference on knowledge discovery \& data mining}}.
  \bibinfo{pages}{3009--3017}.
\newblock


\bibitem[Sachidananda et~al\mbox{.}(2022)]%
        {DBLP:journals/corr/abs-2202-03587}
\bibfield{author}{\bibinfo{person}{Vin Sachidananda},
  \bibinfo{person}{Shao{-}Yen Tseng}, \bibinfo{person}{Erik Marchi},
  \bibinfo{person}{Sachin Kajarekar}, {and} \bibinfo{person}{Panayiotis~G.
  Georgiou}.} \bibinfo{year}{2022}\natexlab{}.
\newblock \showarticletitle{{CALM:} Contrastive Aligned Audio-Language
  Multirate and Multimodal Representations}.
\newblock \bibinfo{journal}{\emph{CoRR}}  \bibinfo{volume}{abs/2202.03587}
  (\bibinfo{year}{2022}).
\newblock


\bibitem[Schroff et~al\mbox{.}(2015)]%
        {DBLP:conf/cvpr/SchroffKP15}
\bibfield{author}{\bibinfo{person}{Florian Schroff}, \bibinfo{person}{Dmitry
  Kalenichenko}, {and} \bibinfo{person}{James Philbin}.}
  \bibinfo{year}{2015}\natexlab{}.
\newblock \showarticletitle{FaceNet: {A} unified embedding for face recognition
  and clustering}. In \bibinfo{booktitle}{\emph{{IEEE} Conference on Computer
  Vision and Pattern Recognition, {CVPR} 2015, Boston, MA, USA, June 7-12,
  2015}}. \bibinfo{publisher}{{IEEE} Computer Society},
  \bibinfo{pages}{815--823}.
\newblock


\bibitem[Shen et~al\mbox{.}(2020)]%
        {DBLP:journals/corr/abs-2009-13818}
\bibfield{author}{\bibinfo{person}{Dinghan Shen}, \bibinfo{person}{Mingzhi
  Zheng}, \bibinfo{person}{Yelong Shen}, \bibinfo{person}{Yanru Qu}, {and}
  \bibinfo{person}{Weizhu Chen}.} \bibinfo{year}{2020}\natexlab{}.
\newblock \showarticletitle{A Simple but Tough-to-Beat Data Augmentation
  Approach for Natural Language Understanding and Generation}.
\newblock \bibinfo{journal}{\emph{CoRR}}  \bibinfo{volume}{abs/2009.13818}
  (\bibinfo{year}{2020}).
\newblock


\bibitem[Shen et~al\mbox{.}(2024)]%
        {shen2024improving}
\bibfield{author}{\bibinfo{person}{Wei Shen}, \bibinfo{person}{Xiaoying Zhang},
  \bibinfo{person}{Yuanshun Yao}, \bibinfo{person}{Rui Zheng},
  \bibinfo{person}{Hongyi Guo}, {and} \bibinfo{person}{Yang Liu}.}
  \bibinfo{year}{2024}\natexlab{}.
\newblock \showarticletitle{Improving Reinforcement Learning from Human
  Feedback Using Contrastive Rewards}.
\newblock \bibinfo{journal}{\emph{arXiv preprint arXiv:2403.07708}}
  (\bibinfo{year}{2024}).
\newblock


\bibitem[Shi et~al\mbox{.}(2021)]%
        {shi2021simple}
\bibfield{author}{\bibinfo{person}{Tian Shi}, \bibinfo{person}{Liuqing Li},
  \bibinfo{person}{Ping Wang}, {and} \bibinfo{person}{Chandan~K Reddy}.}
  \bibinfo{year}{2021}\natexlab{}.
\newblock \showarticletitle{A simple and effective self-supervised contrastive
  learning framework for aspect detection}. In
  \bibinfo{booktitle}{\emph{Proceedings of the AAAI conference on artificial
  intelligence}}, Vol.~\bibinfo{volume}{35}. \bibinfo{pages}{13815--13824}.
\newblock


\bibitem[Shiao et~al\mbox{.}(2022)]%
        {shiao2022link}
\bibfield{author}{\bibinfo{person}{William Shiao}, \bibinfo{person}{Zhichun
  Guo}, \bibinfo{person}{Tong Zhao}, \bibinfo{person}{Evangelos~E Papalexakis},
  \bibinfo{person}{Yozen Liu}, {and} \bibinfo{person}{Neil Shah}.}
  \bibinfo{year}{2022}\natexlab{}.
\newblock \showarticletitle{Link Prediction with Non-Contrastive Learning}. In
  \bibinfo{booktitle}{\emph{The Eleventh International Conference on Learning
  Representations}}.
\newblock


\bibitem[Singh et~al\mbox{.}(2022)]%
        {DBLP:conf/cvpr/SinghHGCGRK22}
\bibfield{author}{\bibinfo{person}{Amanpreet Singh}, \bibinfo{person}{Ronghang
  Hu}, \bibinfo{person}{Vedanuj Goswami}, \bibinfo{person}{Guillaume Couairon},
  \bibinfo{person}{Wojciech Galuba}, \bibinfo{person}{Marcus Rohrbach}, {and}
  \bibinfo{person}{Douwe Kiela}.} \bibinfo{year}{2022}\natexlab{}.
\newblock \showarticletitle{{FLAVA:} {A} Foundational Language And Vision
  Alignment Model}. In \bibinfo{booktitle}{\emph{{IEEE/CVF} Conference on
  Computer Vision and Pattern Recognition, {CVPR} 2022, New Orleans, LA, USA,
  June 18-24, 2022}}. \bibinfo{publisher}{{IEEE}},
  \bibinfo{pages}{15617--15629}.
\newblock


\bibitem[Su et~al\mbox{.}(2022a)]%
        {DBLP:journals/corr/abs-2209-05481}
\bibfield{author}{\bibinfo{person}{Bing Su}, \bibinfo{person}{Dazhao Du},
  \bibinfo{person}{Zhao Yang}, \bibinfo{person}{Yujie Zhou},
  \bibinfo{person}{Jiangmeng Li}, \bibinfo{person}{Anyi Rao},
  \bibinfo{person}{Hao Sun}, \bibinfo{person}{Zhiwu Lu}, {and}
  \bibinfo{person}{Ji{-}Rong Wen}.} \bibinfo{year}{2022}\natexlab{a}.
\newblock \showarticletitle{A Molecular Multimodal Foundation Model Associating
  Molecule Graphs with Natural Language}.
\newblock \bibinfo{journal}{\emph{CoRR}}  \bibinfo{volume}{abs/2209.05481}
  (\bibinfo{year}{2022}).
\newblock


\bibitem[Su et~al\mbox{.}(2022b)]%
        {su2022contrastive}
\bibfield{author}{\bibinfo{person}{Yixuan Su}, \bibinfo{person}{Tian Lan},
  \bibinfo{person}{Yan Wang}, \bibinfo{person}{Dani Yogatama},
  \bibinfo{person}{Lingpeng Kong}, {and} \bibinfo{person}{Nigel Collier}.}
  \bibinfo{year}{2022}\natexlab{b}.
\newblock \showarticletitle{A contrastive framework for neural text
  generation}.
\newblock \bibinfo{journal}{\emph{Advances in Neural Information Processing
  Systems}}  \bibinfo{volume}{35} (\bibinfo{year}{2022}),
  \bibinfo{pages}{21548--21561}.
\newblock


\bibitem[Sun et~al\mbox{.}(2021)]%
        {DBLP:conf/cvpr/SunLCYZ21}
\bibfield{author}{\bibinfo{person}{Bo Sun}, \bibinfo{person}{Banghuai Li},
  \bibinfo{person}{Shengcai Cai}, \bibinfo{person}{Ye Yuan}, {and}
  \bibinfo{person}{Chi Zhang}.} \bibinfo{year}{2021}\natexlab{}.
\newblock \showarticletitle{{FSCE:} Few-Shot Object Detection via Contrastive
  Proposal Encoding}. In \bibinfo{booktitle}{\emph{{IEEE} Conference on
  Computer Vision and Pattern Recognition, {CVPR} 2021, virtual, June 19-25,
  2021}}. \bibinfo{publisher}{Computer Vision Foundation / {IEEE}},
  \bibinfo{pages}{7352--7362}.
\newblock


\bibitem[Sun et~al\mbox{.}(2022)]%
        {DBLP:conf/bic-ta/SunCD022}
\bibfield{author}{\bibinfo{person}{Dengdi Sun}, \bibinfo{person}{Mingxin Cao},
  \bibinfo{person}{Zhuanlian Ding}, {and} \bibinfo{person}{Bin Luo}.}
  \bibinfo{year}{2022}\natexlab{}.
\newblock \showarticletitle{Graph Contrastive Learning with Intrinsic
  Augmentations}. In \bibinfo{booktitle}{\emph{Bio-Inspired Computing: Theories
  and Applications - 17th International Conference, {BIC-TA} 2022, Wuhan,
  China, December 16-18, 2022, Revised Selected Papers}}
  \emph{(\bibinfo{series}{Communications in Computer and Information Science},
  Vol.~\bibinfo{volume}{1801})}, \bibfield{editor}{\bibinfo{person}{Linqiang
  Pan}, \bibinfo{person}{Dongming Zhao}, \bibinfo{person}{Lianghao Li}, {and}
  \bibinfo{person}{Jianqing Lin}} (Eds.). \bibinfo{publisher}{Springer},
  \bibinfo{pages}{343--357}.
\newblock


\bibitem[Sun et~al\mbox{.}(2019)]%
        {sun2019infograph}
\bibfield{author}{\bibinfo{person}{Fan-Yun Sun}, \bibinfo{person}{Jordan
  Hoffman}, \bibinfo{person}{Vikas Verma}, {and} \bibinfo{person}{Jian Tang}.}
  \bibinfo{year}{2019}\natexlab{}.
\newblock \showarticletitle{InfoGraph: Unsupervised and Semi-supervised
  Graph-Level Representation Learning via Mutual Information Maximization}. In
  \bibinfo{booktitle}{\emph{International Conference on Learning
  Representations}}.
\newblock


\bibitem[Sun et~al\mbox{.}(2018)]%
        {sun2018rotate}
\bibfield{author}{\bibinfo{person}{Zhiqing Sun}, \bibinfo{person}{Zhi-Hong
  Deng}, \bibinfo{person}{Jian-Yun Nie}, {and} \bibinfo{person}{Jian Tang}.}
  \bibinfo{year}{2018}\natexlab{}.
\newblock \showarticletitle{RotatE: Knowledge Graph Embedding by Relational
  Rotation in Complex Space}. In \bibinfo{booktitle}{\emph{International
  Conference on Learning Representations}}.
\newblock


\bibitem[Suresh et~al\mbox{.}(2021)]%
        {suresh2021adversarial}
\bibfield{author}{\bibinfo{person}{Susheel Suresh}, \bibinfo{person}{Pan Li},
  \bibinfo{person}{Cong Hao}, {and} \bibinfo{person}{Jennifer Neville}.}
  \bibinfo{year}{2021}\natexlab{}.
\newblock \showarticletitle{Adversarial graph augmentation to improve graph
  contrastive learning}.
\newblock \bibinfo{journal}{\emph{Advances in Neural Information Processing
  Systems}}  \bibinfo{volume}{34} (\bibinfo{year}{2021}),
  \bibinfo{pages}{15920--15933}.
\newblock


\bibitem[Tang et~al\mbox{.}(2023)]%
        {tang2023spatio}
\bibfield{author}{\bibinfo{person}{Jiabin Tang}, \bibinfo{person}{Lianghao
  Xia}, \bibinfo{person}{Jie Hu}, {and} \bibinfo{person}{Chao Huang}.}
  \bibinfo{year}{2023}\natexlab{}.
\newblock \showarticletitle{Spatio-temporal meta contrastive learning}. In
  \bibinfo{booktitle}{\emph{Proceedings of the 32nd ACM International
  Conference on Information and Knowledge Management}}.
  \bibinfo{pages}{2412--2421}.
\newblock


\bibitem[Tashiro et~al\mbox{.}(2021)]%
        {tashiro2021csdi}
\bibfield{author}{\bibinfo{person}{Yusuke Tashiro}, \bibinfo{person}{Jiaming
  Song}, \bibinfo{person}{Yang Song}, {and} \bibinfo{person}{Stefano Ermon}.}
  \bibinfo{year}{2021}\natexlab{}.
\newblock \showarticletitle{Csdi: Conditional score-based diffusion models for
  probabilistic time series imputation}.
\newblock \bibinfo{journal}{\emph{Advances in Neural Information Processing
  Systems}}  \bibinfo{volume}{34} (\bibinfo{year}{2021}),
  \bibinfo{pages}{24804--24816}.
\newblock


\bibitem[Tian et~al\mbox{.}(2020a)]%
        {tian2020contrastive}
\bibfield{author}{\bibinfo{person}{Yonglong Tian}, \bibinfo{person}{Dilip
  Krishnan}, {and} \bibinfo{person}{Phillip Isola}.}
  \bibinfo{year}{2020}\natexlab{a}.
\newblock \showarticletitle{Contrastive multiview coding}. In
  \bibinfo{booktitle}{\emph{European conference on computer vision}}. Springer,
  \bibinfo{pages}{776--794}.
\newblock


\bibitem[Tian et~al\mbox{.}(2020b)]%
        {DBLP:conf/nips/Tian0PKSI20}
\bibfield{author}{\bibinfo{person}{Yonglong Tian}, \bibinfo{person}{Chen Sun},
  \bibinfo{person}{Ben Poole}, \bibinfo{person}{Dilip Krishnan},
  \bibinfo{person}{Cordelia Schmid}, {and} \bibinfo{person}{Phillip Isola}.}
  \bibinfo{year}{2020}\natexlab{b}.
\newblock \showarticletitle{What Makes for Good Views for Contrastive
  Learning?}. In \bibinfo{booktitle}{\emph{Advances in Neural Information
  Processing Systems 33: Annual Conference on Neural Information Processing
  Systems 2020, NeurIPS 2020, December 6-12, 2020, virtual}},
  \bibfield{editor}{\bibinfo{person}{Hugo Larochelle},
  \bibinfo{person}{Marc'Aurelio Ranzato}, \bibinfo{person}{Raia Hadsell},
  \bibinfo{person}{Maria{-}Florina Balcan}, {and} \bibinfo{person}{Hsuan{-}Tien
  Lin}} (Eds.).
\newblock


\bibitem[Tonekaboni et~al\mbox{.}(2020)]%
        {tonekaboni2020unsupervised}
\bibfield{author}{\bibinfo{person}{Sana Tonekaboni}, \bibinfo{person}{Danny
  Eytan}, {and} \bibinfo{person}{Anna Goldenberg}.}
  \bibinfo{year}{2020}\natexlab{}.
\newblock \showarticletitle{Unsupervised Representation Learning for Time
  Series with Temporal Neighborhood Coding}. In
  \bibinfo{booktitle}{\emph{International Conference on Learning
  Representations}}.
\newblock


\bibitem[Trosten et~al\mbox{.}(2021)]%
        {DBLP:conf/cvpr/TrostenLJK21}
\bibfield{author}{\bibinfo{person}{Daniel~J. Trosten}, \bibinfo{person}{Sigurd
  L{\o}kse}, \bibinfo{person}{Robert Jenssen}, {and} \bibinfo{person}{Michael
  Kampffmeyer}.} \bibinfo{year}{2021}\natexlab{}.
\newblock \showarticletitle{Reconsidering Representation Alignment for
  Multi-View Clustering}. In \bibinfo{booktitle}{\emph{{IEEE} Conference on
  Computer Vision and Pattern Recognition, {CVPR} 2021, virtual, June 19-25,
  2021}}. \bibinfo{publisher}{Computer Vision Foundation / {IEEE}},
  \bibinfo{pages}{1255--1265}.
\newblock


\bibitem[Trosten et~al\mbox{.}(2023)]%
        {DBLP:conf/cvpr/TrostenLJK23}
\bibfield{author}{\bibinfo{person}{Daniel~J. Trosten}, \bibinfo{person}{Sigurd
  L{\o}kse}, \bibinfo{person}{Robert Jenssen}, {and}
  \bibinfo{person}{Michael~C. Kampffmeyer}.} \bibinfo{year}{2023}\natexlab{}.
\newblock \showarticletitle{On the Effects of Self-supervision and Contrastive
  Alignment in Deep Multi-view Clustering}. In
  \bibinfo{booktitle}{\emph{{IEEE/CVF} Conference on Computer Vision and
  Pattern Recognition, {CVPR} 2023, Vancouver, BC, Canada, June 17-24, 2023}}.
  \bibinfo{publisher}{{IEEE}}, \bibinfo{pages}{23976--23985}.
\newblock


\bibitem[Tsai et~al\mbox{.}(2020)]%
        {tsai2020mice}
\bibfield{author}{\bibinfo{person}{Tsung~Wei Tsai}, \bibinfo{person}{Chongxuan
  Li}, {and} \bibinfo{person}{Jun Zhu}.} \bibinfo{year}{2020}\natexlab{}.
\newblock \showarticletitle{Mice: Mixture of contrastive experts for
  unsupervised image clustering}. In \bibinfo{booktitle}{\emph{International
  conference on learning representations}}.
\newblock


\bibitem[Vayansky and Kumar(2020)]%
        {vayansky2020review}
\bibfield{author}{\bibinfo{person}{Ike Vayansky} {and}
  \bibinfo{person}{Sathish~AP Kumar}.} \bibinfo{year}{2020}\natexlab{}.
\newblock \showarticletitle{A review of topic modeling methods}.
\newblock \bibinfo{journal}{\emph{Information Systems}}  \bibinfo{volume}{94}
  (\bibinfo{year}{2020}), \bibinfo{pages}{101582}.
\newblock


\bibitem[Velickovic et~al\mbox{.}(2019)]%
        {velickovic2019deep}
\bibfield{author}{\bibinfo{person}{Petar Velickovic}, \bibinfo{person}{William
  Fedus}, \bibinfo{person}{William~L Hamilton}, \bibinfo{person}{Pietro
  Li{\`o}}, \bibinfo{person}{Yoshua Bengio}, {and} \bibinfo{person}{R~Devon
  Hjelm}.} \bibinfo{year}{2019}\natexlab{}.
\newblock \showarticletitle{Deep graph infomax.}
\newblock \bibinfo{journal}{\emph{ICLR (Poster)}} \bibinfo{volume}{2},
  \bibinfo{number}{3} (\bibinfo{year}{2019}), \bibinfo{pages}{4}.
\newblock


\bibitem[Wan et~al\mbox{.}(2023)]%
        {wan2023kelly}
\bibfield{author}{\bibinfo{person}{Yixin Wan}, \bibinfo{person}{George Pu},
  \bibinfo{person}{Jiao Sun}, \bibinfo{person}{Aparna Garimella},
  \bibinfo{person}{Kai-Wei Chang}, {and} \bibinfo{person}{Nanyun Peng}.}
  \bibinfo{year}{2023}\natexlab{}.
\newblock \showarticletitle{" kelly is a warm person, joseph is a role model":
  Gender biases in llm-generated reference letters}.
\newblock \bibinfo{journal}{\emph{arXiv preprint arXiv:2310.09219}}
  (\bibinfo{year}{2023}).
\newblock


\bibitem[Wang et~al\mbox{.}(2021a)]%
        {DBLP:conf/acl/WangCZQL21}
\bibfield{author}{\bibinfo{person}{Danqing Wang}, \bibinfo{person}{Jiaze Chen},
  \bibinfo{person}{Hao Zhou}, \bibinfo{person}{Xipeng Qiu}, {and}
  \bibinfo{person}{Lei Li}.} \bibinfo{year}{2021}\natexlab{a}.
\newblock \showarticletitle{Contrastive Aligned Joint Learning for Multilingual
  Summarization}. In \bibinfo{booktitle}{\emph{Findings of the Association for
  Computational Linguistics: {ACL/IJCNLP} 2021, Online Event, August 1-6,
  2021}} \emph{(\bibinfo{series}{Findings of {ACL}},
  Vol.~\bibinfo{volume}{{ACL/IJCNLP} 2021})},
  \bibfield{editor}{\bibinfo{person}{Chengqing Zong}, \bibinfo{person}{Fei
  Xia}, \bibinfo{person}{Wenjie Li}, {and} \bibinfo{person}{Roberto Navigli}}
  (Eds.). \bibinfo{publisher}{Association for Computational Linguistics},
  \bibinfo{pages}{2739--2750}.
\newblock


\bibitem[Wang et~al\mbox{.}(2020)]%
        {wang2020coarse}
\bibfield{author}{\bibinfo{person}{Deqing Wang}, \bibinfo{person}{Baoyu Jing},
  \bibinfo{person}{Chenwei Lu}, \bibinfo{person}{Junjie Wu},
  \bibinfo{person}{Guannan Liu}, \bibinfo{person}{Chenguang Du}, {and}
  \bibinfo{person}{Fuzhen Zhuang}.} \bibinfo{year}{2020}\natexlab{}.
\newblock \showarticletitle{Coarse alignment of topic and sentiment: A unified
  model for cross-lingual sentiment classification}.
\newblock \bibinfo{journal}{\emph{IEEE Transactions on Neural Networks and
  Learning Systems}} \bibinfo{volume}{32}, \bibinfo{number}{2}
  (\bibinfo{year}{2020}), \bibinfo{pages}{736--747}.
\newblock


\bibitem[Wang et~al\mbox{.}(2021d)]%
        {wang2021cross}
\bibfield{author}{\bibinfo{person}{Deqing Wang}, \bibinfo{person}{Junjie Wu},
  \bibinfo{person}{Jingyuan Yang}, \bibinfo{person}{Baoyu Jing},
  \bibinfo{person}{Wenjie Zhang}, \bibinfo{person}{Xiaonan He}, {and}
  \bibinfo{person}{Hui Zhang}.} \bibinfo{year}{2021}\natexlab{d}.
\newblock \showarticletitle{Cross-lingual knowledge transferring by structural
  correspondence and space transfer}.
\newblock \bibinfo{journal}{\emph{IEEE Transactions on Cybernetics}}
  \bibinfo{volume}{52}, \bibinfo{number}{7} (\bibinfo{year}{2021}),
  \bibinfo{pages}{6555--6566}.
\newblock


\bibitem[Wang et~al\mbox{.}(2023d)]%
        {wang2023characterizing}
\bibfield{author}{\bibinfo{person}{Haohui Wang}, \bibinfo{person}{Baoyu Jing},
  \bibinfo{person}{Kaize Ding}, \bibinfo{person}{Yada Zhu}, {and}
  \bibinfo{person}{Dawei Zhou}.} \bibinfo{year}{2023}\natexlab{d}.
\newblock \showarticletitle{Characterizing Long-Tail Categories on Graphs}.
\newblock \bibinfo{journal}{\emph{arXiv preprint arXiv:2305.09938}}
  (\bibinfo{year}{2023}).
\newblock


\bibitem[Wang et~al\mbox{.}(2023c)]%
        {DBLP:conf/mm/WangFLG23}
\bibfield{author}{\bibinfo{person}{Jing Wang}, \bibinfo{person}{Songhe Feng},
  \bibinfo{person}{Gengyu Lyu}, {and} \bibinfo{person}{Zhibin Gu}.}
  \bibinfo{year}{2023}\natexlab{c}.
\newblock \showarticletitle{Triple-Granularity Contrastive Learning for Deep
  Multi-View Subspace Clustering}. In \bibinfo{booktitle}{\emph{Proceedings of
  the 31st {ACM} International Conference on Multimedia, {MM} 2023, Ottawa, ON,
  Canada, 29 October 2023- 3 November 2023}}. \bibinfo{publisher}{{ACM}},
  \bibinfo{pages}{2994--3002}.
\newblock


\bibitem[Wang(2017)]%
        {wang2017heterogeneous}
\bibfield{author}{\bibinfo{person}{Lidong Wang}.}
  \bibinfo{year}{2017}\natexlab{}.
\newblock \showarticletitle{Heterogeneous data and big data analytics}.
\newblock \bibinfo{journal}{\emph{Automatic Control and Information Sciences}}
  \bibinfo{volume}{3}, \bibinfo{number}{1} (\bibinfo{year}{2017}),
  \bibinfo{pages}{8--15}.
\newblock


\bibitem[Wang et~al\mbox{.}(2022e)]%
        {wang2022simkgc}
\bibfield{author}{\bibinfo{person}{Liang Wang}, \bibinfo{person}{Wei Zhao},
  \bibinfo{person}{Zhuoyu Wei}, {and} \bibinfo{person}{Jingming Liu}.}
  \bibinfo{year}{2022}\natexlab{e}.
\newblock \showarticletitle{SimKGC: Simple Contrastive Knowledge Graph
  Completion with Pre-trained Language Models}. In
  \bibinfo{booktitle}{\emph{Proceedings of the 60th Annual Meeting of the
  Association for Computational Linguistics (Volume 1: Long Papers)}}.
  \bibinfo{pages}{4281--4294}.
\newblock


\bibitem[Wang et~al\mbox{.}(2022a)]%
        {wang2022contrastive}
\bibfield{author}{\bibinfo{person}{Ran Wang}, \bibinfo{person}{Xinyu Dai},
  {et~al\mbox{.}}} \bibinfo{year}{2022}\natexlab{a}.
\newblock \showarticletitle{Contrastive learning-enhanced nearest neighbor
  mechanism for multi-label text classification}. In
  \bibinfo{booktitle}{\emph{Proceedings of the 60th Annual Meeting of the
  Association for Computational Linguistics (Volume 2: Short Papers)}}.
  \bibinfo{pages}{672--679}.
\newblock


\bibitem[Wang et~al\mbox{.}(2023e)]%
        {wang2023deep}
\bibfield{author}{\bibinfo{person}{Rui Wang}, \bibinfo{person}{Chongwei Liu},
  \bibinfo{person}{Xudong Mou}, \bibinfo{person}{Kai Gao},
  \bibinfo{person}{Xiaohui Guo}, \bibinfo{person}{Pin Liu},
  \bibinfo{person}{Tianyu Wo}, {and} \bibinfo{person}{Xudong Liu}.}
  \bibinfo{year}{2023}\natexlab{e}.
\newblock \showarticletitle{Deep contrastive one-class time series anomaly
  detection}. In \bibinfo{booktitle}{\emph{Proceedings of the 2023 SIAM
  International Conference on Data Mining (SDM)}}. SIAM,
  \bibinfo{pages}{694--702}.
\newblock


\bibitem[Wang et~al\mbox{.}(2022c)]%
        {wang2022uncovering}
\bibfield{author}{\bibinfo{person}{Ruijia Wang}, \bibinfo{person}{Xiao Wang},
  \bibinfo{person}{Chuan Shi}, {and} \bibinfo{person}{Le Song}.}
  \bibinfo{year}{2022}\natexlab{c}.
\newblock \showarticletitle{Uncovering the structural fairness in graph
  contrastive learning}.
\newblock \bibinfo{journal}{\emph{Advances in neural information processing
  systems}}  \bibinfo{volume}{35} (\bibinfo{year}{2022}),
  \bibinfo{pages}{32465--32473}.
\newblock


\bibitem[Wang and Lu(2022)]%
        {DBLP:conf/emnlp/WangL22}
\bibfield{author}{\bibinfo{person}{Tianduo Wang} {and} \bibinfo{person}{Wei
  Lu}.} \bibinfo{year}{2022}\natexlab{}.
\newblock \showarticletitle{Differentiable Data Augmentation for Contrastive
  Sentence Representation Learning}. In \bibinfo{booktitle}{\emph{Proceedings
  of the 2022 Conference on Empirical Methods in Natural Language Processing,
  {EMNLP} 2022, Abu Dhabi, United Arab Emirates, December 7-11, 2022}},
  \bibfield{editor}{\bibinfo{person}{Yoav Goldberg}, \bibinfo{person}{Zornitsa
  Kozareva}, {and} \bibinfo{person}{Yue Zhang}} (Eds.).
  \bibinfo{publisher}{Association for Computational Linguistics},
  \bibinfo{pages}{7640--7653}.
\newblock


\bibitem[Wang et~al\mbox{.}(2023a)]%
        {DBLP:journals/ijautcomp/WangCQGWWTG23}
\bibfield{author}{\bibinfo{person}{Xiao Wang}, \bibinfo{person}{Guangyao Chen},
  \bibinfo{person}{Guangwu Qian}, \bibinfo{person}{Pengcheng Gao},
  \bibinfo{person}{Xiao{-}Yong Wei}, \bibinfo{person}{Yaowei Wang},
  \bibinfo{person}{Yonghong Tian}, {and} \bibinfo{person}{Wen Gao}.}
  \bibinfo{year}{2023}\natexlab{a}.
\newblock \showarticletitle{Large-scale Multi-modal Pre-trained Models: {A}
  Comprehensive Survey}.
\newblock \bibinfo{journal}{\emph{Mach. Intell. Res.}} \bibinfo{volume}{20},
  \bibinfo{number}{4} (\bibinfo{year}{2023}), \bibinfo{pages}{447--482}.
\newblock


\bibitem[Wang et~al\mbox{.}(2021b)]%
        {wang2021self}
\bibfield{author}{\bibinfo{person}{Xiao Wang}, \bibinfo{person}{Nian Liu},
  \bibinfo{person}{Hui Han}, {and} \bibinfo{person}{Chuan Shi}.}
  \bibinfo{year}{2021}\natexlab{b}.
\newblock \showarticletitle{Self-supervised heterogeneous graph neural network
  with co-contrastive learning}. In \bibinfo{booktitle}{\emph{Proceedings of
  the 27th ACM SIGKDD conference on knowledge discovery \& data mining}}.
  \bibinfo{pages}{1726--1736}.
\newblock


\bibitem[Wang et~al\mbox{.}(2021e)]%
        {wang2021dense}
\bibfield{author}{\bibinfo{person}{Xinlong Wang}, \bibinfo{person}{Rufeng
  Zhang}, \bibinfo{person}{Chunhua Shen}, \bibinfo{person}{Tao Kong}, {and}
  \bibinfo{person}{Lei Li}.} \bibinfo{year}{2021}\natexlab{e}.
\newblock \showarticletitle{Dense contrastive learning for self-supervised
  visual pre-training}. In \bibinfo{booktitle}{\emph{Proceedings of the
  IEEE/CVF conference on computer vision and pattern recognition}}.
  \bibinfo{pages}{3024--3033}.
\newblock


\bibitem[Wang et~al\mbox{.}(2023b)]%
        {DBLP:conf/acl/WangCZY23}
\bibfield{author}{\bibinfo{person}{Yau{-}Shian Wang},
  \bibinfo{person}{Ta{-}Chung Chi}, \bibinfo{person}{Ruohong Zhang}, {and}
  \bibinfo{person}{Yiming Yang}.} \bibinfo{year}{2023}\natexlab{b}.
\newblock \showarticletitle{{PESCO:} Prompt-enhanced Self Contrastive Learning
  for Zero-shot Text Classification}. In \bibinfo{booktitle}{\emph{Proceedings
  of the 61st Annual Meeting of the Association for Computational Linguistics
  (Volume 1: Long Papers), {ACL} 2023, Toronto, Canada, July 9-14, 2023}},
  \bibfield{editor}{\bibinfo{person}{Anna Rogers}, \bibinfo{person}{Jordan~L.
  Boyd{-}Graber}, {and} \bibinfo{person}{Naoaki Okazaki}} (Eds.).
  \bibinfo{publisher}{Association for Computational Linguistics},
  \bibinfo{pages}{14897--14911}.
\newblock


\bibitem[Wang et~al\mbox{.}(2022d)]%
        {wang2022clusterscl}
\bibfield{author}{\bibinfo{person}{Yanling Wang}, \bibinfo{person}{Jing Zhang},
  \bibinfo{person}{Haoyang Li}, \bibinfo{person}{Yuxiao Dong},
  \bibinfo{person}{Hongzhi Yin}, \bibinfo{person}{Cuiping Li}, {and}
  \bibinfo{person}{Hong Chen}.} \bibinfo{year}{2022}\natexlab{d}.
\newblock \showarticletitle{Clusterscl: Cluster-aware supervised contrastive
  learning on graphs}. In \bibinfo{booktitle}{\emph{Proceedings of the ACM Web
  Conference 2022}}. \bibinfo{pages}{1611--1621}.
\newblock


\bibitem[Wang et~al\mbox{.}(2021c)]%
        {wang2021dnb}
\bibfield{author}{\bibinfo{person}{Zeya Wang}, \bibinfo{person}{Yang Ni},
  \bibinfo{person}{Baoyu Jing}, \bibinfo{person}{Deqing Wang},
  \bibinfo{person}{Hao Zhang}, {and} \bibinfo{person}{Eric Xing}.}
  \bibinfo{year}{2021}\natexlab{c}.
\newblock \showarticletitle{DNB: A joint learning framework for deep Bayesian
  nonparametric clustering}.
\newblock \bibinfo{journal}{\emph{IEEE Transactions on Neural Networks and
  Learning Systems}} \bibinfo{volume}{33}, \bibinfo{number}{12}
  (\bibinfo{year}{2021}), \bibinfo{pages}{7610--7620}.
\newblock


\bibitem[Wang et~al\mbox{.}(2023f)]%
        {DBLP:conf/nips/WangZCHLYTLWZZ23}
\bibfield{author}{\bibinfo{person}{Zehan Wang}, \bibinfo{person}{Yang Zhao},
  \bibinfo{person}{Xize Cheng}, \bibinfo{person}{Haifeng Huang},
  \bibinfo{person}{Jiageng Liu}, \bibinfo{person}{Aoxiong Yin},
  \bibinfo{person}{Li Tang}, \bibinfo{person}{Linjun Li},
  \bibinfo{person}{Yongqi Wang}, \bibinfo{person}{Ziang Zhang}, {and}
  \bibinfo{person}{Zhou Zhao}.} \bibinfo{year}{2023}\natexlab{f}.
\newblock \showarticletitle{Connecting Multi-modal Contrastive
  Representations}. In \bibinfo{booktitle}{\emph{Advances in Neural Information
  Processing Systems 36: Annual Conference on Neural Information Processing
  Systems 2023, NeurIPS 2023, New Orleans, LA, USA, December 10 - 16, 2023}},
  \bibfield{editor}{\bibinfo{person}{Alice Oh}, \bibinfo{person}{Tristan
  Naumann}, \bibinfo{person}{Amir Globerson}, \bibinfo{person}{Kate Saenko},
  \bibinfo{person}{Moritz Hardt}, {and} \bibinfo{person}{Sergey Levine}}
  (Eds.).
\newblock


\bibitem[Wang et~al\mbox{.}(2022b)]%
        {wang2022diffusiondb}
\bibfield{author}{\bibinfo{person}{Zijie~J Wang}, \bibinfo{person}{Evan
  Montoya}, \bibinfo{person}{David Munechika}, \bibinfo{person}{Haoyang Yang},
  \bibinfo{person}{Benjamin Hoover}, {and} \bibinfo{person}{Duen~Horng Chau}.}
  \bibinfo{year}{2022}\natexlab{b}.
\newblock \showarticletitle{Diffusiondb: A large-scale prompt gallery dataset
  for text-to-image generative models}.
\newblock \bibinfo{journal}{\emph{arXiv preprint arXiv:2210.14896}}
  (\bibinfo{year}{2022}).
\newblock


\bibitem[Wei et~al\mbox{.}(2021)]%
        {wei2021contrastive}
\bibfield{author}{\bibinfo{person}{Yinwei Wei}, \bibinfo{person}{Xiang Wang},
  \bibinfo{person}{Qi Li}, \bibinfo{person}{Liqiang Nie}, \bibinfo{person}{Yan
  Li}, \bibinfo{person}{Xuanping Li}, {and} \bibinfo{person}{Tat-Seng Chua}.}
  \bibinfo{year}{2021}\natexlab{}.
\newblock \showarticletitle{Contrastive learning for cold-start
  recommendation}. In \bibinfo{booktitle}{\emph{Proceedings of the 29th ACM
  International Conference on Multimedia}}. \bibinfo{pages}{5382--5390}.
\newblock


\bibitem[Wen and Fang(2023)]%
        {DBLP:conf/sigir/Wen023}
\bibfield{author}{\bibinfo{person}{Zhihao Wen} {and} \bibinfo{person}{Yuan
  Fang}.} \bibinfo{year}{2023}\natexlab{}.
\newblock \showarticletitle{Augmenting Low-Resource Text Classification with
  Graph-Grounded Pre-training and Prompting}. In
  \bibinfo{booktitle}{\emph{Proceedings of the 46th International {ACM} {SIGIR}
  Conference on Research and Development in Information Retrieval, {SIGIR}
  2023, Taipei, Taiwan, July 23-27, 2023}},
  \bibfield{editor}{\bibinfo{person}{Hsin{-}Hsi Chen},
  \bibinfo{person}{Wei{-}Jou~(Edward) Duh}, \bibinfo{person}{Hen{-}Hsen Huang},
  \bibinfo{person}{Makoto~P. Kato}, \bibinfo{person}{Josiane Mothe}, {and}
  \bibinfo{person}{Barbara Poblete}} (Eds.). \bibinfo{publisher}{{ACM}},
  \bibinfo{pages}{506--516}.
\newblock


\bibitem[Woo et~al\mbox{.}(2021)]%
        {woo2021cost}
\bibfield{author}{\bibinfo{person}{Gerald Woo}, \bibinfo{person}{Chenghao Liu},
  \bibinfo{person}{Doyen Sahoo}, \bibinfo{person}{Akshat Kumar}, {and}
  \bibinfo{person}{Steven Hoi}.} \bibinfo{year}{2021}\natexlab{}.
\newblock \showarticletitle{CoST: Contrastive Learning of Disentangled
  Seasonal-Trend Representations for Time Series Forecasting}. In
  \bibinfo{booktitle}{\emph{International Conference on Learning
  Representations}}.
\newblock


\bibitem[Wu et~al\mbox{.}(2022b)]%
        {DBLP:conf/icassp/WuSKB22}
\bibfield{author}{\bibinfo{person}{Ho{-}Hsiang Wu}, \bibinfo{person}{Prem
  Seetharaman}, \bibinfo{person}{Kundan Kumar}, {and}
  \bibinfo{person}{Juan~Pablo Bello}.} \bibinfo{year}{2022}\natexlab{b}.
\newblock \showarticletitle{Wav2CLIP: Learning Robust Audio Representations
  from Clip}. In \bibinfo{booktitle}{\emph{{IEEE} International Conference on
  Acoustics, Speech and Signal Processing, {ICASSP} 2022, Virtual and
  Singapore, 23-27 May 2022}}. \bibinfo{publisher}{{IEEE}},
  \bibinfo{pages}{4563--4567}.
\newblock


\bibitem[Wu et~al\mbox{.}(2022a)]%
        {DBLP:conf/cvpr/0019CZLCZ22}
\bibfield{author}{\bibinfo{person}{Wei Wu}, \bibinfo{person}{Hao Chang},
  \bibinfo{person}{Yonghua Zheng}, \bibinfo{person}{Zhu Li},
  \bibinfo{person}{Zhiwen Chen}, {and} \bibinfo{person}{Ziheng Zhang}.}
  \bibinfo{year}{2022}\natexlab{a}.
\newblock \showarticletitle{Contrastive Learning-based Robust Object Detection
  under Smoky Conditions}. In \bibinfo{booktitle}{\emph{{IEEE/CVF} Conference
  on Computer Vision and Pattern Recognition Workshops, {CVPR} Workshops 2022,
  New Orleans, LA, USA, June 19-20, 2022}}. \bibinfo{publisher}{{IEEE}},
  \bibinfo{pages}{4294--4301}.
\newblock


\bibitem[Wu et~al\mbox{.}(2023)]%
        {DBLP:conf/icassp/WuCZHBD23}
\bibfield{author}{\bibinfo{person}{Yusong Wu}, \bibinfo{person}{Ke Chen},
  \bibinfo{person}{Tianyu Zhang}, \bibinfo{person}{Yuchen Hui},
  \bibinfo{person}{Taylor Berg{-}Kirkpatrick}, {and} \bibinfo{person}{Shlomo
  Dubnov}.} \bibinfo{year}{2023}\natexlab{}.
\newblock \showarticletitle{Large-Scale Contrastive Language-Audio Pretraining
  with Feature Fusion and Keyword-to-Caption Augmentation}. In
  \bibinfo{booktitle}{\emph{{IEEE} International Conference on Acoustics,
  Speech and Signal Processing {ICASSP} 2023, Rhodes Island, Greece, June 4-10,
  2023}}. \bibinfo{publisher}{{IEEE}}, \bibinfo{pages}{1--5}.
\newblock


\bibitem[Wu et~al\mbox{.}(2020)]%
        {DBLP:journals/corr/abs-2012-15466}
\bibfield{author}{\bibinfo{person}{Zhuofeng Wu}, \bibinfo{person}{Sinong Wang},
  \bibinfo{person}{Jiatao Gu}, \bibinfo{person}{Madian Khabsa},
  \bibinfo{person}{Fei Sun}, {and} \bibinfo{person}{Hao Ma}.}
  \bibinfo{year}{2020}\natexlab{}.
\newblock \showarticletitle{{CLEAR:} Contrastive Learning for Sentence
  Representation}.
\newblock \bibinfo{journal}{\emph{CoRR}}  \bibinfo{volume}{abs/2012.15466}
  (\bibinfo{year}{2020}).
\newblock


\bibitem[Xia et~al\mbox{.}(2022)]%
        {DBLP:conf/www/XiaWCHL22}
\bibfield{author}{\bibinfo{person}{Jun Xia}, \bibinfo{person}{Lirong Wu},
  \bibinfo{person}{Jintao Chen}, \bibinfo{person}{Bozhen Hu}, {and}
  \bibinfo{person}{Stan~Z. Li}.} \bibinfo{year}{2022}\natexlab{}.
\newblock \showarticletitle{SimGRACE: {A} Simple Framework for Graph
  Contrastive Learning without Data Augmentation}. In
  \bibinfo{booktitle}{\emph{{WWW} '22: The {ACM} Web Conference 2022, Virtual
  Event, Lyon, France, April 25 - 29, 2022}},
  \bibfield{editor}{\bibinfo{person}{Fr{\'{e}}d{\'{e}}rique Laforest},
  \bibinfo{person}{Rapha{\"{e}}l Troncy}, \bibinfo{person}{Elena Simperl},
  \bibinfo{person}{Deepak Agarwal}, \bibinfo{person}{Aristides Gionis},
  \bibinfo{person}{Ivan Herman}, {and} \bibinfo{person}{Lionel M{\'{e}}dini}}
  (Eds.). \bibinfo{publisher}{{ACM}}, \bibinfo{pages}{1070--1079}.
\newblock


\bibitem[Xiao et~al\mbox{.}(2021)]%
        {DBLP:conf/iclr/Xiao0ED21}
\bibfield{author}{\bibinfo{person}{Tete Xiao}, \bibinfo{person}{Xiaolong Wang},
  \bibinfo{person}{Alexei~A. Efros}, {and} \bibinfo{person}{Trevor Darrell}.}
  \bibinfo{year}{2021}\natexlab{}.
\newblock \showarticletitle{What Should Not Be Contrastive in Contrastive
  Learning}. In \bibinfo{booktitle}{\emph{9th International Conference on
  Learning Representations, {ICLR} 2021, Virtual Event, Austria, May 3-7,
  2021}}. \bibinfo{publisher}{OpenReview.net}.
\newblock


\bibitem[Xie et~al\mbox{.}(2023)]%
        {DBLP:conf/cvpr/XieSXZZZ23}
\bibfield{author}{\bibinfo{person}{Chen{-}Wei Xie}, \bibinfo{person}{Siyang
  Sun}, \bibinfo{person}{Xiong Xiong}, \bibinfo{person}{Yun Zheng},
  \bibinfo{person}{Deli Zhao}, {and} \bibinfo{person}{Jingren Zhou}.}
  \bibinfo{year}{2023}\natexlab{}.
\newblock \showarticletitle{{RA-CLIP:} Retrieval Augmented Contrastive
  Language-Image Pre-Training}. In \bibinfo{booktitle}{\emph{{IEEE/CVF}
  Conference on Computer Vision and Pattern Recognition, {CVPR} 2023,
  Vancouver, BC, Canada, June 17-24, 2023}}. \bibinfo{publisher}{{IEEE}},
  \bibinfo{pages}{19265--19274}.
\newblock


\bibitem[Xie et~al\mbox{.}(2021)]%
        {xie2021detco}
\bibfield{author}{\bibinfo{person}{Enze Xie}, \bibinfo{person}{Jian Ding},
  \bibinfo{person}{Wenhai Wang}, \bibinfo{person}{Xiaohang Zhan},
  \bibinfo{person}{Hang Xu}, \bibinfo{person}{Peize Sun},
  \bibinfo{person}{Zhenguo Li}, {and} \bibinfo{person}{Ping Luo}.}
  \bibinfo{year}{2021}\natexlab{}.
\newblock \showarticletitle{Detco: Unsupervised contrastive learning for object
  detection}. In \bibinfo{booktitle}{\emph{Proceedings of the IEEE/CVF
  international conference on computer vision}}. \bibinfo{pages}{8392--8401}.
\newblock


\bibitem[Xie et~al\mbox{.}(2022)]%
        {xie2022contrastive}
\bibfield{author}{\bibinfo{person}{Xu Xie}, \bibinfo{person}{Fei Sun},
  \bibinfo{person}{Zhaoyang Liu}, \bibinfo{person}{Shiwen Wu},
  \bibinfo{person}{Jinyang Gao}, \bibinfo{person}{Jiandong Zhang},
  \bibinfo{person}{Bolin Ding}, {and} \bibinfo{person}{Bin Cui}.}
  \bibinfo{year}{2022}\natexlab{}.
\newblock \showarticletitle{Contrastive learning for sequential
  recommendation}. In \bibinfo{booktitle}{\emph{2022 IEEE 38th international
  conference on data engineering (ICDE)}}. IEEE, \bibinfo{pages}{1259--1273}.
\newblock


\bibitem[Xu et~al\mbox{.}(2023b)]%
        {DBLP:journals/corr/abs-2310-02263}
\bibfield{author}{\bibinfo{person}{Canwen Xu}, \bibinfo{person}{Corby Rosset},
  \bibinfo{person}{Luciano~Del Corro}, \bibinfo{person}{Shweti Mahajan},
  \bibinfo{person}{Julian~J. McAuley}, \bibinfo{person}{Jennifer Neville},
  \bibinfo{person}{Ahmed~Hassan Awadallah}, {and} \bibinfo{person}{Nikhil
  Rao}.} \bibinfo{year}{2023}\natexlab{b}.
\newblock \showarticletitle{Contrastive Post-training Large Language Models on
  Data Curriculum}.
\newblock \bibinfo{journal}{\emph{CoRR}}  \bibinfo{volume}{abs/2310.02263}
  (\bibinfo{year}{2023}).
\newblock


\bibitem[Xu et~al\mbox{.}(2021)]%
        {xu2021infogcl}
\bibfield{author}{\bibinfo{person}{Dongkuan Xu}, \bibinfo{person}{Wei Cheng},
  \bibinfo{person}{Dongsheng Luo}, \bibinfo{person}{Haifeng Chen}, {and}
  \bibinfo{person}{Xiang Zhang}.} \bibinfo{year}{2021}\natexlab{}.
\newblock \showarticletitle{Infogcl: Information-aware graph contrastive
  learning}.
\newblock \bibinfo{journal}{\emph{Advances in Neural Information Processing
  Systems}}  \bibinfo{volume}{34} (\bibinfo{year}{2021}),
  \bibinfo{pages}{30414--30425}.
\newblock


\bibitem[Xu et~al\mbox{.}(2022a)]%
        {DBLP:conf/cvpr/XuT0P0022}
\bibfield{author}{\bibinfo{person}{Jie Xu}, \bibinfo{person}{Huayi Tang},
  \bibinfo{person}{Yazhou Ren}, \bibinfo{person}{Liang Peng},
  \bibinfo{person}{Xiaofeng Zhu}, {and} \bibinfo{person}{Lifang He}.}
  \bibinfo{year}{2022}\natexlab{a}.
\newblock \showarticletitle{Multi-level Feature Learning for Contrastive
  Multi-view Clustering}. In \bibinfo{booktitle}{\emph{{IEEE/CVF} Conference on
  Computer Vision and Pattern Recognition, {CVPR} 2022, New Orleans, LA, USA,
  June 18-24, 2022}}. \bibinfo{publisher}{{IEEE}},
  \bibinfo{pages}{16030--16039}.
\newblock


\bibitem[Xu et~al\mbox{.}(2024)]%
        {xu2024survey}
\bibfield{author}{\bibinfo{person}{Mengwei Xu}, \bibinfo{person}{Wangsong Yin},
  \bibinfo{person}{Dongqi Cai}, \bibinfo{person}{Rongjie Yi},
  \bibinfo{person}{Daliang Xu}, \bibinfo{person}{Qipeng Wang},
  \bibinfo{person}{Bingyang Wu}, \bibinfo{person}{Yihao Zhao},
  \bibinfo{person}{Chen Yang}, \bibinfo{person}{Shihe Wang}, {et~al\mbox{.}}}
  \bibinfo{year}{2024}\natexlab{}.
\newblock \showarticletitle{A survey of resource-efficient llm and multimodal
  foundation models}.
\newblock \bibinfo{journal}{\emph{arXiv preprint arXiv:2401.08092}}
  (\bibinfo{year}{2024}).
\newblock


\bibitem[Xu et~al\mbox{.}(2022b)]%
        {xu2022sequence}
\bibfield{author}{\bibinfo{person}{Shusheng Xu}, \bibinfo{person}{Xingxing
  Zhang}, \bibinfo{person}{Yi Wu}, {and} \bibinfo{person}{Furu Wei}.}
  \bibinfo{year}{2022}\natexlab{b}.
\newblock \showarticletitle{Sequence level contrastive learning for text
  summarization}. In \bibinfo{booktitle}{\emph{Proceedings of the AAAI
  conference on artificial intelligence}}, Vol.~\bibinfo{volume}{36}.
  \bibinfo{pages}{11556--11565}.
\newblock


\bibitem[Xu et~al\mbox{.}(2023a)]%
        {xu2023temporal}
\bibfield{author}{\bibinfo{person}{Yi Xu}, \bibinfo{person}{Junjie Ou},
  \bibinfo{person}{Hui Xu}, {and} \bibinfo{person}{Luoyi Fu}.}
  \bibinfo{year}{2023}\natexlab{a}.
\newblock \showarticletitle{Temporal knowledge graph reasoning with historical
  contrastive learning}. In \bibinfo{booktitle}{\emph{Proceedings of the AAAI
  Conference on Artificial Intelligence}}, Vol.~\bibinfo{volume}{37}.
  \bibinfo{pages}{4765--4773}.
\newblock


\bibitem[Xu et~al\mbox{.}(2023c)]%
        {DBLP:conf/wsdm/Xu0QLXHH23}
\bibfield{author}{\bibinfo{person}{Ziyun Xu}, \bibinfo{person}{Chengyu Wang},
  \bibinfo{person}{Minghui Qiu}, \bibinfo{person}{Fuli Luo},
  \bibinfo{person}{Runxin Xu}, \bibinfo{person}{Songfang Huang}, {and}
  \bibinfo{person}{Jun Huang}.} \bibinfo{year}{2023}\natexlab{c}.
\newblock \showarticletitle{Making Pre-trained Language Models End-to-end
  Few-shot Learners with Contrastive Prompt Tuning}. In
  \bibinfo{booktitle}{\emph{Proceedings of the Sixteenth {ACM} International
  Conference on Web Search and Data Mining, {WSDM} 2023, Singapore, 27 February
  2023 - 3 March 2023}}, \bibfield{editor}{\bibinfo{person}{Tat{-}Seng Chua},
  \bibinfo{person}{Hady~W. Lauw}, \bibinfo{person}{Luo Si},
  \bibinfo{person}{Evimaria Terzi}, {and} \bibinfo{person}{Panayiotis
  Tsaparas}} (Eds.). \bibinfo{publisher}{{ACM}}, \bibinfo{pages}{438--446}.
\newblock


\bibitem[Yan et~al\mbox{.}(2021a)]%
        {yan2021deep}
\bibfield{author}{\bibinfo{person}{Xiaoqiang Yan}, \bibinfo{person}{Shizhe Hu},
  \bibinfo{person}{Yiqiao Mao}, \bibinfo{person}{Yangdong Ye}, {and}
  \bibinfo{person}{Hui Yu}.} \bibinfo{year}{2021}\natexlab{a}.
\newblock \showarticletitle{Deep multi-view learning methods: A review}.
\newblock \bibinfo{journal}{\emph{Neurocomputing}}  \bibinfo{volume}{448}
  (\bibinfo{year}{2021}), \bibinfo{pages}{106--129}.
\newblock


\bibitem[Yan et~al\mbox{.}(2024)]%
        {yan2024reconciling}
\bibfield{author}{\bibinfo{person}{Yuchen Yan}, \bibinfo{person}{Baoyu Jing},
  \bibinfo{person}{Lihui Liu}, \bibinfo{person}{Ruijie Wang},
  \bibinfo{person}{Jinning Li}, \bibinfo{person}{Tarek Abdelzaher}, {and}
  \bibinfo{person}{Hanghang Tong}.} \bibinfo{year}{2024}\natexlab{}.
\newblock \showarticletitle{Reconciling Competing Sampling Strategies of
  Network Embedding}.
\newblock \bibinfo{journal}{\emph{Advances in Neural Information Processing
  Systems}}  \bibinfo{volume}{36} (\bibinfo{year}{2024}).
\newblock


\bibitem[Yan et~al\mbox{.}(2021b)]%
        {yan2021dynamic}
\bibfield{author}{\bibinfo{person}{Yuchen Yan}, \bibinfo{person}{Lihui Liu},
  \bibinfo{person}{Yikun Ban}, \bibinfo{person}{Baoyu Jing}, {and}
  \bibinfo{person}{Hanghang Tong}.} \bibinfo{year}{2021}\natexlab{b}.
\newblock \showarticletitle{Dynamic knowledge graph alignment}. In
  \bibinfo{booktitle}{\emph{Proceedings of the AAAI conference on artificial
  intelligence}}, Vol.~\bibinfo{volume}{35}. \bibinfo{pages}{4564--4572}.
\newblock


\bibitem[Yang et~al\mbox{.}(2022c)]%
        {DBLP:journals/corr/abs-2211-01335}
\bibfield{author}{\bibinfo{person}{An Yang}, \bibinfo{person}{Junshu Pan},
  \bibinfo{person}{Junyang Lin}, \bibinfo{person}{Rui Men},
  \bibinfo{person}{Yichang Zhang}, \bibinfo{person}{Jingren Zhou}, {and}
  \bibinfo{person}{Chang Zhou}.} \bibinfo{year}{2022}\natexlab{c}.
\newblock \showarticletitle{Chinese {CLIP:} Contrastive Vision-Language
  Pretraining in Chinese}.
\newblock \bibinfo{journal}{\emph{CoRR}}  \bibinfo{volume}{abs/2211.01335}
  (\bibinfo{year}{2022}).
\newblock


\bibitem[Yang et~al\mbox{.}(2022b)]%
        {yang2022unified}
\bibfield{author}{\bibinfo{person}{Jianwei Yang}, \bibinfo{person}{Chunyuan
  Li}, \bibinfo{person}{Pengchuan Zhang}, \bibinfo{person}{Bin Xiao},
  \bibinfo{person}{Ce Liu}, \bibinfo{person}{Lu Yuan}, {and}
  \bibinfo{person}{Jianfeng Gao}.} \bibinfo{year}{2022}\natexlab{b}.
\newblock \showarticletitle{Unified contrastive learning in image-text-label
  space}. In \bibinfo{booktitle}{\emph{Proceedings of the IEEE/CVF Conference
  on Computer Vision and Pattern Recognition}}. \bibinfo{pages}{19163--19173}.
\newblock


\bibitem[Yang et~al\mbox{.}(2023c)]%
        {DBLP:conf/acl/YangYNGX23}
\bibfield{author}{\bibinfo{person}{Jiuding Yang}, \bibinfo{person}{Yakun Yu},
  \bibinfo{person}{Di Niu}, \bibinfo{person}{Weidong Guo}, {and}
  \bibinfo{person}{Yu Xu}.} \bibinfo{year}{2023}\natexlab{c}.
\newblock \showarticletitle{ConFEDE: Contrastive Feature Decomposition for
  Multimodal Sentiment Analysis}. In \bibinfo{booktitle}{\emph{Proceedings of
  the 61st Annual Meeting of the Association for Computational Linguistics
  (Volume 1: Long Papers), {ACL} 2023, Toronto, Canada, July 9-14, 2023}},
  \bibfield{editor}{\bibinfo{person}{Anna Rogers}, \bibinfo{person}{Jordan~L.
  Boyd{-}Graber}, {and} \bibinfo{person}{Naoaki Okazaki}} (Eds.).
  \bibinfo{publisher}{Association for Computational Linguistics},
  \bibinfo{pages}{7617--7630}.
\newblock


\bibitem[Yang and Hong(2022)]%
        {yang2022unsupervised}
\bibfield{author}{\bibinfo{person}{Ling Yang} {and} \bibinfo{person}{Shenda
  Hong}.} \bibinfo{year}{2022}\natexlab{}.
\newblock \showarticletitle{Unsupervised time-series representation learning
  with iterative bilinear temporal-spectral fusion}. In
  \bibinfo{booktitle}{\emph{International conference on machine learning}}.
  PMLR, \bibinfo{pages}{25038--25054}.
\newblock


\bibitem[Yang et~al\mbox{.}(2023a)]%
        {DBLP:conf/iccv/YangHY23}
\bibfield{author}{\bibinfo{person}{Serin Yang}, \bibinfo{person}{Hyunmin
  Hwang}, {and} \bibinfo{person}{Jong~Chul Ye}.}
  \bibinfo{year}{2023}\natexlab{a}.
\newblock \showarticletitle{Zero-Shot Contrastive Loss for Text-Guided
  Diffusion Image Style Transfer}. In \bibinfo{booktitle}{\emph{{IEEE/CVF}
  International Conference on Computer Vision, {ICCV} 2023, Paris, France,
  October 1-6, 2023}}. \bibinfo{publisher}{{IEEE}},
  \bibinfo{pages}{22816--22825}.
\newblock


\bibitem[Yang et~al\mbox{.}(2022a)]%
        {yang2022knowledge}
\bibfield{author}{\bibinfo{person}{Yuhao Yang}, \bibinfo{person}{Chao Huang},
  \bibinfo{person}{Lianghao Xia}, {and} \bibinfo{person}{Chenliang Li}.}
  \bibinfo{year}{2022}\natexlab{a}.
\newblock \showarticletitle{Knowledge graph contrastive learning for
  recommendation}. In \bibinfo{booktitle}{\emph{Proceedings of the 45th
  International ACM SIGIR Conference on Research and Development in Information
  Retrieval}}. \bibinfo{pages}{1434--1443}.
\newblock


\bibitem[Yang et~al\mbox{.}(2023b)]%
        {yang2023generative}
\bibfield{author}{\bibinfo{person}{Yonghui Yang}, \bibinfo{person}{Zhengwei
  Wu}, \bibinfo{person}{Le Wu}, \bibinfo{person}{Kun Zhang},
  \bibinfo{person}{Richang Hong}, \bibinfo{person}{Zhiqiang Zhang},
  \bibinfo{person}{Jun Zhou}, {and} \bibinfo{person}{Meng Wang}.}
  \bibinfo{year}{2023}\natexlab{b}.
\newblock \showarticletitle{Generative-contrastive graph learning for
  recommendation}. In \bibinfo{booktitle}{\emph{Proceedings of the 46th
  International ACM SIGIR Conference on Research and Development in Information
  Retrieval}}. \bibinfo{pages}{1117--1126}.
\newblock


\bibitem[Yang et~al\mbox{.}(2023d)]%
        {yang2023dcdetector}
\bibfield{author}{\bibinfo{person}{Yiyuan Yang}, \bibinfo{person}{Chaoli
  Zhang}, \bibinfo{person}{Tian Zhou}, \bibinfo{person}{Qingsong Wen}, {and}
  \bibinfo{person}{Liang Sun}.} \bibinfo{year}{2023}\natexlab{d}.
\newblock \showarticletitle{Dcdetector: Dual attention contrastive
  representation learning for time series anomaly detection}. In
  \bibinfo{booktitle}{\emph{Proceedings of the 29th ACM SIGKDD Conference on
  Knowledge Discovery and Data Mining}}. \bibinfo{pages}{3033--3045}.
\newblock


\bibitem[Ye et~al\mbox{.}(2021b)]%
        {ye2021contrastive}
\bibfield{author}{\bibinfo{person}{Hongbin Ye}, \bibinfo{person}{Ningyu Zhang},
  \bibinfo{person}{Shumin Deng}, \bibinfo{person}{Mosha Chen},
  \bibinfo{person}{Chuanqi Tan}, \bibinfo{person}{Fei Huang}, {and}
  \bibinfo{person}{Huajun Chen}.} \bibinfo{year}{2021}\natexlab{b}.
\newblock \showarticletitle{Contrastive triple extraction with generative
  transformer}. In \bibinfo{booktitle}{\emph{Proceedings of the AAAI conference
  on artificial intelligence}}, Vol.~\bibinfo{volume}{35}.
  \bibinfo{pages}{14257--14265}.
\newblock


\bibitem[Ye et~al\mbox{.}(2021a)]%
        {DBLP:conf/emnlp/YeKO21}
\bibfield{author}{\bibinfo{person}{Seonghyeon Ye}, \bibinfo{person}{Jiseon
  Kim}, {and} \bibinfo{person}{Alice Oh}.} \bibinfo{year}{2021}\natexlab{a}.
\newblock \showarticletitle{Efficient Contrastive Learning via Novel Data
  Augmentation and Curriculum Learning}. In
  \bibinfo{booktitle}{\emph{Proceedings of the 2021 Conference on Empirical
  Methods in Natural Language Processing, {EMNLP} 2021, Virtual Event / Punta
  Cana, Dominican Republic, 7-11 November, 2021}},
  \bibfield{editor}{\bibinfo{person}{Marie{-}Francine Moens},
  \bibinfo{person}{Xuanjing Huang}, \bibinfo{person}{Lucia Specia}, {and}
  \bibinfo{person}{Scott~Wen{-}tau Yih}} (Eds.).
  \bibinfo{publisher}{Association for Computational Linguistics},
  \bibinfo{pages}{1832--1838}.
\newblock


\bibitem[Ye et~al\mbox{.}(2023)]%
        {DBLP:conf/acl/YeHRJLHYZ23}
\bibfield{author}{\bibinfo{person}{Zhenhui Ye}, \bibinfo{person}{Rongjie
  Huang}, \bibinfo{person}{Yi Ren}, \bibinfo{person}{Ziyue Jiang},
  \bibinfo{person}{Jinglin Liu}, \bibinfo{person}{Jinzheng He},
  \bibinfo{person}{Xiang Yin}, {and} \bibinfo{person}{Zhou Zhao}.}
  \bibinfo{year}{2023}\natexlab{}.
\newblock \showarticletitle{CLAPSpeech: Learning Prosody from Text Context with
  Contrastive Language-Audio Pre-Training}. In
  \bibinfo{booktitle}{\emph{Proceedings of the 61st Annual Meeting of the
  Association for Computational Linguistics (Volume 1: Long Papers), {ACL}
  2023, Toronto, Canada, July 9-14, 2023}},
  \bibfield{editor}{\bibinfo{person}{Anna Rogers}, \bibinfo{person}{Jordan~L.
  Boyd{-}Graber}, {and} \bibinfo{person}{Naoaki Okazaki}} (Eds.).
  \bibinfo{publisher}{Association for Computational Linguistics},
  \bibinfo{pages}{9317--9331}.
\newblock


\bibitem[Yeh et~al\mbox{.}(2023)]%
        {yeh2023toward}
\bibfield{author}{\bibinfo{person}{Chin-Chia~Michael Yeh}, \bibinfo{person}{Xin
  Dai}, \bibinfo{person}{Huiyuan Chen}, \bibinfo{person}{Yan Zheng},
  \bibinfo{person}{Yujie Fan}, \bibinfo{person}{Audrey Der},
  \bibinfo{person}{Vivian Lai}, \bibinfo{person}{Zhongfang Zhuang},
  \bibinfo{person}{Junpeng Wang}, \bibinfo{person}{Liang Wang},
  {et~al\mbox{.}}} \bibinfo{year}{2023}\natexlab{}.
\newblock \showarticletitle{Toward a foundation model for time series data}. In
  \bibinfo{booktitle}{\emph{Proceedings of the 32nd ACM International
  Conference on Information and Knowledge Management}}.
  \bibinfo{pages}{4400--4404}.
\newblock


\bibitem[Yin et~al\mbox{.}(2023)]%
        {yin2023coco}
\bibfield{author}{\bibinfo{person}{Nan Yin}, \bibinfo{person}{Li Shen},
  \bibinfo{person}{Mengzhu Wang}, \bibinfo{person}{Long Lan},
  \bibinfo{person}{Zeyu Ma}, \bibinfo{person}{Chong Chen},
  \bibinfo{person}{Xian-Sheng Hua}, {and} \bibinfo{person}{Xiao Luo}.}
  \bibinfo{year}{2023}\natexlab{}.
\newblock \showarticletitle{Coco: A coupled contrastive framework for
  unsupervised domain adaptive graph classification}. In
  \bibinfo{booktitle}{\emph{International Conference on Machine Learning}}.
  PMLR, \bibinfo{pages}{40040--40053}.
\newblock


\bibitem[Yin et~al\mbox{.}(2022)]%
        {yin2022autogcl}
\bibfield{author}{\bibinfo{person}{Yihang Yin}, \bibinfo{person}{Qingzhong
  Wang}, \bibinfo{person}{Siyu Huang}, \bibinfo{person}{Haoyi Xiong}, {and}
  \bibinfo{person}{Xiang Zhang}.} \bibinfo{year}{2022}\natexlab{}.
\newblock \showarticletitle{Autogcl: Automated graph contrastive learning via
  learnable view generators}. In \bibinfo{booktitle}{\emph{Proceedings of the
  AAAI conference on artificial intelligence}}, Vol.~\bibinfo{volume}{36}.
  \bibinfo{pages}{8892--8900}.
\newblock


\bibitem[You et~al\mbox{.}(2021)]%
        {DBLP:conf/icml/YouCSW21}
\bibfield{author}{\bibinfo{person}{Yuning You}, \bibinfo{person}{Tianlong
  Chen}, \bibinfo{person}{Yang Shen}, {and} \bibinfo{person}{Zhangyang Wang}.}
  \bibinfo{year}{2021}\natexlab{}.
\newblock \showarticletitle{Graph Contrastive Learning Automated}. In
  \bibinfo{booktitle}{\emph{Proceedings of the 38th International Conference on
  Machine Learning, {ICML} 2021, 18-24 July 2021, Virtual Event}}
  \emph{(\bibinfo{series}{Proceedings of Machine Learning Research},
  Vol.~\bibinfo{volume}{139})}, \bibfield{editor}{\bibinfo{person}{Marina
  Meila} {and} \bibinfo{person}{Tong Zhang}} (Eds.).
  \bibinfo{publisher}{{PMLR}}, \bibinfo{pages}{12121--12132}.
\newblock


\bibitem[You et~al\mbox{.}(2020)]%
        {DBLP:conf/nips/YouCSCWS20}
\bibfield{author}{\bibinfo{person}{Yuning You}, \bibinfo{person}{Tianlong
  Chen}, \bibinfo{person}{Yongduo Sui}, \bibinfo{person}{Ting Chen},
  \bibinfo{person}{Zhangyang Wang}, {and} \bibinfo{person}{Yang Shen}.}
  \bibinfo{year}{2020}\natexlab{}.
\newblock \showarticletitle{Graph Contrastive Learning with Augmentations}. In
  \bibinfo{booktitle}{\emph{Advances in Neural Information Processing Systems
  33: Annual Conference on Neural Information Processing Systems 2020, NeurIPS
  2020, December 6-12, 2020, virtual}}, \bibfield{editor}{\bibinfo{person}{Hugo
  Larochelle}, \bibinfo{person}{Marc'Aurelio Ranzato}, \bibinfo{person}{Raia
  Hadsell}, \bibinfo{person}{Maria{-}Florina Balcan}, {and}
  \bibinfo{person}{Hsuan{-}Tien Lin}} (Eds.).
\newblock


\bibitem[You et~al\mbox{.}(2022)]%
        {DBLP:conf/wsdm/YouCWS22}
\bibfield{author}{\bibinfo{person}{Yuning You}, \bibinfo{person}{Tianlong
  Chen}, \bibinfo{person}{Zhangyang Wang}, {and} \bibinfo{person}{Yang Shen}.}
  \bibinfo{year}{2022}\natexlab{}.
\newblock \showarticletitle{Bringing Your Own View: Graph Contrastive Learning
  without Prefabricated Data Augmentations}. In
  \bibinfo{booktitle}{\emph{{WSDM} '22: The Fifteenth {ACM} International
  Conference on Web Search and Data Mining, Virtual Event / Tempe, AZ, USA,
  February 21 - 25, 2022}}, \bibfield{editor}{\bibinfo{person}{K.~Selcuk
  Candan}, \bibinfo{person}{Huan Liu}, \bibinfo{person}{Leman Akoglu},
  \bibinfo{person}{Xin~Luna Dong}, {and} \bibinfo{person}{Jiliang Tang}}
  (Eds.). \bibinfo{publisher}{{ACM}}, \bibinfo{pages}{1300--1309}.
\newblock


\bibitem[Yu et~al\mbox{.}(2022a)]%
        {yu2022coca}
\bibfield{author}{\bibinfo{person}{Jiahui Yu}, \bibinfo{person}{Zirui Wang},
  \bibinfo{person}{Vijay Vasudevan}, \bibinfo{person}{Legg Yeung},
  \bibinfo{person}{Mojtaba Seyedhosseini}, {and} \bibinfo{person}{Yonghui Wu}.}
  \bibinfo{year}{2022}\natexlab{a}.
\newblock \showarticletitle{Coca: Contrastive captioners are image-text
  foundation models}.
\newblock \bibinfo{journal}{\emph{arXiv preprint arXiv:2205.01917}}
  (\bibinfo{year}{2022}).
\newblock


\bibitem[Yu et~al\mbox{.}(2022b)]%
        {yu2022graph}
\bibfield{author}{\bibinfo{person}{Junliang Yu}, \bibinfo{person}{Hongzhi Yin},
  \bibinfo{person}{Xin Xia}, \bibinfo{person}{Tong Chen},
  \bibinfo{person}{Lizhen Cui}, {and} \bibinfo{person}{Quoc Viet~Hung Nguyen}.}
  \bibinfo{year}{2022}\natexlab{b}.
\newblock \showarticletitle{Are graph augmentations necessary? simple graph
  contrastive learning for recommendation}. In
  \bibinfo{booktitle}{\emph{Proceedings of the 45th international ACM SIGIR
  conference on research and development in information retrieval}}.
  \bibinfo{pages}{1294--1303}.
\newblock


\bibitem[Yu et~al\mbox{.}(2023)]%
        {DBLP:conf/acl/YuZQSWGWYN23}
\bibfield{author}{\bibinfo{person}{Yakun Yu}, \bibinfo{person}{Mingjun Zhao},
  \bibinfo{person}{Shiang Qi}, \bibinfo{person}{Feiran Sun},
  \bibinfo{person}{Baoxun Wang}, \bibinfo{person}{Weidong Guo},
  \bibinfo{person}{Xiaoli Wang}, \bibinfo{person}{Lei Yang}, {and}
  \bibinfo{person}{Di Niu}.} \bibinfo{year}{2023}\natexlab{}.
\newblock \showarticletitle{ConKI: Contrastive Knowledge Injection for
  Multimodal Sentiment Analysis}. In \bibinfo{booktitle}{\emph{Findings of the
  Association for Computational Linguistics: {ACL} 2023, Toronto, Canada, July
  9-14, 2023}}, \bibfield{editor}{\bibinfo{person}{Anna Rogers},
  \bibinfo{person}{Jordan~L. Boyd{-}Graber}, {and} \bibinfo{person}{Naoaki
  Okazaki}} (Eds.). \bibinfo{publisher}{Association for Computational
  Linguistics}, \bibinfo{pages}{13610--13624}.
\newblock


\bibitem[Yue et~al\mbox{.}(2022)]%
        {DBLP:conf/aaai/YueWDYHTX22}
\bibfield{author}{\bibinfo{person}{Zhihan Yue}, \bibinfo{person}{Yujing Wang},
  \bibinfo{person}{Juanyong Duan}, \bibinfo{person}{Tianmeng Yang},
  \bibinfo{person}{Congrui Huang}, \bibinfo{person}{Yunhai Tong}, {and}
  \bibinfo{person}{Bixiong Xu}.} \bibinfo{year}{2022}\natexlab{}.
\newblock \showarticletitle{TS2Vec: Towards Universal Representation of Time
  Series}. In \bibinfo{booktitle}{\emph{Thirty-Sixth {AAAI} Conference on
  Artificial Intelligence, {AAAI} 2022, Thirty-Fourth Conference on Innovative
  Applications of Artificial Intelligence, {IAAI} 2022, The Twelveth Symposium
  on Educational Advances in Artificial Intelligence, {EAAI} 2022 Virtual
  Event, February 22 - March 1, 2022}}. \bibinfo{publisher}{{AAAI} Press},
  \bibinfo{pages}{8980--8987}.
\newblock


\bibitem[Zhan et~al\mbox{.}(2022)]%
        {DBLP:conf/cvpr/ZhanYWZLZ22}
\bibfield{author}{\bibinfo{person}{Fangneng Zhan}, \bibinfo{person}{Yingchen
  Yu}, \bibinfo{person}{Rongliang Wu}, \bibinfo{person}{Jiahui Zhang},
  \bibinfo{person}{Shijian Lu}, {and} \bibinfo{person}{Changgong Zhang}.}
  \bibinfo{year}{2022}\natexlab{}.
\newblock \showarticletitle{Marginal Contrastive Correspondence for Guided
  Image Generation}. In \bibinfo{booktitle}{\emph{{IEEE/CVF} Conference on
  Computer Vision and Pattern Recognition, {CVPR} 2022, New Orleans, LA, USA,
  June 18-24, 2022}}. \bibinfo{publisher}{{IEEE}},
  \bibinfo{pages}{10653--10662}.
\newblock


\bibitem[Zhang et~al\mbox{.}(2022g)]%
        {DBLP:conf/acl/ZhangXZMA22}
\bibfield{author}{\bibinfo{person}{Dejiao Zhang}, \bibinfo{person}{Wei Xiao},
  \bibinfo{person}{Henghui Zhu}, \bibinfo{person}{Xiaofei Ma}, {and}
  \bibinfo{person}{Andrew~O. Arnold}.} \bibinfo{year}{2022}\natexlab{g}.
\newblock \showarticletitle{Virtual Augmentation Supported Contrastive Learning
  of Sentence Representations}. In \bibinfo{booktitle}{\emph{Findings of the
  Association for Computational Linguistics: {ACL} 2022, Dublin, Ireland, May
  22-27, 2022}}, \bibfield{editor}{\bibinfo{person}{Smaranda Muresan},
  \bibinfo{person}{Preslav Nakov}, {and} \bibinfo{person}{Aline Villavicencio}}
  (Eds.). \bibinfo{publisher}{Association for Computational Linguistics},
  \bibinfo{pages}{864--876}.
\newblock


\bibitem[Zhang et~al\mbox{.}(2022c)]%
        {zhang2022fairness}
\bibfield{author}{\bibinfo{person}{Fengda Zhang}, \bibinfo{person}{Kun Kuang},
  \bibinfo{person}{Long Chen}, \bibinfo{person}{Yuxuan Liu},
  \bibinfo{person}{Chao Wu}, {and} \bibinfo{person}{Jun Xiao}.}
  \bibinfo{year}{2022}\natexlab{c}.
\newblock \showarticletitle{Fairness-aware contrastive learning with partially
  annotated sensitive attributes}. In \bibinfo{booktitle}{\emph{The Eleventh
  International Conference on Learning Representations}}.
\newblock


\bibitem[Zhang et~al\mbox{.}(2021)]%
        {DBLP:conf/cvpr/0010KBLY21}
\bibfield{author}{\bibinfo{person}{Han Zhang}, \bibinfo{person}{Jing~Yu Koh},
  \bibinfo{person}{Jason Baldridge}, \bibinfo{person}{Honglak Lee}, {and}
  \bibinfo{person}{Yinfei Yang}.} \bibinfo{year}{2021}\natexlab{}.
\newblock \showarticletitle{Cross-Modal Contrastive Learning for Text-to-Image
  Generation}. In \bibinfo{booktitle}{\emph{{IEEE} Conference on Computer
  Vision and Pattern Recognition, {CVPR} 2021, virtual, June 19-25, 2021}}.
  \bibinfo{publisher}{Computer Vision Foundation / {IEEE}},
  \bibinfo{pages}{833--842}.
\newblock


\bibitem[Zhang et~al\mbox{.}(2023b)]%
        {DBLP:conf/iconip/0003LZG023}
\bibfield{author}{\bibinfo{person}{Hu Zhang}, \bibinfo{person}{Kunrui Li},
  \bibinfo{person}{Guangjun Zhang}, \bibinfo{person}{Yong Guan}, {and}
  \bibinfo{person}{Ru Li}.} \bibinfo{year}{2023}\natexlab{b}.
\newblock \showarticletitle{Multi-granularity Contrastive Siamese Networks for
  Abstractive Text Summarization}. In \bibinfo{booktitle}{\emph{Neural
  Information Processing - 30th International Conference, {ICONIP} 2023,
  Changsha, China, November 20-23, 2023, Proceedings, Part {XII}}}
  \emph{(\bibinfo{series}{Communications in Computer and Information Science},
  Vol.~\bibinfo{volume}{1966})}, \bibfield{editor}{\bibinfo{person}{Biao Luo},
  \bibinfo{person}{Long Cheng}, \bibinfo{person}{Zheng{-}Guang Wu},
  \bibinfo{person}{Hongyi Li}, {and} \bibinfo{person}{Chaojie Li}} (Eds.).
  \bibinfo{publisher}{Springer}, \bibinfo{pages}{198--210}.
\newblock


\bibitem[Zhang et~al\mbox{.}(2023a)]%
        {DBLP:journals/corr/abs-2304-00685}
\bibfield{author}{\bibinfo{person}{Jingyi Zhang}, \bibinfo{person}{Jiaxing
  Huang}, \bibinfo{person}{Sheng Jin}, {and} \bibinfo{person}{Shijian Lu}.}
  \bibinfo{year}{2023}\natexlab{a}.
\newblock \showarticletitle{Vision-Language Models for Vision Tasks: {A}
  Survey}.
\newblock \bibinfo{journal}{\emph{CoRR}}  \bibinfo{volume}{abs/2304.00685}
  (\bibinfo{year}{2023}).
\newblock


\bibitem[Zhang et~al\mbox{.}(2022d)]%
        {DBLP:conf/naacl/ZhangMAHK22}
\bibfield{author}{\bibinfo{person}{Miaoran Zhang}, \bibinfo{person}{Marius
  Mosbach}, \bibinfo{person}{David~Ifeoluwa Adelani},
  \bibinfo{person}{Michael~A. Hedderich}, {and} \bibinfo{person}{Dietrich
  Klakow}.} \bibinfo{year}{2022}\natexlab{d}.
\newblock \showarticletitle{{MCSE:} Multimodal Contrastive Learning of Sentence
  Embeddings}. In \bibinfo{booktitle}{\emph{Proceedings of the 2022 Conference
  of the North American Chapter of the Association for Computational
  Linguistics: Human Language Technologies, {NAACL} 2022, Seattle, WA, United
  States, July 10-15, 2022}}, \bibfield{editor}{\bibinfo{person}{Marine
  Carpuat}, \bibinfo{person}{Marie{-}Catherine de~Marneffe}, {and}
  \bibinfo{person}{Iv{\'{a}}n Vladimir~Meza Ru{\'{\i}}z}} (Eds.).
  \bibinfo{publisher}{Association for Computational Linguistics},
  \bibinfo{pages}{5959--5969}.
\newblock


\bibitem[Zhang and R{\'{e}}(2022)]%
        {DBLP:conf/nips/ZhangR22}
\bibfield{author}{\bibinfo{person}{Michael Zhang} {and}
  \bibinfo{person}{Christopher R{\'{e}}}.} \bibinfo{year}{2022}\natexlab{}.
\newblock \showarticletitle{Contrastive Adapters for Foundation Model Group
  Robustness}. In \bibinfo{booktitle}{\emph{Advances in Neural Information
  Processing Systems 35: Annual Conference on Neural Information Processing
  Systems 2022, NeurIPS 2022, New Orleans, LA, USA, November 28 - December 9,
  2022}}, \bibfield{editor}{\bibinfo{person}{Sanmi Koyejo},
  \bibinfo{person}{S.~Mohamed}, \bibinfo{person}{A.~Agarwal},
  \bibinfo{person}{Danielle Belgrave}, \bibinfo{person}{K.~Cho}, {and}
  \bibinfo{person}{A.~Oh}} (Eds.).
\newblock


\bibitem[Zhang et~al\mbox{.}(2018)]%
        {zhang2018binary}
\bibfield{author}{\bibinfo{person}{Min-Ling Zhang}, \bibinfo{person}{Yu-Kun
  Li}, \bibinfo{person}{Xu-Ying Liu}, {and} \bibinfo{person}{Xin Geng}.}
  \bibinfo{year}{2018}\natexlab{}.
\newblock \showarticletitle{Binary relevance for multi-label learning: an
  overview}.
\newblock \bibinfo{journal}{\emph{Frontiers of Computer Science}}
  \bibinfo{volume}{12}, \bibinfo{number}{2} (\bibinfo{year}{2018}),
  \bibinfo{pages}{191--202}.
\newblock


\bibitem[Zhang et~al\mbox{.}(2022b)]%
        {zhang-etal-2022-contrastive-data}
\bibfield{author}{\bibinfo{person}{Rui Zhang}, \bibinfo{person}{Yangfeng Ji},
  \bibinfo{person}{Yue Zhang}, {and} \bibinfo{person}{Rebecca~J. Passonneau}.}
  \bibinfo{year}{2022}\natexlab{b}.
\newblock \showarticletitle{Contrastive Data and Learning for Natural Language
  Processing}. In \bibinfo{booktitle}{\emph{Proceedings of the 2022 Conference
  of the North American Chapter of the Association for Computational
  Linguistics: Human Language Technologies: Tutorial Abstracts}}.
  \bibinfo{publisher}{Association for Computational Linguistics},
  \bibinfo{address}{Seattle, United States}.
\newblock


\bibitem[Zhang et~al\mbox{.}(2020)]%
        {zhang2020commdgi}
\bibfield{author}{\bibinfo{person}{Tianqi Zhang}, \bibinfo{person}{Yun Xiong},
  \bibinfo{person}{Jiawei Zhang}, \bibinfo{person}{Yao Zhang},
  \bibinfo{person}{Yizhu Jiao}, {and} \bibinfo{person}{Yangyong Zhu}.}
  \bibinfo{year}{2020}\natexlab{}.
\newblock \showarticletitle{CommDGI: community detection oriented deep graph
  infomax}. In \bibinfo{booktitle}{\emph{Proceedings of the 29th ACM
  international conference on information \& knowledge management}}.
  \bibinfo{pages}{1843--1852}.
\newblock


\bibitem[Zhang et~al\mbox{.}(2022h)]%
        {zhang2022frequency}
\bibfield{author}{\bibinfo{person}{Tong Zhang}, \bibinfo{person}{Wei Ye},
  \bibinfo{person}{Baosong Yang}, \bibinfo{person}{Long Zhang},
  \bibinfo{person}{Xingzhang Ren}, \bibinfo{person}{Dayiheng Liu},
  \bibinfo{person}{Jinan Sun}, \bibinfo{person}{Shikun Zhang},
  \bibinfo{person}{Haibo Zhang}, {and} \bibinfo{person}{Wen Zhao}.}
  \bibinfo{year}{2022}\natexlab{h}.
\newblock \showarticletitle{Frequency-aware contrastive learning for neural
  machine translation}. In \bibinfo{booktitle}{\emph{Proceedings of the AAAI
  Conference on Artificial Intelligence}}, Vol.~\bibinfo{volume}{36}.
  \bibinfo{pages}{11712--11720}.
\newblock


\bibitem[Zhang et~al\mbox{.}(2022e)]%
        {DBLP:conf/emnlp/Zhang0MCTN022}
\bibfield{author}{\bibinfo{person}{Wenqi Zhang}, \bibinfo{person}{Yongliang
  Shen}, \bibinfo{person}{Yanna Ma}, \bibinfo{person}{Xiaoxia Cheng},
  \bibinfo{person}{Zeqi Tan}, \bibinfo{person}{Qingpeng Nong}, {and}
  \bibinfo{person}{Weiming Lu}.} \bibinfo{year}{2022}\natexlab{e}.
\newblock \showarticletitle{Multi-View Reasoning: Consistent Contrastive
  Learning for Math Word Problem}. In \bibinfo{booktitle}{\emph{Findings of the
  Association for Computational Linguistics: {EMNLP} 2022, Abu Dhabi, United
  Arab Emirates, December 7-11, 2022}}. \bibinfo{publisher}{Association for
  Computational Linguistics}, \bibinfo{pages}{1103--1116}.
\newblock


\bibitem[Zhang et~al\mbox{.}(2022i)]%
        {zhang2022self}
\bibfield{author}{\bibinfo{person}{Xiang Zhang}, \bibinfo{person}{Ziyuan Zhao},
  \bibinfo{person}{Theodoros Tsiligkaridis}, {and} \bibinfo{person}{Marinka
  Zitnik}.} \bibinfo{year}{2022}\natexlab{i}.
\newblock \showarticletitle{Self-supervised contrastive pre-training for time
  series via time-frequency consistency}.
\newblock \bibinfo{journal}{\emph{Advances in Neural Information Processing
  Systems}}  \bibinfo{volume}{35} (\bibinfo{year}{2022}),
  \bibinfo{pages}{3988--4003}.
\newblock


\bibitem[Zhang et~al\mbox{.}(2022a)]%
        {DBLP:conf/cvpr/ZhangC022}
\bibfield{author}{\bibinfo{person}{Yanan Zhang}, \bibinfo{person}{Jiaxin Chen},
  {and} \bibinfo{person}{Di Huang}.} \bibinfo{year}{2022}\natexlab{a}.
\newblock \showarticletitle{CAT-Det: Contrastively Augmented Transformer for
  Multimodal 3D Object Detection}. In \bibinfo{booktitle}{\emph{{IEEE/CVF}
  Conference on Computer Vision and Pattern Recognition, {CVPR} 2022, New
  Orleans, LA, USA, June 18-24, 2022}}. \bibinfo{publisher}{{IEEE}},
  \bibinfo{pages}{898--907}.
\newblock


\bibitem[Zhang et~al\mbox{.}(2022f)]%
        {DBLP:conf/ijcnn/ZhangSCXZL22}
\bibfield{author}{\bibinfo{person}{Zijian Zhang}, \bibinfo{person}{Chang Shu},
  \bibinfo{person}{Youxin Chen}, \bibinfo{person}{Jing Xiao},
  \bibinfo{person}{Qian Zhang}, {and} \bibinfo{person}{Zheng Lu}.}
  \bibinfo{year}{2022}\natexlab{f}.
\newblock \showarticletitle{{ICAF:} Iterative Contrastive Alignment Framework
  for Multimodal Abstractive Summarization}. In
  \bibinfo{booktitle}{\emph{International Joint Conference on Neural Networks,
  {IJCNN} 2022, Padua, Italy, July 18-23, 2022}}. \bibinfo{publisher}{{IEEE}},
  \bibinfo{pages}{1--8}.
\newblock


\bibitem[Zhao et~al\mbox{.}(2021)]%
        {zhao2021graph}
\bibfield{author}{\bibinfo{person}{Han Zhao}, \bibinfo{person}{Xu Yang},
  \bibinfo{person}{Zhenru Wang}, \bibinfo{person}{Erkun Yang}, {and}
  \bibinfo{person}{Cheng Deng}.} \bibinfo{year}{2021}\natexlab{}.
\newblock \showarticletitle{Graph Debiased Contrastive Learning with Joint
  Representation Clustering.}. In \bibinfo{booktitle}{\emph{IJCAI}}.
  \bibinfo{pages}{3434--3440}.
\newblock


\bibitem[Zhao et~al\mbox{.}(2023)]%
        {DBLP:journals/corr/abs-2309-08929}
\bibfield{author}{\bibinfo{person}{Kaiyan Zhao}, \bibinfo{person}{Qiyu Wu},
  \bibinfo{person}{Xin{-}Qiang Cai}, {and} \bibinfo{person}{Yoshimasa
  Tsuruoka}.} \bibinfo{year}{2023}\natexlab{}.
\newblock \showarticletitle{Leveraging Multi-lingual Positive Instances in
  Contrastive Learning to Improve Sentence Embedding}.
\newblock \bibinfo{journal}{\emph{CoRR}}  \bibinfo{volume}{abs/2309.08929}
  (\bibinfo{year}{2023}).
\newblock


\bibitem[Zheng et~al\mbox{.}(2023b)]%
        {DBLP:conf/acl/ZhengSYLPZXZ23}
\bibfield{author}{\bibinfo{person}{Kai Zheng}, \bibinfo{person}{Qingfeng Sun},
  \bibinfo{person}{Yaming Yang}, \bibinfo{person}{Tengchao Lv},
  \bibinfo{person}{Yeyong Pi}, \bibinfo{person}{Changlin Zhao},
  \bibinfo{person}{Fei Xu}, {and} \bibinfo{person}{Qi Zhang}.}
  \bibinfo{year}{2023}\natexlab{b}.
\newblock \showarticletitle{Adversarial Knowledge Stimulated Contrastive
  Prompting for Few-shot Language Learners}. In
  \bibinfo{booktitle}{\emph{Findings of the Association for Computational
  Linguistics: {ACL} 2023, Toronto, Canada, July 9-14, 2023}},
  \bibfield{editor}{\bibinfo{person}{Anna Rogers}, \bibinfo{person}{Jordan~L.
  Boyd{-}Graber}, {and} \bibinfo{person}{Naoaki Okazaki}} (Eds.).
  \bibinfo{publisher}{Association for Computational Linguistics},
  \bibinfo{pages}{13495--13507}.
\newblock


\bibitem[Zheng et~al\mbox{.}(2024)]%
        {zheng2024multi}
\bibfield{author}{\bibinfo{person}{Lecheng Zheng}, \bibinfo{person}{Zhengzhang
  Chen}, \bibinfo{person}{Jingrui He}, {and} \bibinfo{person}{Haifeng Chen}.}
  \bibinfo{year}{2024}\natexlab{}.
\newblock \showarticletitle{Multi-modal Causal Structure Learning and Root
  Cause Analysis}.
\newblock \bibinfo{journal}{\emph{arXiv preprint arXiv:2402.02357}}
  (\bibinfo{year}{2024}).
\newblock


\bibitem[Zheng et~al\mbox{.}(2019)]%
        {DBLP:conf/sdm/ZhengCH19}
\bibfield{author}{\bibinfo{person}{Lecheng Zheng}, \bibinfo{person}{Yu Cheng},
  {and} \bibinfo{person}{Jingrui He}.} \bibinfo{year}{2019}\natexlab{}.
\newblock \showarticletitle{Deep Multimodality Model for Multi-task Multi-view
  Learning}. In \bibinfo{booktitle}{\emph{Proceedings of the 2019 {SIAM}
  International Conference on Data Mining, {SDM} 2019, Calgary, Alberta,
  Canada, May 2-4, 2019}}, \bibfield{editor}{\bibinfo{person}{Tanya~Y.
  Berger{-}Wolf} {and} \bibinfo{person}{Nitesh~V. Chawla}} (Eds.).
  \bibinfo{publisher}{{SIAM}}, \bibinfo{pages}{10--18}.
\newblock


\bibitem[Zheng et~al\mbox{.}(2021a)]%
        {DBLP:conf/www/ZhengCYCH21}
\bibfield{author}{\bibinfo{person}{Lecheng Zheng}, \bibinfo{person}{Yu Cheng},
  \bibinfo{person}{Hongxia Yang}, \bibinfo{person}{Nan Cao}, {and}
  \bibinfo{person}{Jingrui He}.} \bibinfo{year}{2021}\natexlab{a}.
\newblock \showarticletitle{Deep Co-Attention Network for Multi-View Subspace
  Learning}. In \bibinfo{booktitle}{\emph{{WWW} '21: The Web Conference 2021,
  Virtual Event / Ljubljana, Slovenia, April 19-23, 2021}},
  \bibfield{editor}{\bibinfo{person}{Jure Leskovec}, \bibinfo{person}{Marko
  Grobelnik}, \bibinfo{person}{Marc Najork}, \bibinfo{person}{Jie Tang}, {and}
  \bibinfo{person}{Leila Zia}} (Eds.). \bibinfo{publisher}{{ACM} / {IW3C2}},
  \bibinfo{pages}{1528--1539}.
\newblock


\bibitem[Zheng et~al\mbox{.}(2021b)]%
        {zheng2021deeper}
\bibfield{author}{\bibinfo{person}{Lecheng Zheng}, \bibinfo{person}{Dongqi Fu},
  \bibinfo{person}{Ross Maciejewski}, {and} \bibinfo{person}{Jingrui He}.}
  \bibinfo{year}{2021}\natexlab{b}.
\newblock \showarticletitle{Deeper-GXX: deepening arbitrary GNNs}.
\newblock \bibinfo{journal}{\emph{arXiv preprint arXiv:2110.13798}}
  (\bibinfo{year}{2021}).
\newblock


\bibitem[Zheng et~al\mbox{.}(2022)]%
        {DBLP:conf/kdd/ZhengXZH22}
\bibfield{author}{\bibinfo{person}{Lecheng Zheng}, \bibinfo{person}{Jinjun
  Xiong}, \bibinfo{person}{Yada Zhu}, {and} \bibinfo{person}{Jingrui He}.}
  \bibinfo{year}{2022}\natexlab{}.
\newblock \showarticletitle{Contrastive Learning with Complex Heterogeneity}.
  In \bibinfo{booktitle}{\emph{{KDD} '22: The 28th {ACM} {SIGKDD} Conference on
  Knowledge Discovery and Data Mining, Washington, DC, USA, August 14 - 18,
  2022}}, \bibfield{editor}{\bibinfo{person}{Aidong Zhang} {and}
  \bibinfo{person}{Huzefa Rangwala}} (Eds.). \bibinfo{publisher}{{ACM}},
  \bibinfo{pages}{2594--2604}.
\newblock


\bibitem[Zheng et~al\mbox{.}(2023c)]%
        {zheng2023fairness}
\bibfield{author}{\bibinfo{person}{Lecheng Zheng}, \bibinfo{person}{Yada Zhu},
  {and} \bibinfo{person}{Jingrui He}.} \bibinfo{year}{2023}\natexlab{c}.
\newblock \showarticletitle{Fairness-aware multi-view clustering}. In
  \bibinfo{booktitle}{\emph{Proceedings of the 2023 SIAM International
  Conference on Data Mining (SDM)}}. SIAM, \bibinfo{pages}{856--864}.
\newblock


\bibitem[Zheng et~al\mbox{.}(2021d)]%
        {DBLP:journals/corr/abs-2105-09401}
\bibfield{author}{\bibinfo{person}{Lecheng Zheng}, \bibinfo{person}{Yada Zhu},
  \bibinfo{person}{Jingrui He}, {and} \bibinfo{person}{Jinjun Xiong}.}
  \bibinfo{year}{2021}\natexlab{d}.
\newblock \showarticletitle{Heterogeneous Contrastive Learning}.
\newblock \bibinfo{journal}{\emph{CoRR}}  \bibinfo{volume}{abs/2105.09401}
  (\bibinfo{year}{2021}).
\newblock


\bibitem[Zheng et~al\mbox{.}(2021c)]%
        {DBLP:conf/iccv/Zheng0Y0Z0021}
\bibfield{author}{\bibinfo{person}{Mingkai Zheng}, \bibinfo{person}{Fei Wang},
  \bibinfo{person}{Shan You}, \bibinfo{person}{Chen Qian},
  \bibinfo{person}{Changshui Zhang}, \bibinfo{person}{Xiaogang Wang}, {and}
  \bibinfo{person}{Chang Xu}.} \bibinfo{year}{2021}\natexlab{c}.
\newblock \showarticletitle{Weakly Supervised Contrastive Learning}. In
  \bibinfo{booktitle}{\emph{2021 {IEEE/CVF} International Conference on
  Computer Vision, {ICCV} 2021, Montreal, QC, Canada, October 10-17, 2021}}.
  \bibinfo{publisher}{{IEEE}}, \bibinfo{pages}{10022--10031}.
\newblock


\bibitem[Zheng et~al\mbox{.}(2023a)]%
        {zheng2023simts}
\bibfield{author}{\bibinfo{person}{Xiaochen Zheng}, \bibinfo{person}{Xingyu
  Chen}, \bibinfo{person}{Manuel Sch{\"u}rch}, \bibinfo{person}{Amina
  Mollaysa}, \bibinfo{person}{Ahmed Allam}, {and} \bibinfo{person}{Michael
  Krauthammer}.} \bibinfo{year}{2023}\natexlab{a}.
\newblock \showarticletitle{SimTS: rethinking contrastive representation
  learning for time series forecasting}.
\newblock \bibinfo{journal}{\emph{arXiv preprint arXiv:2303.18205}}
  (\bibinfo{year}{2023}).
\newblock


\bibitem[Zhou et~al\mbox{.}(2022)]%
        {DBLP:conf/cikm/ZhouZF0H22}
\bibfield{author}{\bibinfo{person}{Dawei Zhou}, \bibinfo{person}{Lecheng
  Zheng}, \bibinfo{person}{Dongqi Fu}, \bibinfo{person}{Jiawei Han}, {and}
  \bibinfo{person}{Jingrui He}.} \bibinfo{year}{2022}\natexlab{}.
\newblock \showarticletitle{MentorGNN: Deriving Curriculum for Pre-Training
  GNNs}. In \bibinfo{booktitle}{\emph{Proceedings of the 31st {ACM}
  International Conference on Information {\&} Knowledge Management, Atlanta,
  GA, USA, October 17-21, 2022}},
  \bibfield{editor}{\bibinfo{person}{Mohammad~Al Hasan} {and}
  \bibinfo{person}{Li~Xiong}} (Eds.). \bibinfo{publisher}{{ACM}},
  \bibinfo{pages}{2721--2731}.
\newblock


\bibitem[Zhou et~al\mbox{.}(2020)]%
        {DBLP:conf/www/ZhouZZLH20}
\bibfield{author}{\bibinfo{person}{Dawei Zhou}, \bibinfo{person}{Lecheng
  Zheng}, \bibinfo{person}{Yada Zhu}, \bibinfo{person}{Jianbo Li}, {and}
  \bibinfo{person}{Jingrui He}.} \bibinfo{year}{2020}\natexlab{}.
\newblock \showarticletitle{Domain Adaptive Multi-Modality Neural Attention
  Network for Financial Forecasting}. In \bibinfo{booktitle}{\emph{{WWW} '20:
  The Web Conference 2020, Taipei, Taiwan, April 20-24, 2020}},
  \bibfield{editor}{\bibinfo{person}{Yennun Huang}, \bibinfo{person}{Irwin
  King}, \bibinfo{person}{Tie{-}Yan Liu}, {and} \bibinfo{person}{Maarten van
  Steen}} (Eds.). \bibinfo{publisher}{{ACM} / {IW3C2}},
  \bibinfo{pages}{2230--2240}.
\newblock


\bibitem[Zhou et~al\mbox{.}(2021)]%
        {zhou2021soda}
\bibfield{author}{\bibinfo{person}{Jieli Zhou}, \bibinfo{person}{Baoyu Jing},
  \bibinfo{person}{Zeya Wang}, \bibinfo{person}{Hongyi Xin}, {and}
  \bibinfo{person}{Hanghang Tong}.} \bibinfo{year}{2021}\natexlab{}.
\newblock \showarticletitle{Soda: Detecting covid-19 in chest x-rays with
  semi-supervised open set domain adaptation}.
\newblock \bibinfo{journal}{\emph{IEEE/ACM Transactions on Computational
  Biology and Bioinformatics}} \bibinfo{volume}{19}, \bibinfo{number}{5}
  (\bibinfo{year}{2021}), \bibinfo{pages}{2605--2612}.
\newblock


\bibitem[Zhou et~al\mbox{.}(2024)]%
        {zhou2024one}
\bibfield{author}{\bibinfo{person}{Tian Zhou}, \bibinfo{person}{Peisong Niu},
  \bibinfo{person}{Liang Sun}, \bibinfo{person}{Rong Jin}, {et~al\mbox{.}}}
  \bibinfo{year}{2024}\natexlab{}.
\newblock \showarticletitle{One fits all: Power general time series analysis by
  pretrained lm}.
\newblock \bibinfo{journal}{\emph{Advances in neural information processing
  systems}}  \bibinfo{volume}{36} (\bibinfo{year}{2024}).
\newblock


\bibitem[Zhu et~al\mbox{.}(2021)]%
        {DBLP:conf/www/0001XYLWW21}
\bibfield{author}{\bibinfo{person}{Yanqiao Zhu}, \bibinfo{person}{Yichen Xu},
  \bibinfo{person}{Feng Yu}, \bibinfo{person}{Qiang Liu}, \bibinfo{person}{Shu
  Wu}, {and} \bibinfo{person}{Liang Wang}.} \bibinfo{year}{2021}\natexlab{}.
\newblock \showarticletitle{Graph Contrastive Learning with Adaptive
  Augmentation}. In \bibinfo{booktitle}{\emph{{WWW} '21: The Web Conference
  2021, Virtual Event / Ljubljana, Slovenia, April 19-23, 2021}},
  \bibfield{editor}{\bibinfo{person}{Jure Leskovec}, \bibinfo{person}{Marko
  Grobelnik}, \bibinfo{person}{Marc Najork}, \bibinfo{person}{Jie Tang}, {and}
  \bibinfo{person}{Leila Zia}} (Eds.). \bibinfo{publisher}{{ACM} / {IW3C2}},
  \bibinfo{pages}{2069--2080}.
\newblock


\bibitem[Zoph et~al\mbox{.}(2020)]%
        {zoph2020learning}
\bibfield{author}{\bibinfo{person}{Barret Zoph}, \bibinfo{person}{Ekin~D
  Cubuk}, \bibinfo{person}{Golnaz Ghiasi}, \bibinfo{person}{Tsung-Yi Lin},
  \bibinfo{person}{Jonathon Shlens}, {and} \bibinfo{person}{Quoc~V Le}.}
  \bibinfo{year}{2020}\natexlab{}.
\newblock \showarticletitle{Learning data augmentation strategies for object
  detection}. In \bibinfo{booktitle}{\emph{Computer Vision--ECCV 2020: 16th
  European Conference, Glasgow, UK, August 23--28, 2020, Proceedings, Part
  XXVII 16}}. Springer, \bibinfo{pages}{566--583}.
\newblock


\bibitem[Zou et~al\mbox{.}(2023)]%
        {DBLP:conf/mm/ZouHS23}
\bibfield{author}{\bibinfo{person}{Shihao Zou}, \bibinfo{person}{Xianying
  Huang}, {and} \bibinfo{person}{Xudong Shen}.}
  \bibinfo{year}{2023}\natexlab{}.
\newblock \showarticletitle{Multimodal Prompt Transformer with Hybrid
  Contrastive Learning for Emotion Recognition in Conversation}. In
  \bibinfo{booktitle}{\emph{Proceedings of the 31st {ACM} International
  Conference on Multimedia, {MM} 2023, Ottawa, ON, Canada, 29 October 2023- 3
  November 2023}}, \bibfield{editor}{\bibinfo{person}{Abdulmotaleb
  El{-}Saddik}, \bibinfo{person}{Tao Mei}, \bibinfo{person}{Rita Cucchiara},
  \bibinfo{person}{Marco Bertini}, \bibinfo{person}{Diana Patricia~Tobon
  Vallejo}, \bibinfo{person}{Pradeep~K. Atrey}, {and}
  \bibinfo{person}{M.~Shamim Hossain}} (Eds.). \bibinfo{publisher}{{ACM}},
  \bibinfo{pages}{5994--6003}.
\newblock


\end{thebibliography}

% %%
% %% If your work has an appendix, this is the place to put it.
% \appendix









\end{document}